\documentclass[a4paper]{Microsoft}

\usepackage{lipsum}
\usepackage{hyperref}
\usepackage{url}

\usepackage{graphicx}
\usepackage{xcolor}
\graphicspath{ {./images/} }

\usepackage{amsfonts}

\usepackage{mathtools}
\usepackage{listings}
\usepackage{bbm}
\usepackage{placeins}
\usepackage{enumitem}
\usepackage{comment}
\usepackage{subcaption}
\usepackage[percent]{overpic}

\usepackage{amsthm}
\usepackage[utf8]{inputenc}

\title{Scaling Laws for Pre-training Agents and World Models}

\author{Tim Pearce$^*$, Tabish Rashid$^*$, Dave Bignell, Raluca Georgescu, Sam Devlin, Katja Hofmann \\
Microsoft Research \\
$^*$Equal contribution 
}

\begin{document}

\maketitle

\begin{abstract}

The performance of embodied agents has been shown to improve by increasing model parameters, dataset size, and compute. 
This has been demonstrated in domains from robotics to video games, when generative learning objectives on offline datasets (pre-training) are used to model an agent's behavior (imitation learning) or their environment (world modeling).
This paper characterizes the role of scale in these tasks more precisely. 
Going beyond the simple intuition that `bigger is better', we show that the same types of power laws found in language modeling also arise in world modeling and imitation learning (e.g. between loss and optimal model size).
However, the coefficients of these laws are heavily influenced by the tokenizer, task \& architecture -- this has important implications on the optimal sizing of models and data.

\end{abstract}

\section{Introduction}
\label{sec_intro}

Much progress in AI in the early 2020's has been driven by increasing model size, dataset size, and training compute. Whilst conceptually simple, the importance of this practice has led to an emerging subfield studying the \textit{science of scaling}. This field answers questions such as how to estimate the benefit of increased compute investment, or how to optimally trade-off model and dataset size.

The role of scale in pre-training is until now best understood in the context of large language models (LLMs).
Following the observation that the empirical relationship between loss and key scaling quantities can be accurately described by power laws \citep{kaplan2020scaling}, ensuing work studied the precise trade-off between model and dataset size \citep{hoffmann2022training}, as well as considerations about inference compute \citep{sardana2023beyondchinchilla}, 
repeated training data \citep{muennighoff2024repeatscaling}, 
parameter counting \citep{pearce2024reconciling}, and more (Section \ref{sec_related}).

In comparison, less is understood about scaling in \textit{embodied AI}. Recent high-impact works show increasing model and dataset size can lead to ever more capable agents for two pre-training objectives; behavior cloning (BC) \citep{reed2022gato, baker2022vpt, brohan2023rt2} and world modeling (WM) \citep{hafner2020mastering, hu2023gaia, yang2023unisim, bruce2024genie}. 
Such works typically demonstrate the benefit of scale through ablations over a few model sizes, shown in terms of downstream agent performance, confirming the intuition that `bigger is better' (\cite{sartor2024neuralscaleembodied} provide an aggregated analysis).
However, this leaves a large gap to the precise understanding of scale in LLMs, where for a given increase in compute, models can be sized optimally, and their performance accurately predicted.

\begin{figure}[t!]
\centering
\begin{subfigure}{0.48\textwidth}
    \centering
    \caption{WM-Token-256}
    \vspace{-0.25in}
    \begin{picture}(200,170)
        \put(0,0){\includegraphics[width=\textwidth]{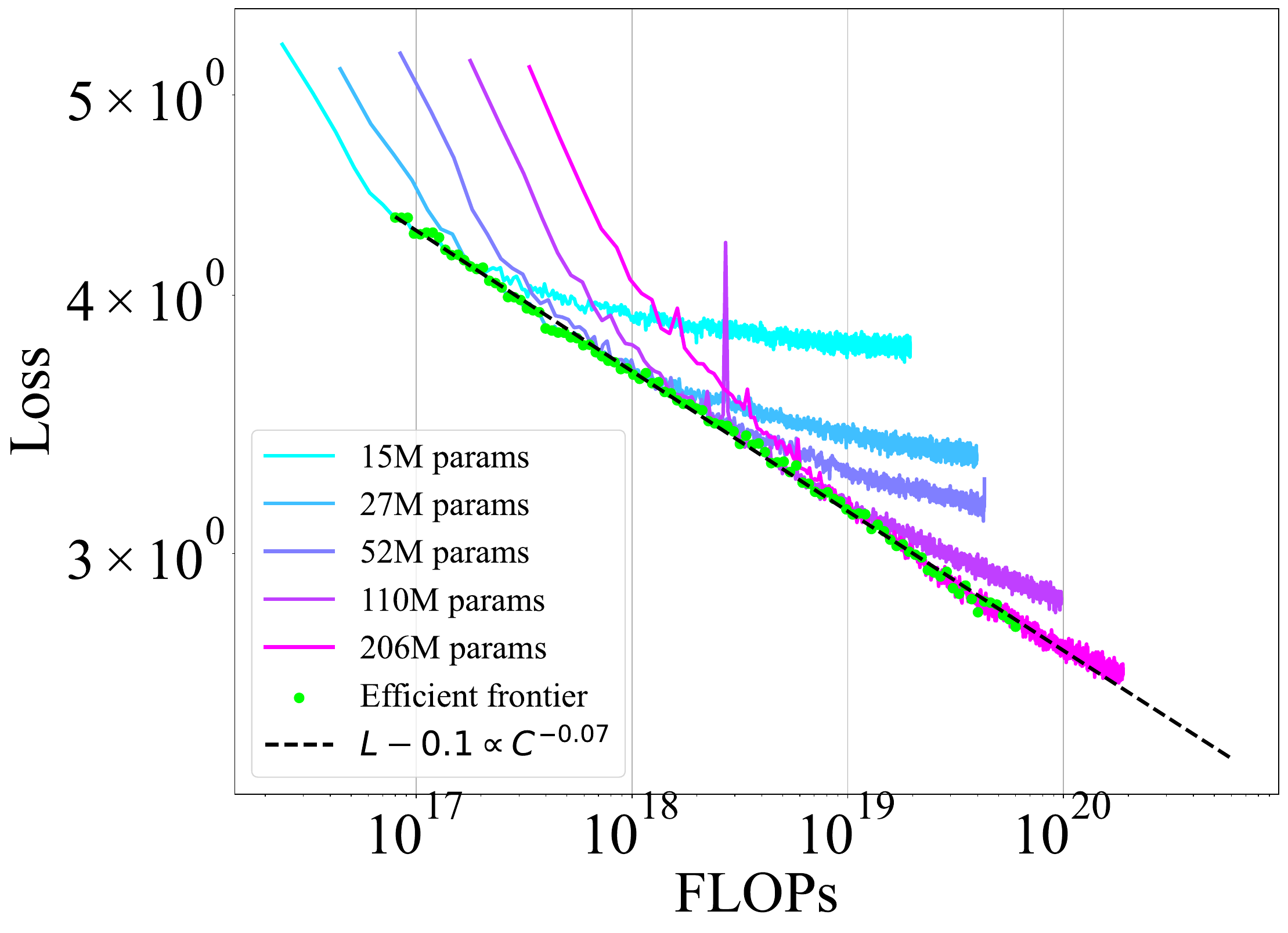}}
        \put(135,140){\small $N_\text{optimal} \propto C^{0.49}$}
        \put(135,120){\small $D_\text{optimal} \propto C^{0.51}$}
    \end{picture}
\end{subfigure}
\hfill
\begin{subfigure}{0.48\textwidth}
    \centering
    \caption{WM-Token-540}
    \vspace{-0.25in}
    \begin{picture}(200,170)
        \put(0,0){\includegraphics[width=\textwidth]{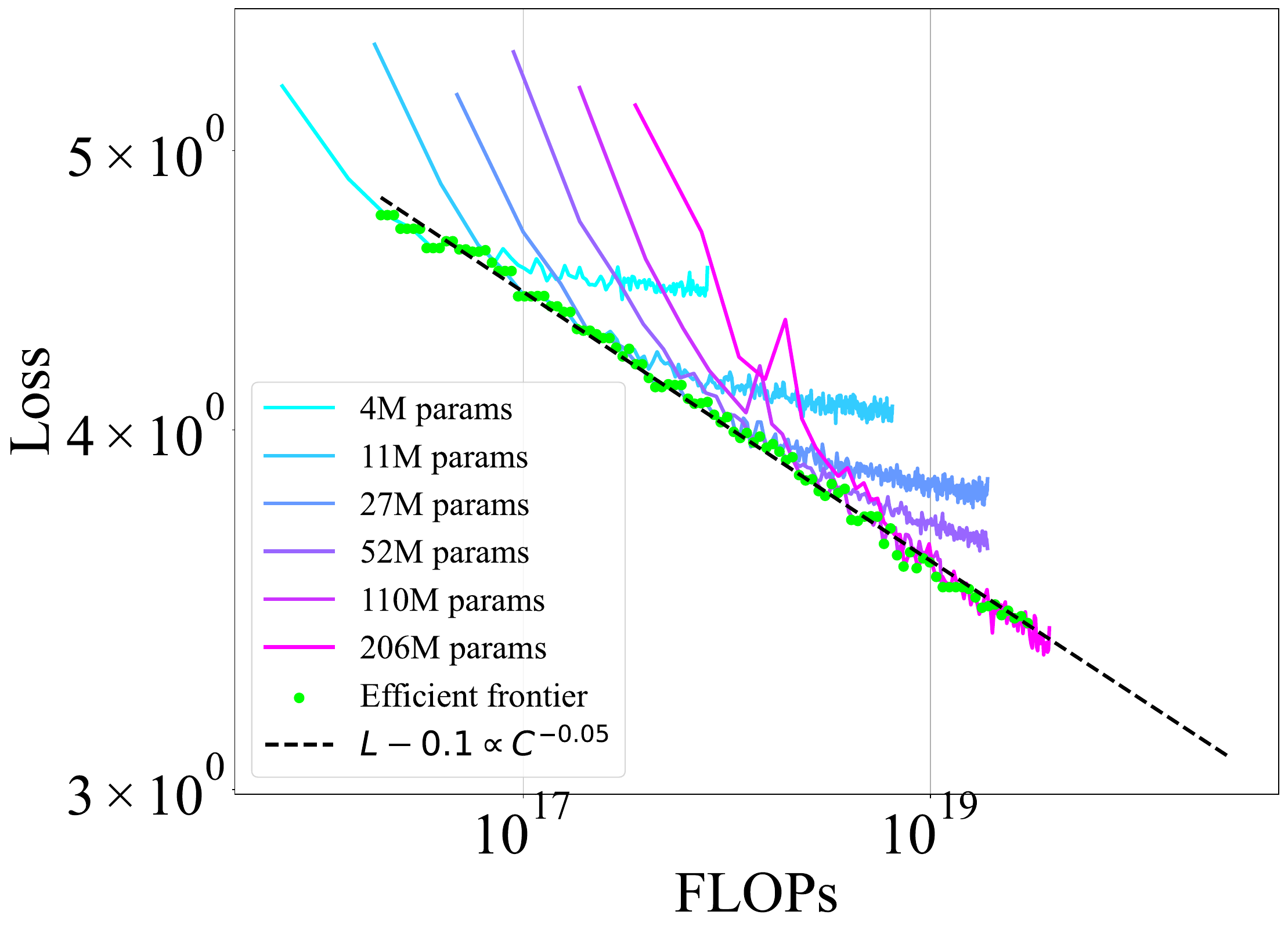}}
        \put(135,140){\small $N_\text{optimal} \propto C^{0.62}$}
        \put(135,120){\small $D_\text{optimal} \propto C^{0.38}$}
    \end{picture}
\end{subfigure}

\begin{subfigure}{0.48\textwidth}
    \centering
    \caption{BC-Token-540}
    \vspace{-0.25in}
    \begin{picture}(200,170)
        \put(0,0){\includegraphics[width=\textwidth]{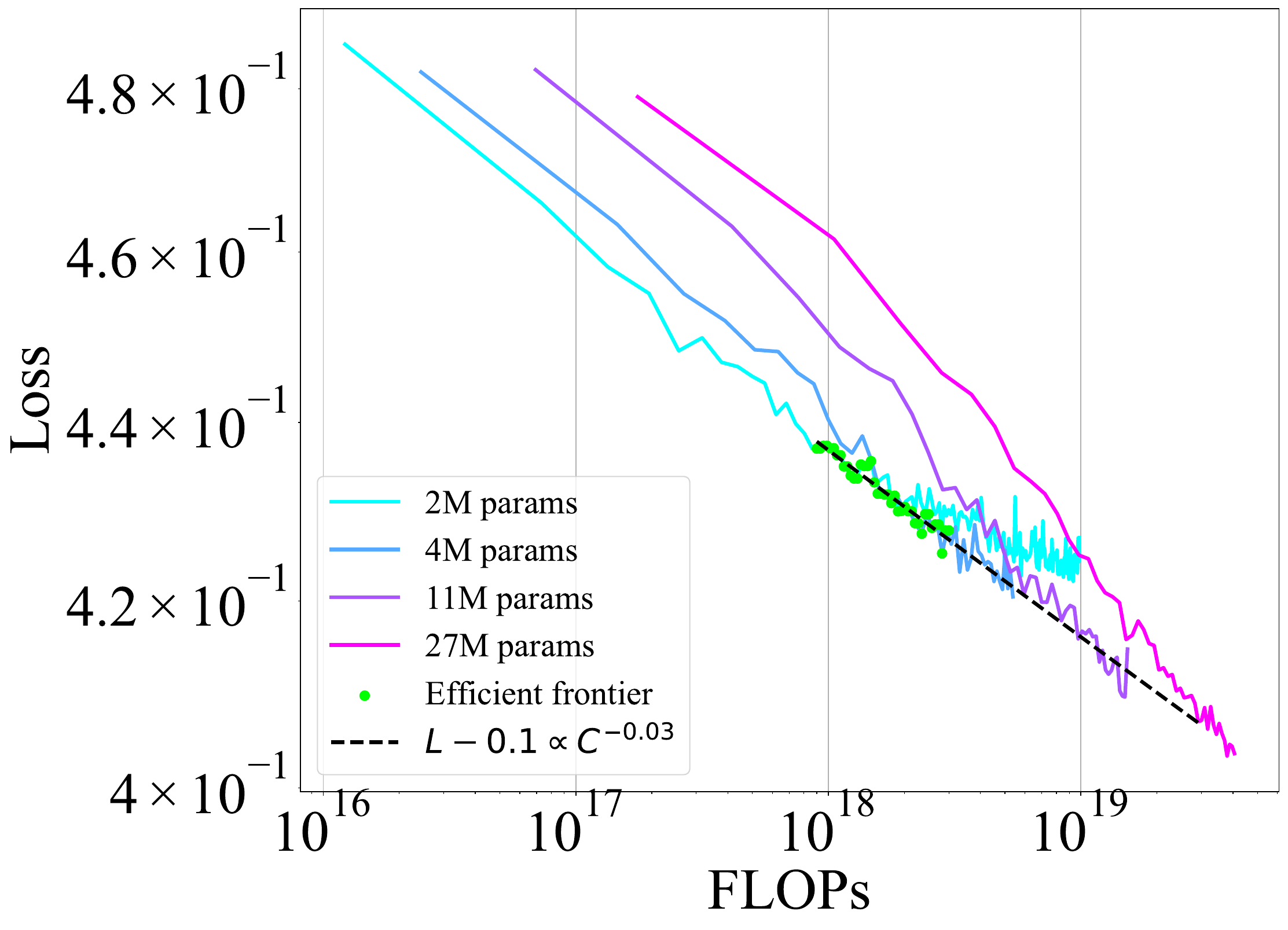}}
        \put(135,140){\small $N_\text{optimal} \propto C^{0.32}$}
        \put(135,120){\small $D_\text{optimal} \propto C^{0.78}$}
    \end{picture}
\end{subfigure}
\hfill
\begin{subfigure}{0.48\textwidth}
    \centering
    \caption{BC-CNN}
    \vspace{-0.25in}
    \begin{picture}(200,170)
        \put(0,0){\includegraphics[width=\textwidth]{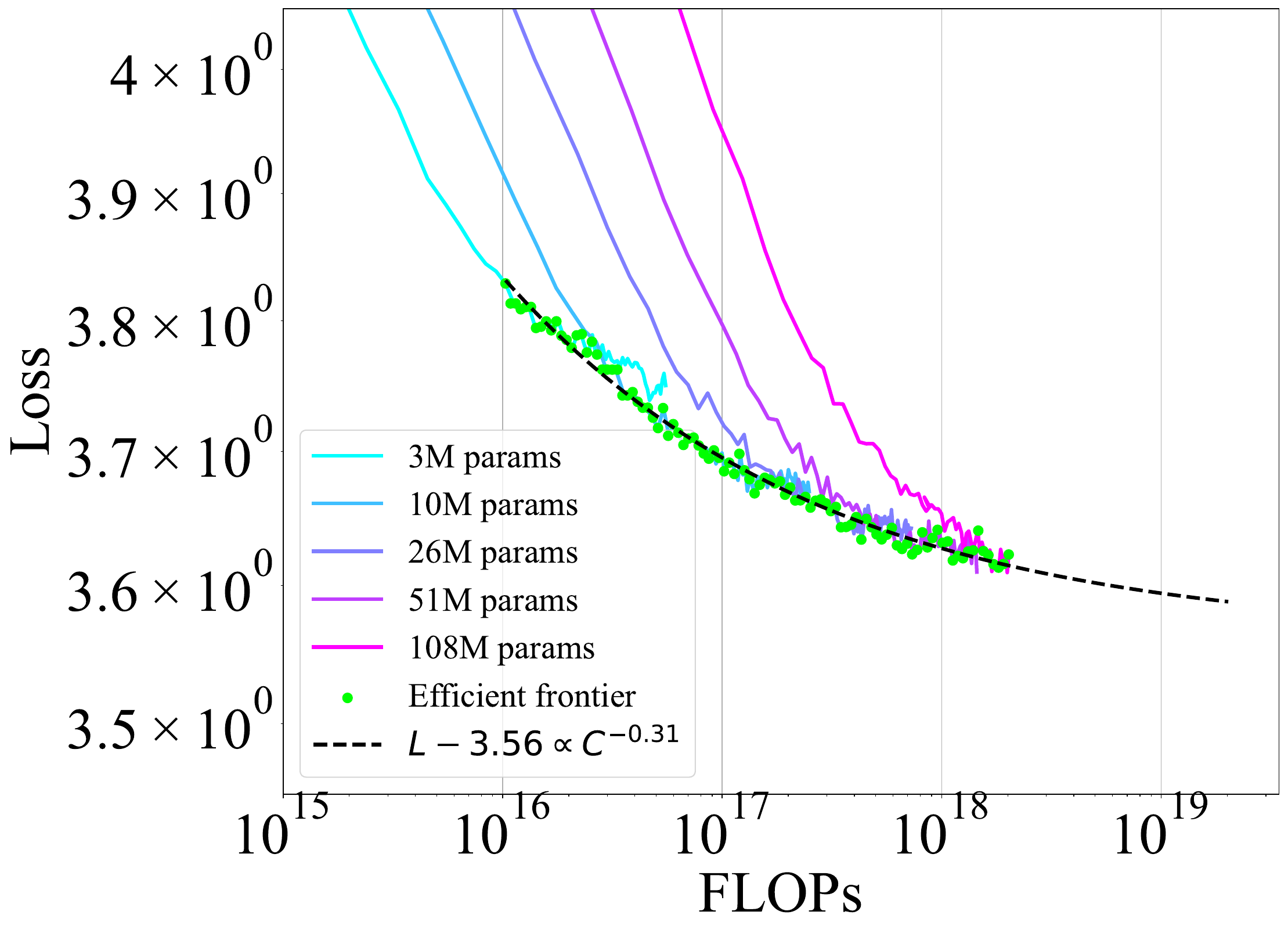}}
        \put(135,140){\small $N_\text{optimal} \propto C^{0.66}$}
        \put(135,120){\small $D_\text{optimal} \propto C^{0.34}$}
    \end{picture}
\end{subfigure}
\caption{This paper observes that scaling laws, as originally found in LLMs, also emerge in the tasks of world modeling and BC, when studying the pre-training loss on large datasets of human behavior. (a) (b) For world modeling, the power law coefficient determining optimal model size is affected by the compression rate of the tokenizer. (c) In BC with tokenized image observations (BC-Token), small models need a large FLOPs budget to saturate, making these scaling laws less clear cut. (d) However, moving to a single continuous embedding per observation remedies this (BC-CNN), producing prototypical scaling laws and a more balanced optimal model size coefficient.}
\label{fig_intro_overview}
\end{figure}
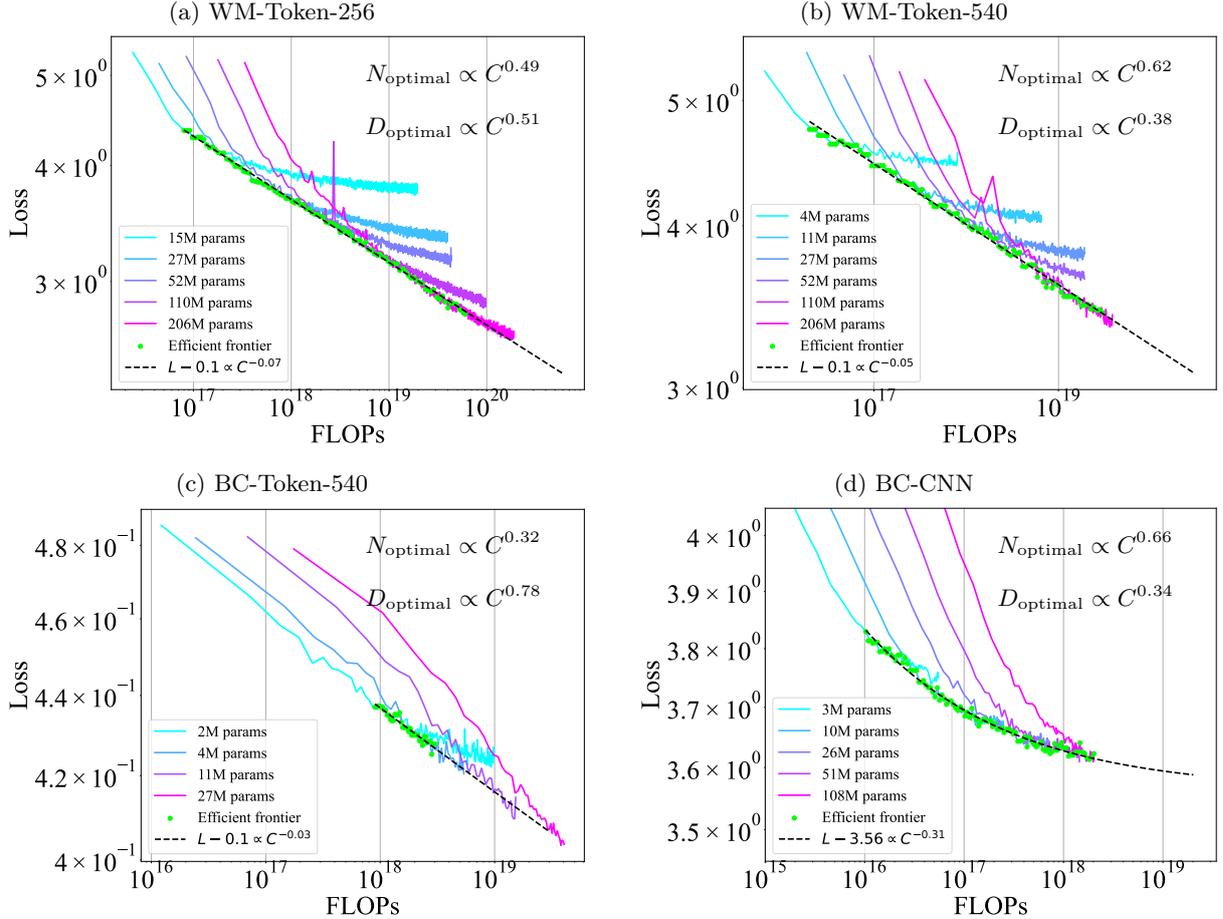

This paper helps close this gap. Similar to the study of scale in LLMs, we focus on the effect of scaling on a \textit{generative pre-training loss} (rather than on downstream agent performance, or reward- or representation-centric objectives), in the infinite data regime, on a fixed offline dataset. Under this setting, we train families of transformers on next-token prediction tasks using architectures popular in both world modeling and BC tasks.
This leads to several contributions, summarized in Figure \ref{fig_intro_overview}.

\begin{itemize}[left=4pt]
    \item Scaling laws similar to those in LLMs can be observed in world modeling with tokenized observations and actions (Section \ref{sec_mainresults_WM}, Figure \ref{fig_intro_overview}a).
    \item The optimal trade-off between model and dataset size in world modeling is influenced by the tokenizer's compression rate (number of tokens per observation) (Section \ref{sec_mainresults_WM}, Figure \ref{fig_intro_overview}a \& b).
    \item Scaling laws for BC with tokenized observations are harder to observe under modest compute budgets.
    The optimal trade-off favors smaller models and more data (Section \ref{sec_mainresults_BC}, Figure \ref{fig_intro_overview}c). 
    \item Scaling laws similar to those in LLMs can once again be observed in BC with one continuous encoding per observation (Section \ref{sec_mainresults_BC}, Figure \ref{fig_intro_overview}d).
    \item Our findings can be understood through small-scale language modeling experiments (Section \ref{sec_understanding}).
\end{itemize}

\textbf{Organization.} 
Section \ref{sec_related} provides detailed related work, contrasting the current understanding of scaling in embodied AI with other domains, and justifying pre-training loss as a proxy for online reward.
Section \ref{sec_setup} introduces details for our main experiments, including the architectures \& datasets considered, and details of scaling laws analyses.
Section \ref{sec_mainresults} presents our main results in world modeling and BC.
Section \ref{sec_understanding} presents insights behind our main results, including a set of tiny-scale language experiments mimicking aspects of our main experiments.
Section \ref{sec_discussion} discusses our findings and notes limitations.

\section{Background}
This section introduces related work, and outlines arguments and evidence supporting using pre-training loss to study scaling in embodied AI.

\subsection{Related work}
\label{sec_related}

\textbf{Scaling laws origin.} The term \textit{scaling laws} is used throughout the engineering and physical sciences to denote power law relationships between two quantities,
e.g. duration of a volcanic eruption and the probability of it continuing \citep{cannavo2016volcanic}.
The name derives from the \textit{scale-invariant}\footnote{For two variables $x$ \& $y$, the power law $y=ax^b$ is invariant to scaling $x$ by a constant $c$.\\ Formally: $a(cx)^b = c^b a x^b \implies c^b y = c^b a x^b \implies y=a x^b$.} property of power laws.
While early work suggested that power laws could be good empirical descriptors of pre-training loss in deep learning \citep{hestness2017deep, rosenfeld2019constructive}, it was \cite{kaplan2020scaling} who provided a comprehensive study of power laws in transformer LLMs, and popularized the usage of \textit{scaling laws} in this context. 

\textbf{Scaling laws in LLMs.}
As the real-world value of LLMs was understood,  
scaling in LLMs became a high-priority research topic. 
\cite{hoffmann2022training} conducted a precise analysis into the trade-off of model and dataset size, finding they should be increased in equal proportions. This conflicted with the suggestion that model size should be prioritized \cite{kaplan2020scaling} -- an incorrect conclusion that \cite{pearce2024reconciling} showed largely arose from counting only non-embedding parameters.

Many other aspects of LLM scaling analyses are beginning to be refined. \cite{su2024unraveling} revisited the methodology used to find scaling coefficients.
\cite{hagele2024scaling} found that multiple independent cosine schedules could be reproduced more efficiently through a constant learning rate with multiple short decays, or stochastic weight averaging.
\cite{pearce2024reconciling} \& \cite{porian2024resolving}  found that well-tuned constant learning rates were sufficient to recover certain coefficients.
\cite{bi2024deepseek} study the effect of various hyperparameters on scaling.
\cite{muennighoff2024repeatscaling} looked at repeated epochs, finding up to four epochs produce negligible departures from the infinite data regime. \cite{sardana2023beyondchinchilla} factored in inference compute to the definition of what is compute-optimal. \cite{isik2024downstream} study the link between pre-training loss and downstream performance.
A further line of research aims to explain \textit{why} power laws are such a good descriptor of empirical deep learning \citep{hutter2021learning, maloney2022solvable, bahri2024explaining}.

\textbf{Scaling laws in image and video generation.}
Scaling laws have also been observed in auto-regressive modeling of video and images \cite{henighan2020scaling, tian2024visual}. \cite{henighan2020scaling} found the optimal trade off between model and dataset size to match their reported LLM coefficient ($N_\text{optimal} \propto C^{0.7}$) and was not affected by tokenizer. Our experiments offer different findings in the domain of world modeling -- using updated methodologies to measure this trade-off \citep{hoffmann2022training}, we find it \textit{is} affected by the tokenizer. 

\textbf{Scaling in embodied AI.} Compared to LLMs, an understanding of scale in embodied settings is less advanced. Early successes in competitive games showed that running reinforcement learning (RL) at scale could surpass human performance \citep{silver2017mastering, berner2019dota}. 
In self-play RL, power laws were observed between certain quantities by \cite{neumann2022scaling}.
Meanwhile, \cite{2023singelagentscaling} noted that, in general, reward signals do not follow power laws, and defined a transformation of reward (intrinsic performance) that created self-consistent scaling laws.

Inspired by the effectiveness of scaling in LLMs, embodied AI research has recently begun to explore the effectiveness of generative pre-training objectives on offline datasets, when executed at scale.
This includes behavior cloning objectives in video games \citep{baker2022vpt, raad2024scaling}, robotics \citep{brohan2022rt1, brohan2023rt2, padalkar2023openx, bousmalis2023robocat}, or multiple domains \citep{reed2022gato}, as well as world modeling objectives \citep{hu2023gaia, yang2023unisim, bruce2024genie}. 
In these studies, the benefit of scale is generally shown through increasing model size on a specific downstream task of interest (e.g. measured by completion rate) -- an aggregated survey is provided by \cite{sartor2024neuralscaleembodied}.

\cite{tuyls2023scalingimitation} offer a valuable initial investigation into scaling laws for BC.
They fit power laws to both BC pre-training loss and online return, when scaling width of single-layer LSTM models on datasets generated by fixed high-reward policies. We extend this line of investigation by studying transformer models trained on datasets of human behavior, discovering effects of architecture choices on scaling coefficients. In addition we study scaling laws in world models for the first time.

\begin{figure}[t!]
\begin{center}
\hspace{0.2in} NetHack \hspace{0.85in} Bank Heist \hspace{0.85in} Breakout \hspace{0.75in} Space Invaders    \\
\includegraphics[width=0.245\columnwidth]{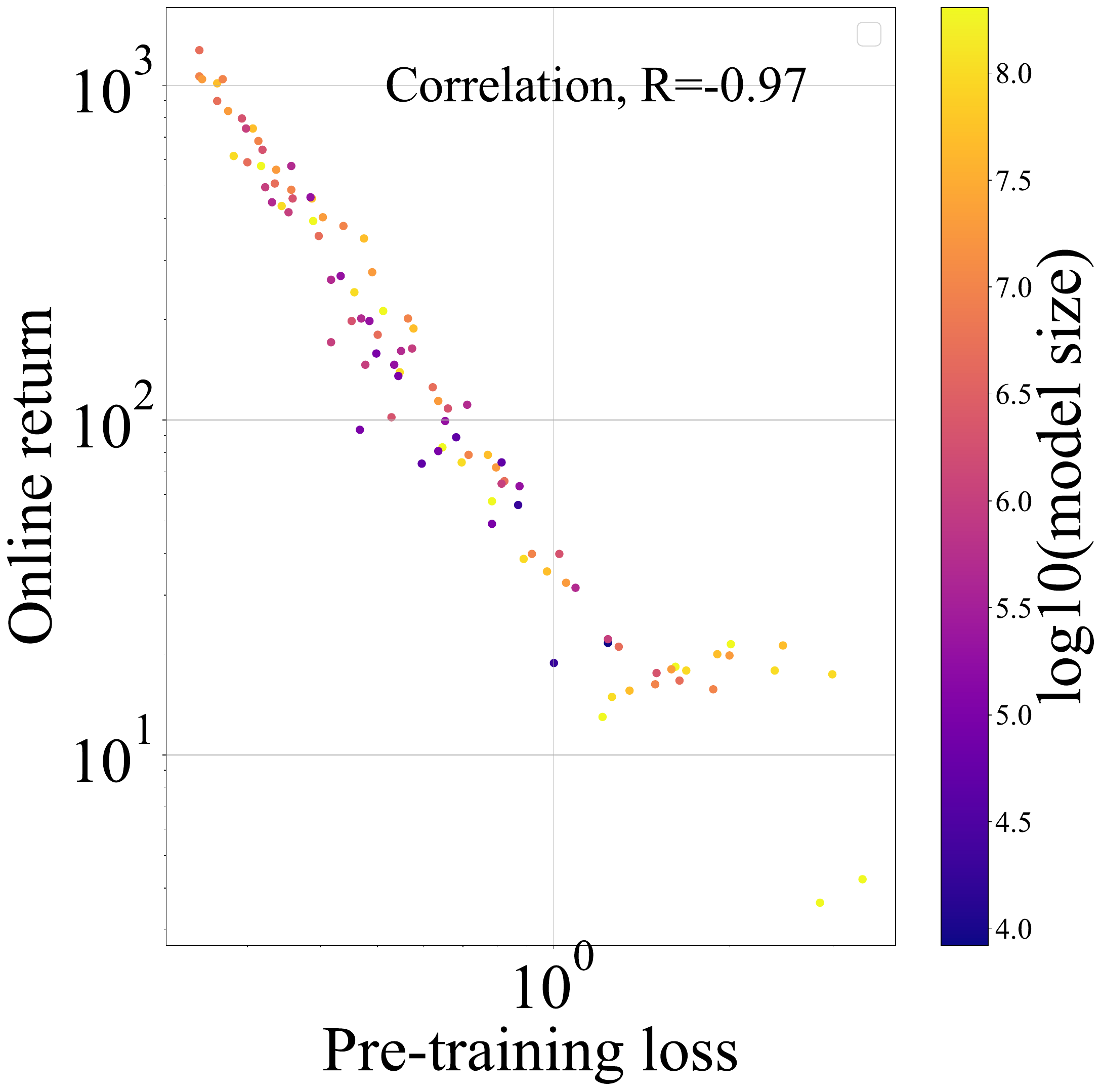}
\includegraphics[width=0.245\columnwidth]{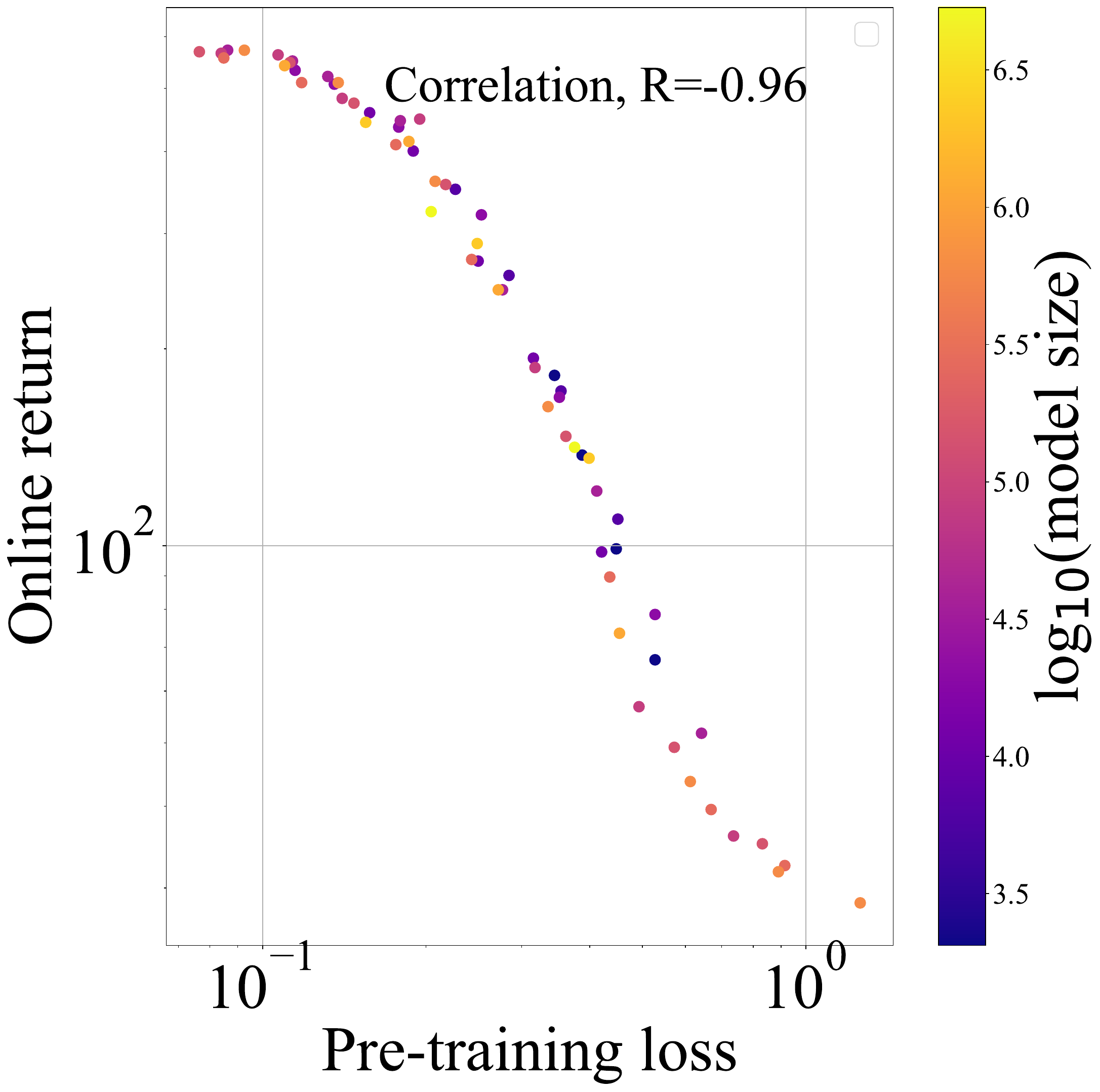}
\includegraphics[width=0.245\columnwidth]{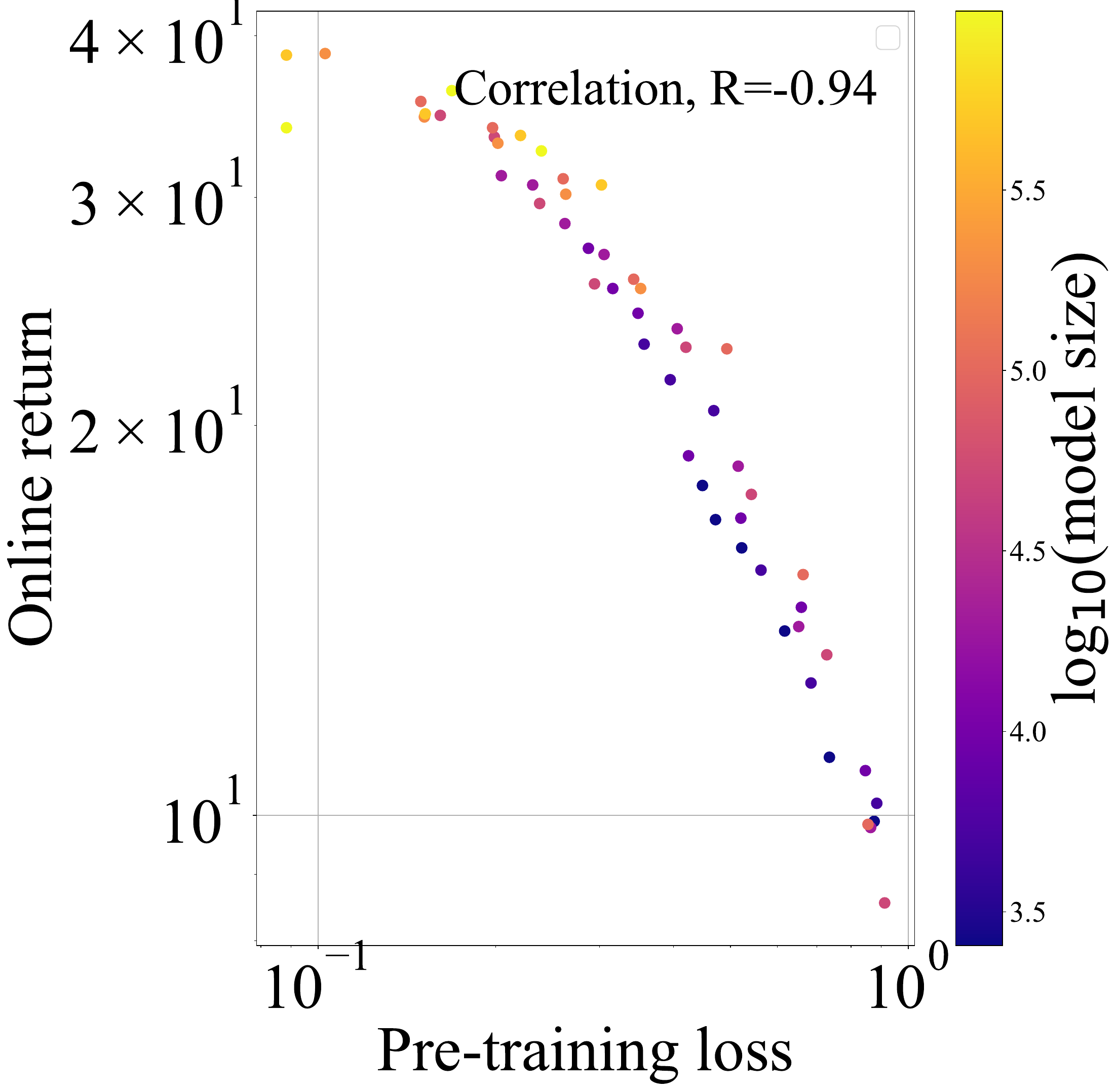}
\includegraphics[width=0.245\columnwidth]{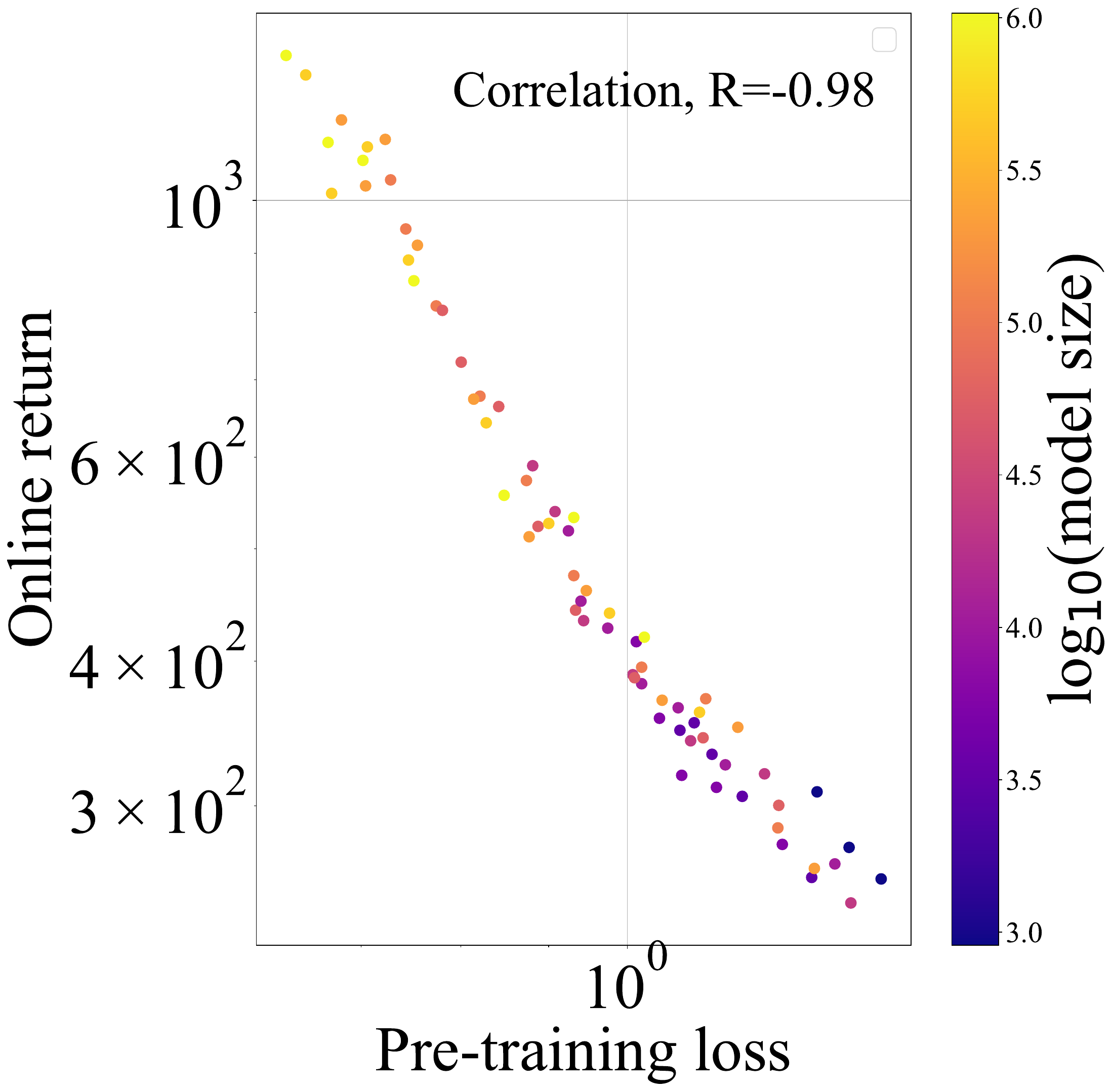}
\caption{Our meta-analysis of \cite{tuyls2023scalingimitation} evidences that pre-training loss is strongly correlated with reward in BC tasks when in the infinite data regime. 
}
\label{fig_tuyls_meta_main}
\end{center}
\end{figure}

\subsection{Pre-training loss as a proxy for performance}
\label{sec_pretraining_proxy}

A major difference between scaling research in LLMs and embodied AI is that \textit{LLM research uses pre-training loss} as the main variable of interest, while \textit{embodied AI has focused on downstream online task performance}. This handicaps embodied AI scaling research -- measuring online performance for a single model checkpoint in complex environments like robotics or modern video games is expensive in time and hardware, requiring multiple repeated runs to allow statistically significant comparisons. Furthermore, models may first require a period of fine-tuning before evaluation.
By contrast, pre-training loss is available for free at any point of a model's training.

We believe embodied AI's focus stems from reports that validation loss is only weakly correlated with online performance \cite{hussenot2021hyperparameter, li2024simpler}. However, such observations have been made with fixed-sized training datasets and held out validation sets, where effects of overfitting may be slightly beneficial. In contrast, scaling law studies are conducted in an \textbf{infinite data regime}, where datapoints are not trained on more than once, making train and test losses equivalent, and overfitting effects not applicable. 

\begin{figure}[t!]
\begin{center}
\begin{overpic}[width=0.49\columnwidth]{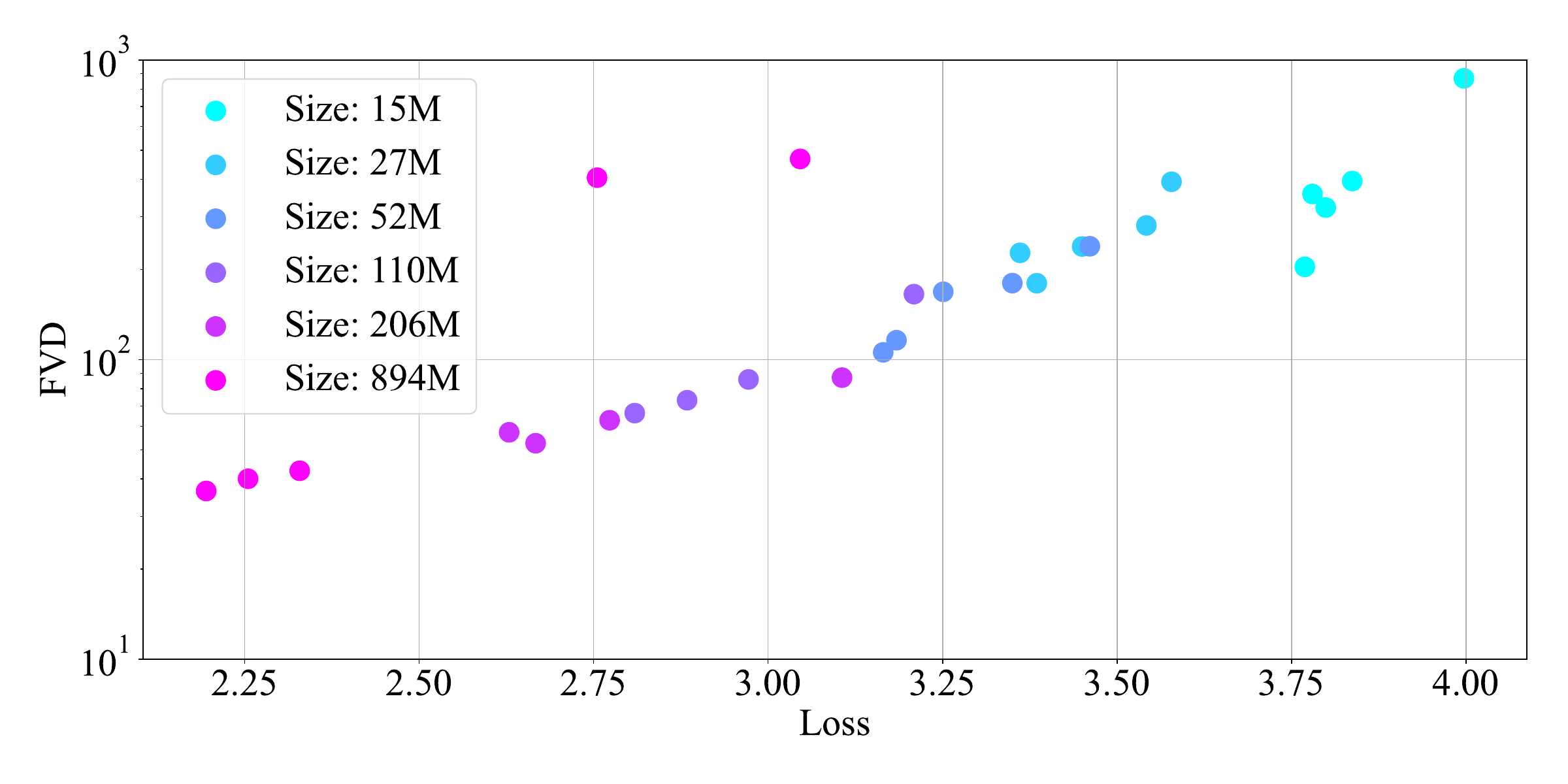}
    \put(55, 10){\footnotesize Correlation, $R=0.83$}
\end{overpic}
\begin{overpic}[width=0.49\columnwidth]{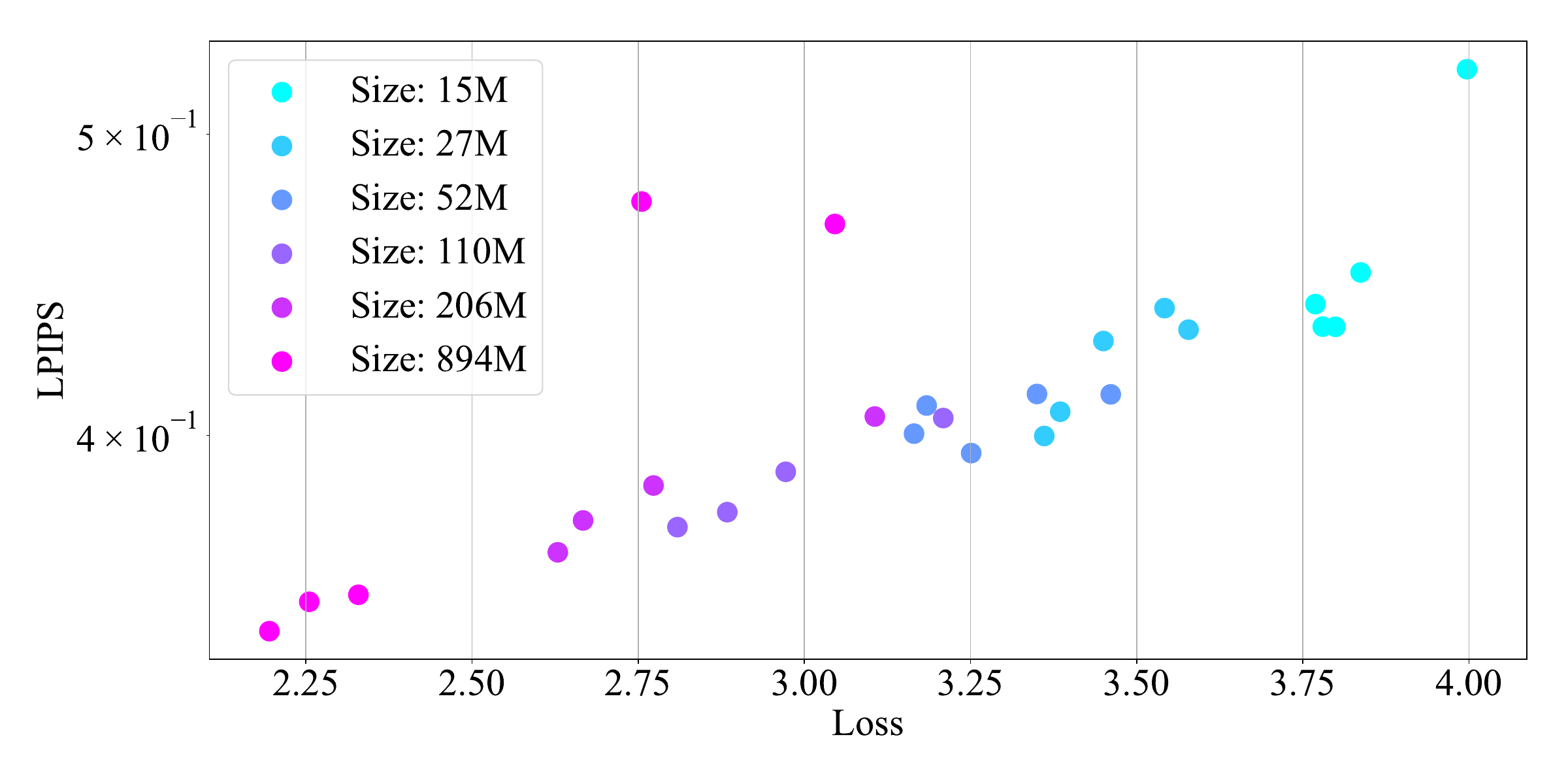}
    \put(55, 10){\footnotesize Correlation, $R=0.77$}
\end{overpic}
\caption{Our experiments suggest pre-training loss is a good proxy for world model quality. Further details in Appendix \ref{sec_app_pretrain_evidence}.}
\label{fig_wmtok256_loss_vs_metrics_main}
\end{center}
\end{figure}

To evidence that pre-training loss can be a good proxy for online return in the infinite data regime, we conducted a meta-analysis of \citep{tuyls2023scalingimitation}, who were able to roll out a large number of checkpoints for two reasons. 1) They used simple lightweight environments (Atari \& NetHack). 2) Their demonstration policy was high-skill removing any need for fine-tuning.
Figure \ref{fig_tuyls_meta_main} plots online environment return vs. pre-training loss for several environments (computed by tabulating pairs of points from Figure 6 \& 10 in \citep{tuyls2023scalingimitation}). The correlation coefficient for all environments is stronger that -0.94. Figure \ref{fig_wmtok256_loss_vs_metrics_main} shares evidence from our experiments that pre-training loss is well correlated with the video-generation quality of world models -- providing correlation coefficients around 0.8. Further details in Appendix \ref{sec_app_pretrain_evidence}.

More intuitively, improving a next-token prediction loss in BC and WM requires models to `know more' about behaviors and the environment, creating more useful pre-trained checkpoints for specialization to downstream tasks.
In BC, better predicting the next action in a dataset of human behavior requires understanding the objectives human's are trying to complete, alternative social behaviors they might choose to perform, as well as making in-context inferences about the skill level and mental state of individuals.
In WM, decreasing next-token prediction loss might follow a curriculum, first requiring a model to capture basic shapes and colors, then textures and physics, followed by rare object interactions, and finally even complex stochastic elements in the environment such as other intelligent agents.

\section{Methodology}
\label{sec_setup}

This section provides details for our main experiments. We describe the pre-training tasks, architectures, and datasets considered. We also detail the methodology used in the scaling law analyses.

\subsection{Tasks}
\label{sec_setup_tasks}

We consider trajectories constructed as sequences of alternating observations $\mathbf{o}_t$ and actions $\mathbf{a}_t$ for timestep $t \in \mathbb{N}$. In this work, observations are always images, $\mathbf{o}_t \in \mathbb{R}^{3 \times w \times h}$ and any continuous actions are discretized during preprocessing leaving, $\mathbf{a}_t \in \{0,1\}^{d_a}$.

Given this data format, we consider two tasks. \textit{World modeling} (WM) \citep{ha2018world} predicts future observations from previous observations and actions. This allows an agent to explicitly understand how its environment works, which can be used for planning, or dyna-style RL \citep{sutton2018rltextbook}. \textit{Behavior cloning} (BC) predicts the future actions that the dataset's demonstrators take \citep{bakker1996robot}. This creates a policy that can be directly used to act in the environment, either as-is or following further fine-tuning.
Concretely, these two tasks require modeling the following quantities,
\begin{align}
    \text{World modeling:}& \;\; P(\mathbf{o}_{t+1} | \mathbf{o}_{t} \dots \mathbf{o}_{t-k}, \mathbf{a}_{t} \dots \mathbf{a}_{t-k}), \\
    \text{Behavior cloning:}& \;\; P(\mathbf{a}_{t} | \mathbf{o}_{t} \dots \mathbf{o}_{t-k}, \mathbf{a}_{t-1} \dots \mathbf{a}_{t-k-1}).
\end{align}
This work focuses on \textit{generative} pre-training aiming to model this full conditional probability distribution. We leave a study of scaling laws for alternative objectives, e.g., explicitly targeting representation learning \citep{nair2022r3m} or reward-centric models \citep{hafner2020mastering}, to future work.

\subsection{Architectures}
\label{sec_setup_architectures}

\begin{figure}[t!]
\begin{center}
\includegraphics[width=0.999\columnwidth]{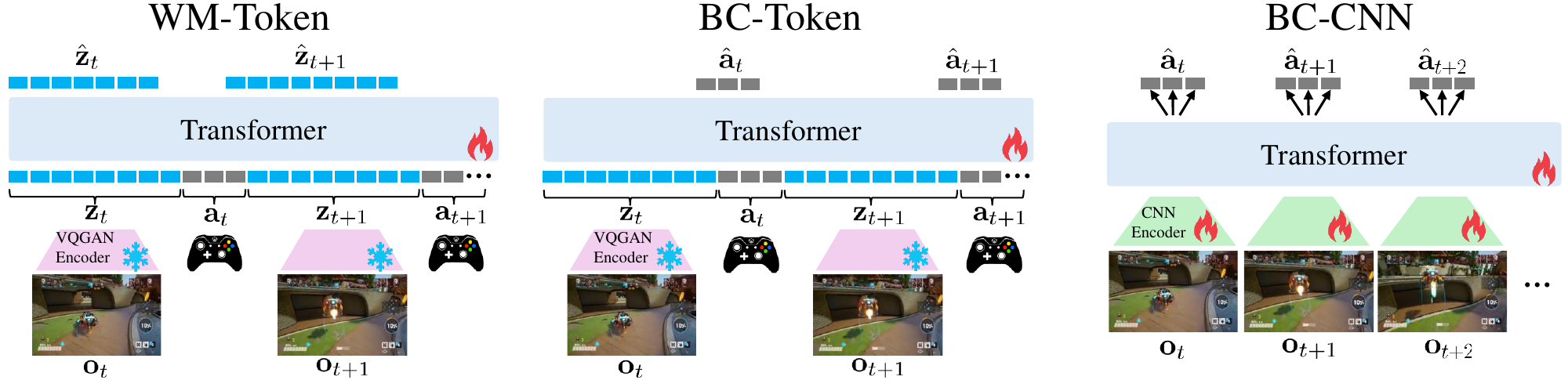}
\caption{The World Modelling (WM) and Behavior Cloning (BC) tasks \& architecture combinations considered in this work. The fire symbol signifies trainable components, the ice symbol signifies frozen pre-trained components.}
\label{fig_architectures_tasks}
\end{center}
\end{figure}

All experiments revolve around GPT-2 style causal transformers \citep{radford2019gpt2} as the core of the model. However we consider two different methods for inputting image observations, summarized in Figure \ref{fig_architectures_tasks}. Section \ref{sec_setup_scaling} details how we measure the model size of each.

\textbf{Tokenized architecture.} Our first architecture tokenizes each image observation into multiple discrete tokens. This is done with a frozen VQGAN encoder $\operatorname{Enc}_\theta(\mathbf{o}_t) \to \mathbf{z}_t$, where $\mathbf{z}_t \in \{1,2,...,V_o\}^{d_z}$, for vocabulary size $V_o$ and latent dimension $d_z$.
Discretized actions are mapped to a non-overlapping vocabulary.
Following tokenization, training sequences take the form, 
\begin{equation}
[z_t^1, z_t^2, ... ,  z_t^{d_z}, a_t^1, a_t^2, ... , a_t^{d_a}, z_{t+1}^1, z_{t+1}^2, ... z_{t+1}^{d_z}, a_{t+1}^1, a_{t+1}^2, ... , a_{t+1}^{d_a}, ...],
\end{equation}
where each item of the sequence is an integer within our vocabulary. A transformer is then trained to maximize the likelihood of either the latent image tokens (world modeling), or action tokens (BC).

This tokenized architecture is widely used both in world modeling \citep{micheli2022iris} and BC tasks \citep{bousmalis2023robocat}. 
Gato \citep{reed2022gato} used a similar design but with continuous patches rather than discrete tokens. 
Our implementation tests both a `small' (28M parameters, $d_z=256$) and `large' (150M parameters, $d_z=540$) VQGAN -- further details in Appendix \ref{sec_app_main_exp}.

\textbf{CNN architecture.} 
Our second architecture differs in two ways. 1) Each image observation is input into the transformer as a single continuous embedding, extracted from a small trainable convolutional neural network (CNN). 2) Action dimensions are predicted independently (rather than in series), assuming $P(\mathbf{a}_{t}| \dots) \approx \prod_{i=1}^{d_a}  P(a_{t}^i | \dots)$. A single forward pass of the transformer is needed per action prediction.

This produces an architecture similar to \cite{baker2022vpt} (VPT additionally used a transformer-XL and a refined hierarchical action space).
Our implementation uses an Impala-style \citep{espeholt2018impala} CNN with 0.6M parameters for embedding image observations.

\subsection{Datasets}
\label{sec_setup_datasets}

This paper focuses on the effect of scaling on the pre-training loss over an offline dataset. To study this cleanly, datasets must meet two criteria. 
\begin{enumerate}
    \item \textbf{Dataset size.} Repeated training on the same data alters the effect of scaling -- datasets should be large enough that training is done in the infinite data regime.
    \item \textbf{Dataset diversity.} The behavior and environment must contain enough richness and variety that pre-training loss does not saturate across model sizes tested. 
\end{enumerate}

Many existing benchmark datasets fail to fulfill these criteria -- if not due to limited size, then because behavior is generated from a pre-trained fixed policy, or the environment is too simple.

\begin{figure}[t!]
\begin{center}
\includegraphics[width=0.99\columnwidth]{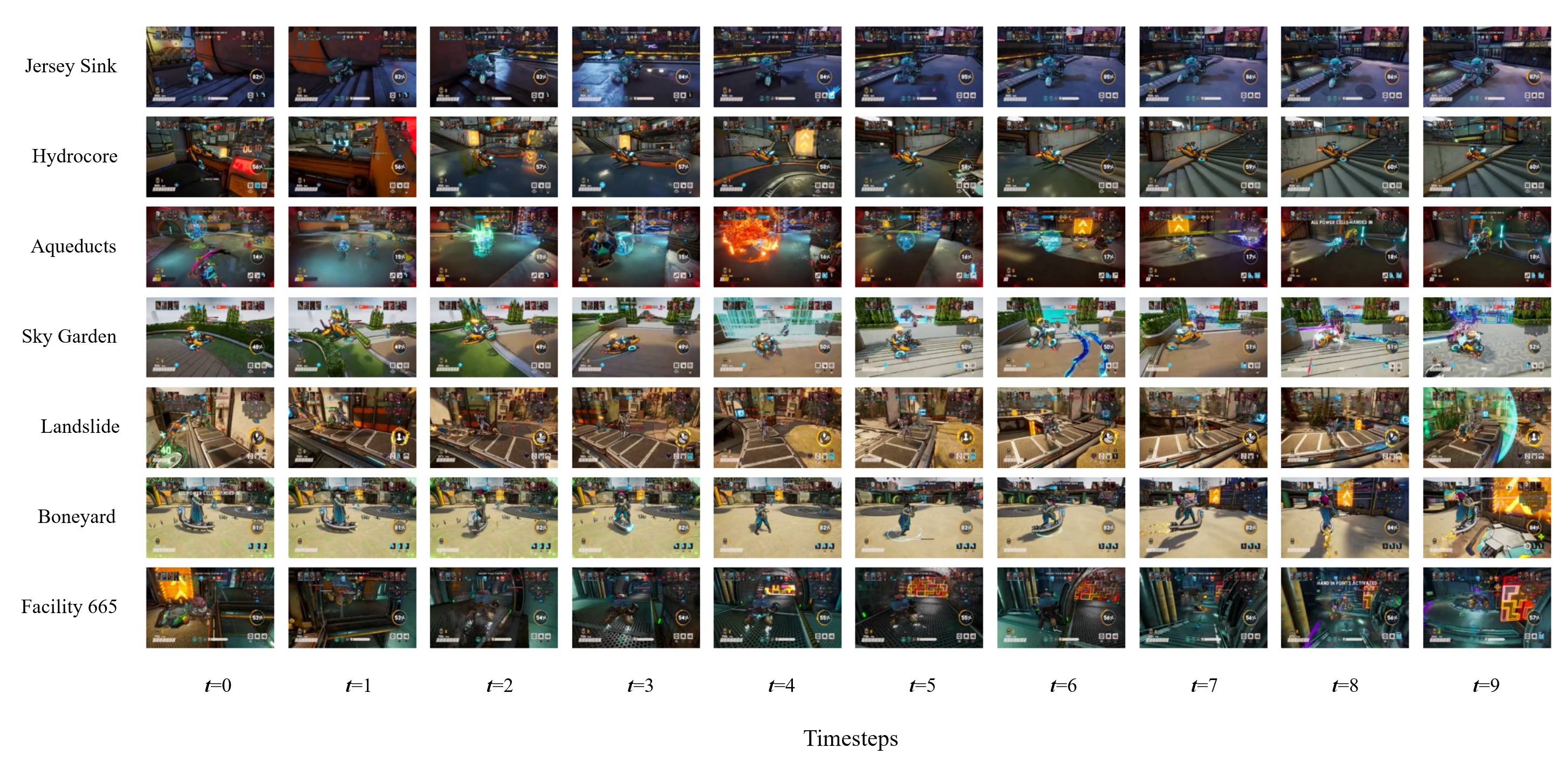}
\caption{Example trajectories from a dataset of 8.6 years of human gameplay in the video game \textit{Bleeding Edge} across 7 maps.}
\label{fig_dataset_egs}
\end{center}
\end{figure}

Our work primarily focuses on a dataset of human behavior collected in a video game \textit{Bleeding Edge}. 
This is a fast-paced 4 vs 4 multiplayer  game, with a range of characters, abilities and maps. 
Game play is highly complex due to the cooperative and competitive dynamics. Success requires selecting high-level strategies (e.g. choosing which map regions to fight for), as well as fine-grained reactive control during combat.
Figure \ref{fig_dataset_egs} shows example sequences from our dataset.

Supported by the game's developer \textit{Ninja Theory}, we compiled a dataset of 8.6 years of anonymized game play, containing both image observations and controller actions. We refer to this as the \textit{7 map dataset}. We also use a subset of this for some experiments, of around 1.1 years from a single map, which we name the \textit{Sky Garden dataset}. Appendix \ref{subsec:data-details} provides further details.

As a secondary dataset we use RT-1 \citep{brohan2022rt1}, comprising 14 days of human's operating a robotic arm doing a range of manipulation tasks such as `pick banana from white bowl'.
Using this smaller dataset allows us both to verify that conclusions on the large-scale video games dataset hold in a real-world robotics domain, and also allows us to run several small scale ablations in WM. Appendix \ref{sec_app_robotics} provides further details about the dataset and tokenizers used.

\subsection{Scaling analysis methodology}
\label{sec_setup_scaling}

We study the relationship between several quantities defined below.
\begin{itemize}
    \item Model size $N$, the \textit{total} number of trainable parameters (ignoring VQGAN parameters for WM-Token \& BC-Token, but including the fixed-size CNN for BC-CNN). Embedding parameters are included in the count following \cite{pearce2024reconciling}.
    \item Dataset size $D$, the total number of \textit{inputs} the transformer sees during training. For WM-Token and BC-Token this is $d_z + d_a$ per observation \& action pair, and for BC-CNN this is one per observation \& action pair.
    \item Compute $C$, the number of floating point operations (FLOPs) used during training. The common approximation of $C=6ND$ \citep{kaplan2020scaling} is used.
    \item Loss $L$, the standard classification cross-entropy loss (all targets are discretized). 
    We assume training loss is an accurate proxy for test loss (Appendix \ref{sec_appendix_infinite} analyzes further).
\end{itemize}


More specifically, we are interested in `compute-optimal' versions of each quantity. For loss, this is defined as the minimal loss possible for a given FLOPs budget,
\begin{align}
    L_\text{optimal}(C) &= \underset{\text{s.t.}\; C=6N D}{\operatorname{min}} L(N,D),
\end{align}
where $L(N,D)$ is the empirical loss achieved with an $N$ parameter model trained on $D$ tokens. We further define optimal model and dataset sizes as the configuration that produce this minimal loss given a FLOPs budget,
\begin{align}
    N_\text{optimal}(C), D_\text{optimal}(C) = \underset{N, D \; \text{s.t.}\; C=6N D}{\operatorname{argmin}} L(N,D).
\end{align}

\textbf{Scaling analysis.} 
The heart of scaling law analysis is fitting power law relationships predicting these compute-optimal quantities. For predicting optimal model and dataset size, we use,
\begin{align}
    \hat{N}_\text{optimal}(C) &= a_0 C ^{a} \;\;\;\;\;\; 
    \hat{D}_\text{optimal}(C) = b_0 C ^{b},
\end{align}
with fitted constants $a_0, a, b_0, b$.\footnote{Note that by subscribing to $C=6ND$ we find $a=1-b$; $N \propto C^a \implies C/D \propto C^a \implies D \propto C^{1-a}$. Hence, at times we only describe relationships in terms of $N \propto C^a$, with $N \propto D^{1-a}$ implied.}
We consider two methods to fit these relationships, introduced by \cite{hoffmann2022training}. Their Method 1, which we term \textit{Frontier fit}, classifies efficient models as those falling on the \textit{efficient frontier} (see Figure \ref{fig_intro_overview}). Coefficients can then be estimated straightforwardly through a line of best fit on a plot of FLOPs vs parameters or data for these efficient models.

Frontier fit is our preferred method when available -- it avoids making any assumptions about the training curves, directly fitting the best models observed. 
However, it requires training models past the point where they are the optimal configuration (seen on a loss-FLOPs plot as overlapping curves). In some of our experiments (BC-Token and Section \ref{sec_understand_q1}), this was not possible.

In these situations, we resort to Method 3 of \cite{hoffmann2022training}, which we term \textit{Parametric fit}. This fits the coefficients $\alpha, \beta, N_c, D_c, E$ to a parametric loss form,
\begin{align}
    \hat{L}(N,D) &= \frac{N_c}{N^{\alpha}}+ \frac{D_c}{D^{\beta}} + E \label{eq_loss_ND} ,
\end{align}
to the empirical training curves.
In our implementation, we use SciPy's \verb|curve_fit| function. We find $a = \beta/(\alpha+\beta), b = \alpha/(\alpha+\beta)$. This makes a very strong assumption about the training curves, but allows coefficients to be estimated at a smaller compute budget.

For loss prediction we use the form recommended by \cite{pearce2024reconciling},
\begin{align}
    \hat{L}_\text{optimal}(N,D) &= c_0 C^{-c} + E.
\end{align}
We again use the \verb|curve_fit| function, fitted to models along the efficient frontier. During fitting, we set bounds on the variables: $c_0 \in [0, \infty], c \in [-1,1], E \in [0.1, \infty]$.

\textbf{Training details.} While early scaling studies conducted sweeps over multiple cosine decays of differing lengths \citep{kaplan2020scaling, hoffmann2022training}, follow up work found this redundant \citep{pearce2024reconciling, hagele2024scaling, porian2024resolving}. We follow the approach of using a constant learning rate per model, so each requires only one training run. 
We aim to train models until they have passed their compute efficient FLOPs budget. 
We only modify the parameters of the transformer, following the configurations documented in Appendix \ref{sec_app_main_exp}.

\section{Scaling analysis in embodied AI}
\label{sec_mainresults}

This section presents our main results. 
We begin by considering the scaling laws for the task of world modelling in Section \ref{sec_mainresults_WM} with two different tokenizers (turning image observations into $256$ and $540$ tokens for the \textit{small} and \textit{large} variants respectively). 
Section \ref{sec_mainresults_BC} then considers the task of BC both with \textit{tokenized} and \textit{CNN} architectures.
Finally, Section \ref{sec_mainresults_extrap_wm} tests the extrapolation capability of these scaling laws for the task of world modeling.

\begin{table}[t!]
\caption{Summary of fitted scaling coefficients for our main experiments. Note that we favor the Frontier fit when available, and only use the Parametric fit for BC-Token-540 (see Section \ref{sec_setup_scaling}).}
\begin{center}
\label{tbl_main_coeffs}
\resizebox{0.9 \columnwidth}{!}{
\begin{tabular}{lrrrr}
     \multicolumn{1}{c}{} & \multicolumn{2}{c}{\bf Frontier fit}& \multicolumn{2}{c}{\bf Parametric fit}\\
     \multicolumn{1}{c}{\bf Experiment}  
     & \multicolumn{1}{c}{\bf $N_\text{optimal} \propto C^a$}
     & \multicolumn{1}{c}{\bf $D_\text{optimal} \propto C^b$}
     & \multicolumn{1}{c}{\bf $N_\text{optimal} \propto C^a$}
     & \multicolumn{1}{c}{\bf $D_\text{optimal} \propto C^b$}
     \\ \hline \\
    WM-Token-256  & 0.49 & 0.51 & 0.52 & 0.48 \\
    WM-Token-540  & 0.62 & 0.37 & 0.78 & 0.22 \\
    BC-Token-540  & N/A & N/A & 0.32 & 0.68 \\
    BC-CNN & 0.66 & 0.34 & 0.47 & 0.53
\end{tabular}
}
\end{center}
\end{table}

\subsection{Scaling analysis in world modeling}
\label{sec_mainresults_WM}

\begin{figure}[h!]
\begin{center}
\includegraphics[width=0.32\columnwidth]{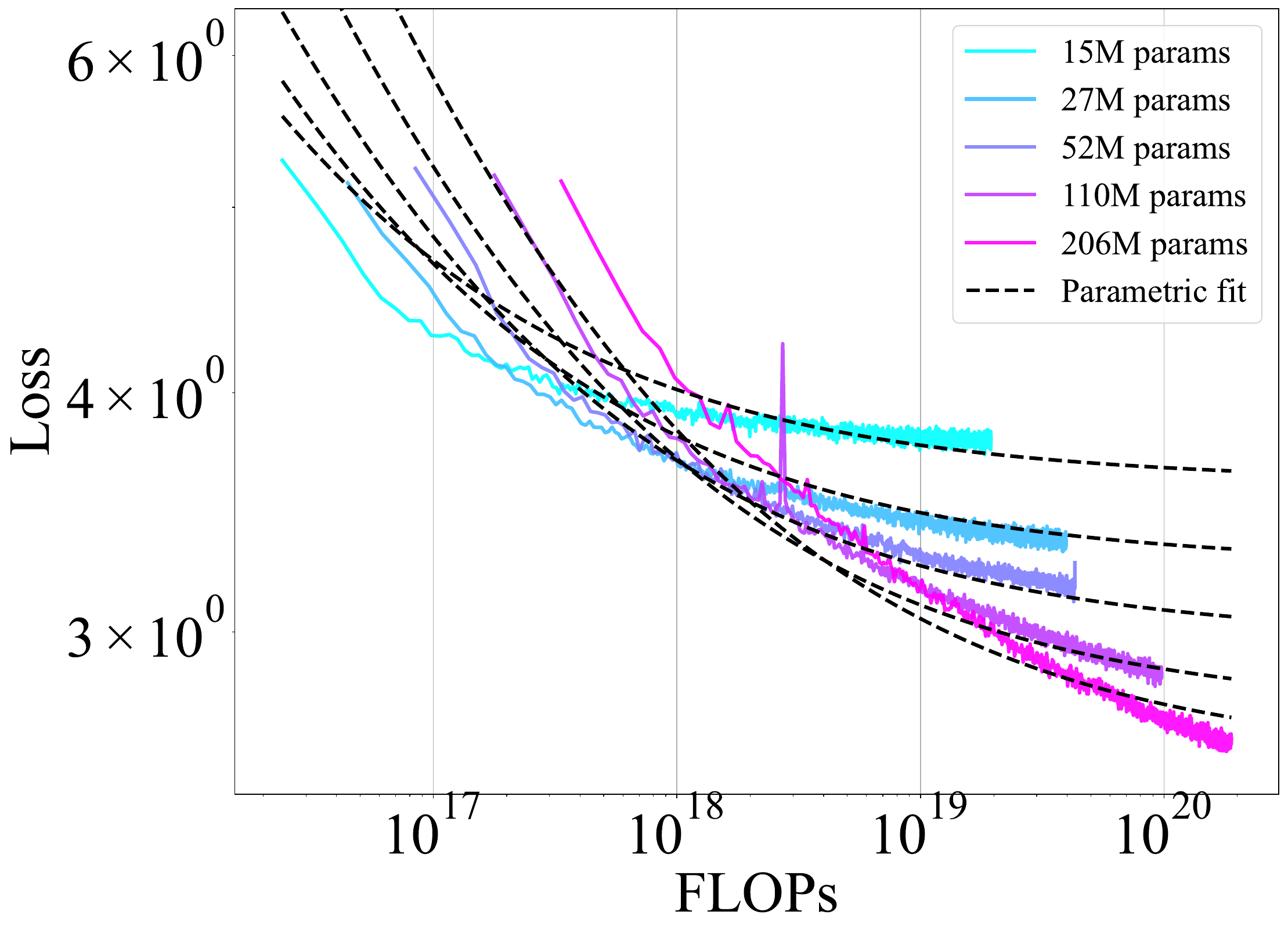}
\includegraphics[width=0.32\columnwidth]{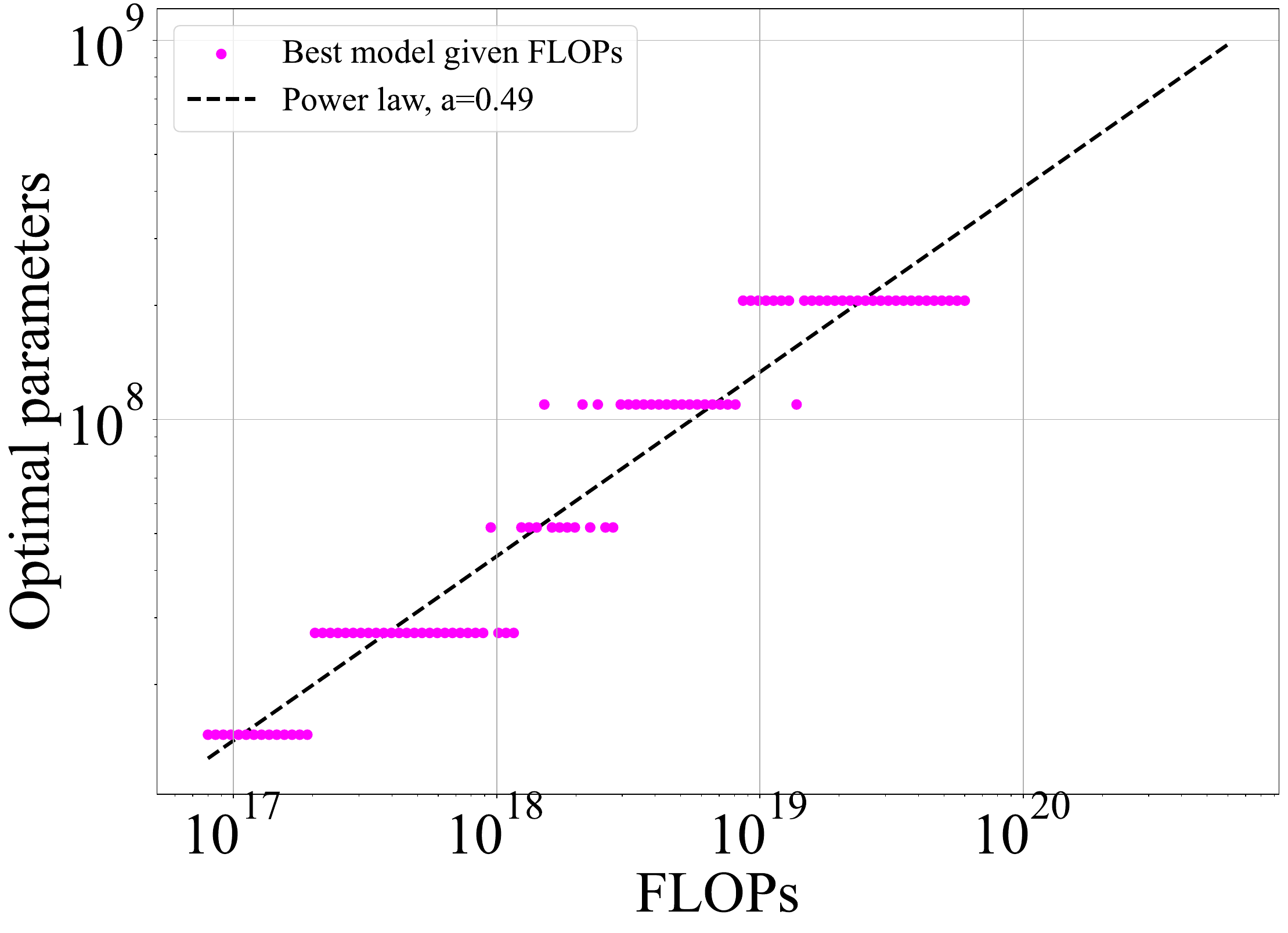}
\includegraphics[width=0.32\columnwidth]{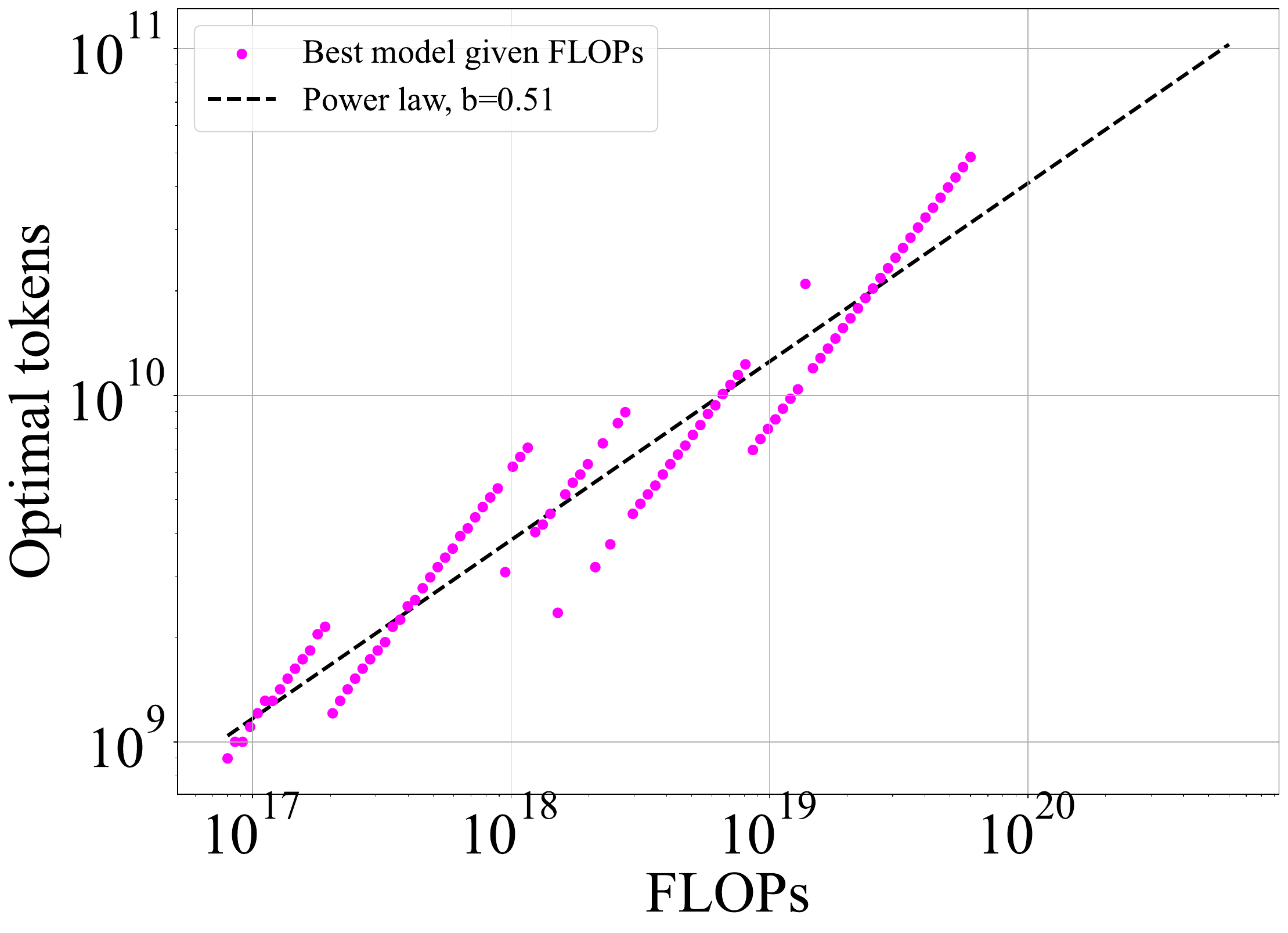}
\caption{WM-Token scaling with $d_z=$256 tokens per image observation. Left shows the \textit{parametric fit}. Middle \& right show the \textit{frontier fit} estimating optimal model \& dataset size respectively.}
\label{fig_WM_256_tok}
\end{center}
\end{figure}

\begin{figure}[h!]
\begin{center}
\includegraphics[width=0.32\columnwidth]{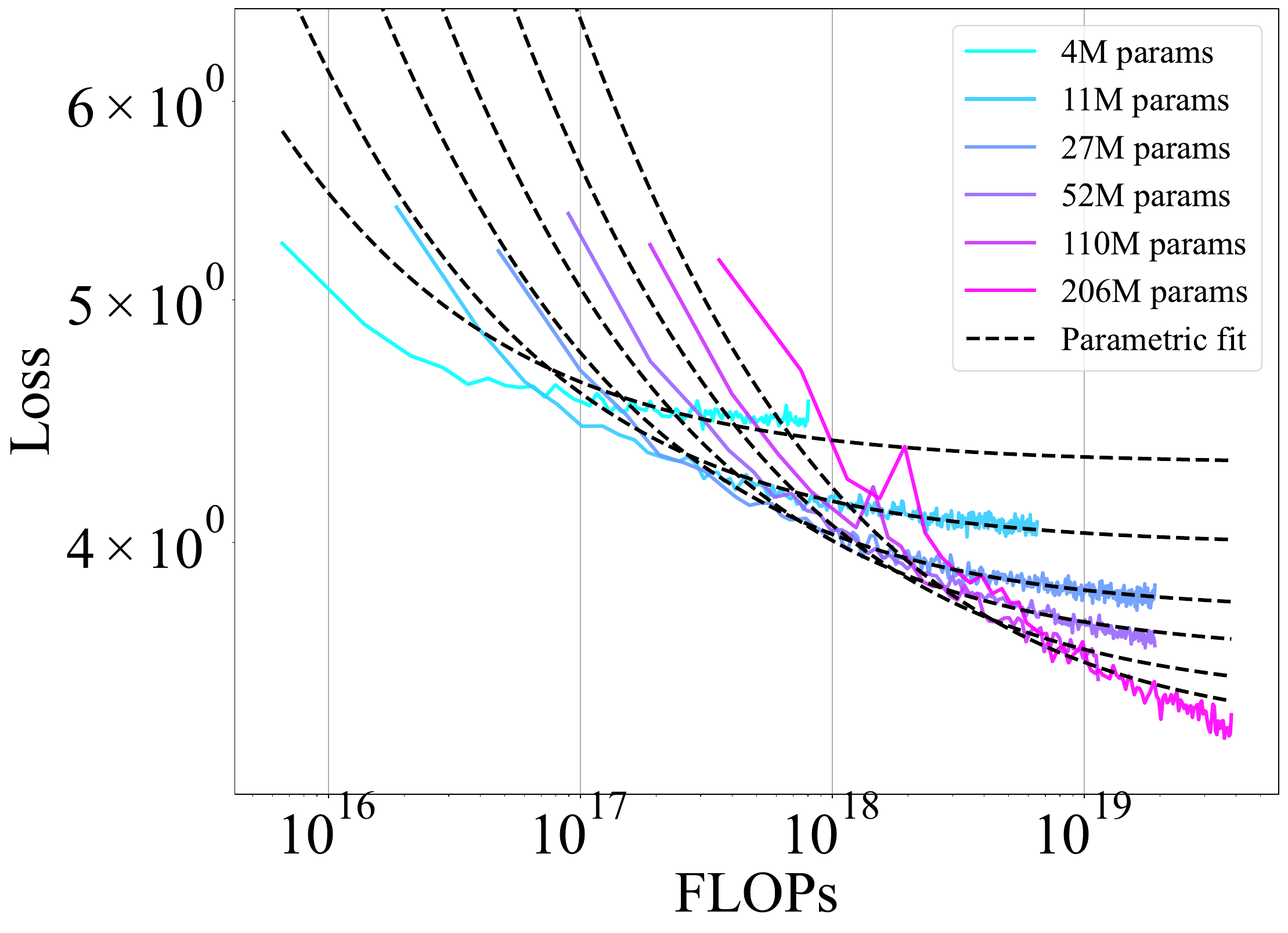}
\includegraphics[width=0.32\columnwidth]{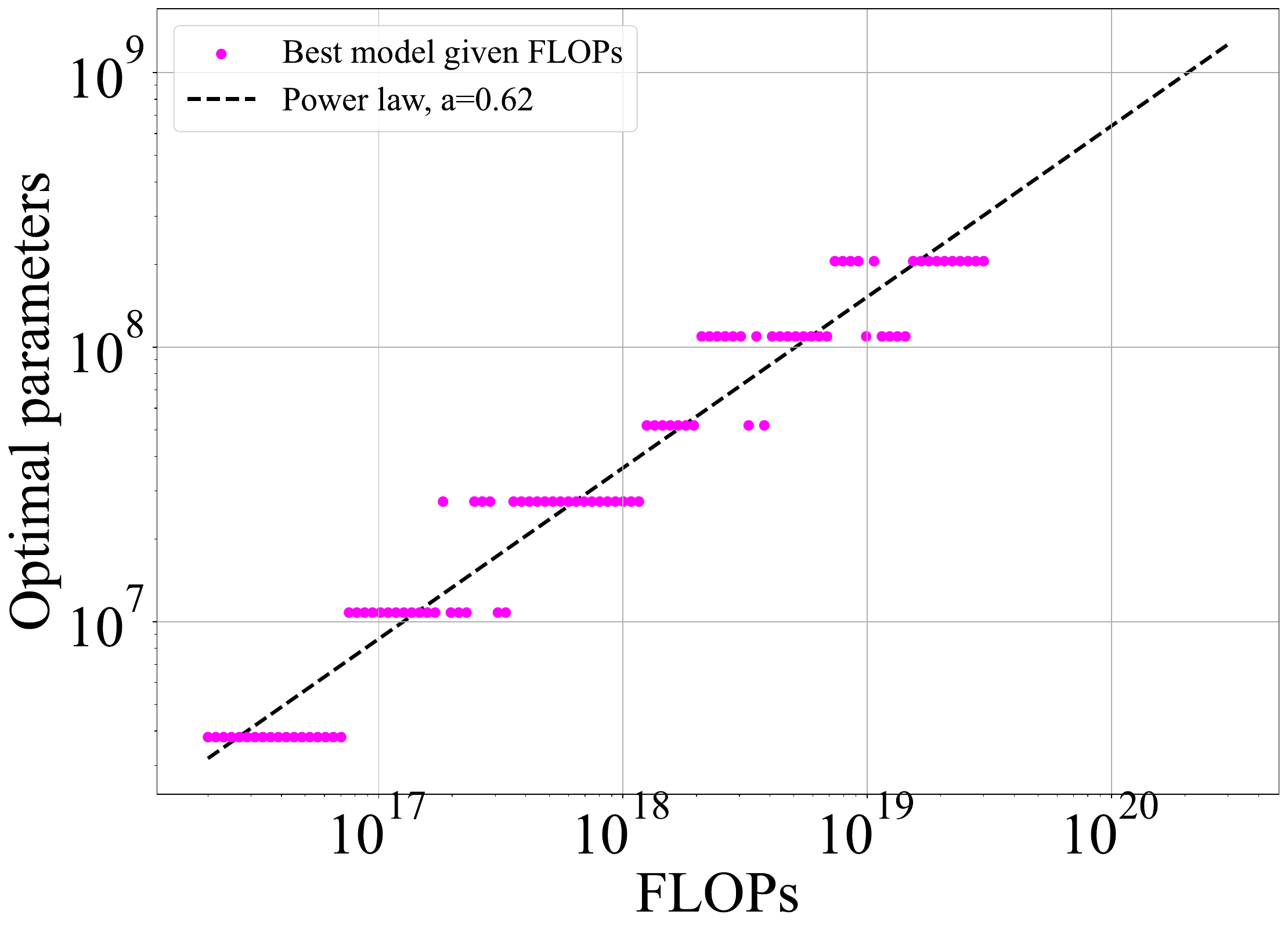}
\includegraphics[width=0.32\columnwidth]{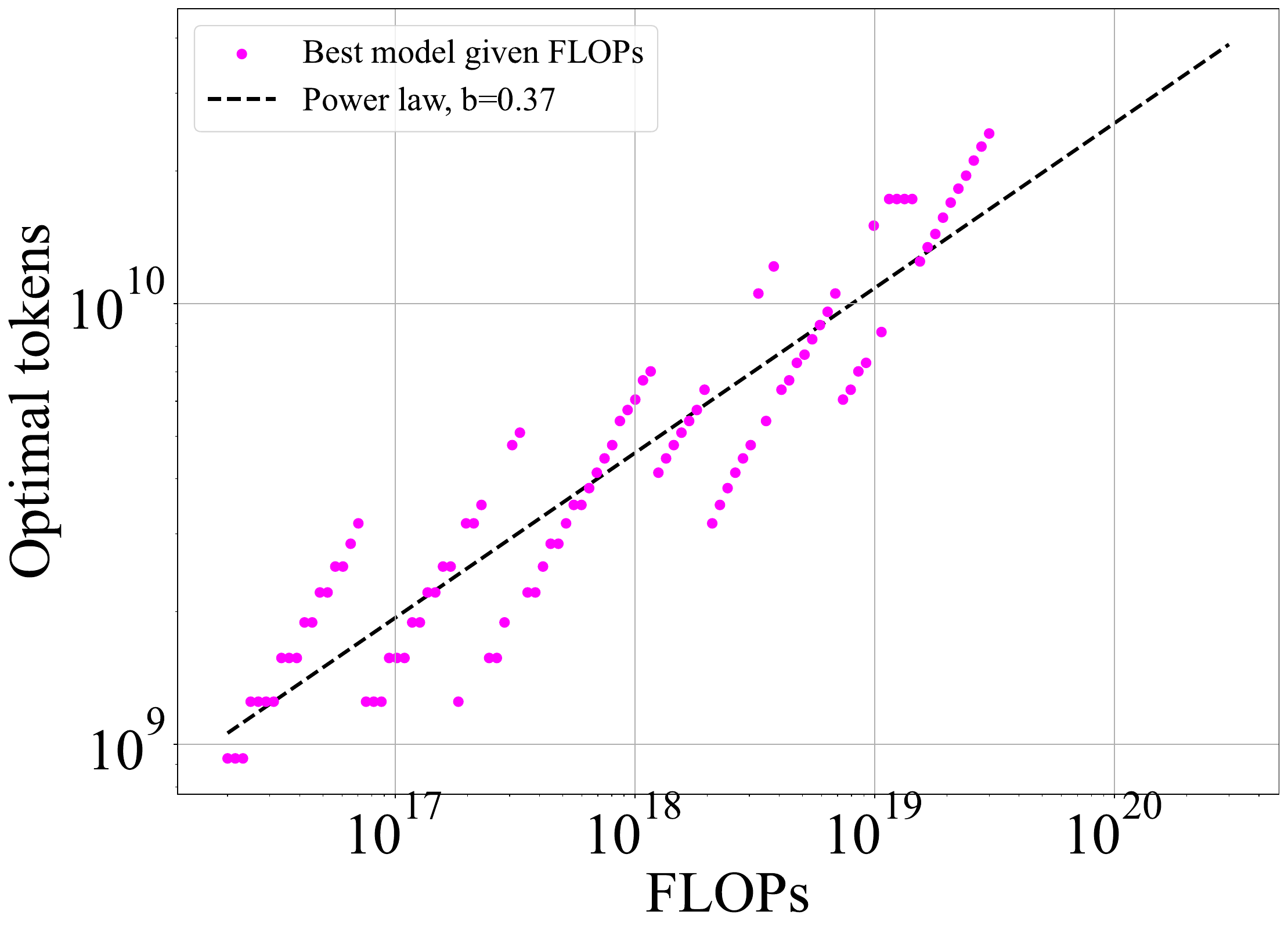}
\caption{WM-Token scaling with $d_z=$540 tokens per image observation.
 Left shows the \textit{parametric fit}. Middle \& right show the \textit{frontier fit} estimating optimal model \& dataset size respectively.
 Compared to the results for WM-Token-256, the power law coefficient for $N_\text{optimal}$ increases from $0.49$ to $0.62$.
}
\label{fig_WM_540_tok}
\end{center}
\end{figure}

Figures \ref{fig_WM_256_tok} \& \ref{fig_WM_540_tok} present our results for the task of world modeling, with the scaling law coefficients summarized in Table \ref{tbl_main_coeffs}.
For WM-Token-256 we find that the optimal coefficients for model and dataset size are both $\approx 0.5$, e.g. one should increase both model and dataset size in the same proportions.
This matches the scaling laws observed in LLMs \citep{hoffmann2022training}.
Increasing the number of tokens per image to $540$ for WM-Token-540 changes the optimal trade-off between model and dataset size, skewing towards model size; $N_\text{optimal}=0.62$, $D_\text{optimal}=0.37$. We discuss this further in Section \ref{sec_understand_q3}.

Appendix Figure \ref{fig_rt1_scaling} visualizes the power law fits for the RT-1 robotics dataset, confirming that this predictable scaling behavior is not specific to human behavior in video games, and also emerges on real-world robotics tasks with high-skill human operators.

\subsection{Scaling analysis in behavior cloning}
\label{sec_mainresults_BC}

\begin{figure}[h!]
\begin{center}
\includegraphics[width=0.32\columnwidth]{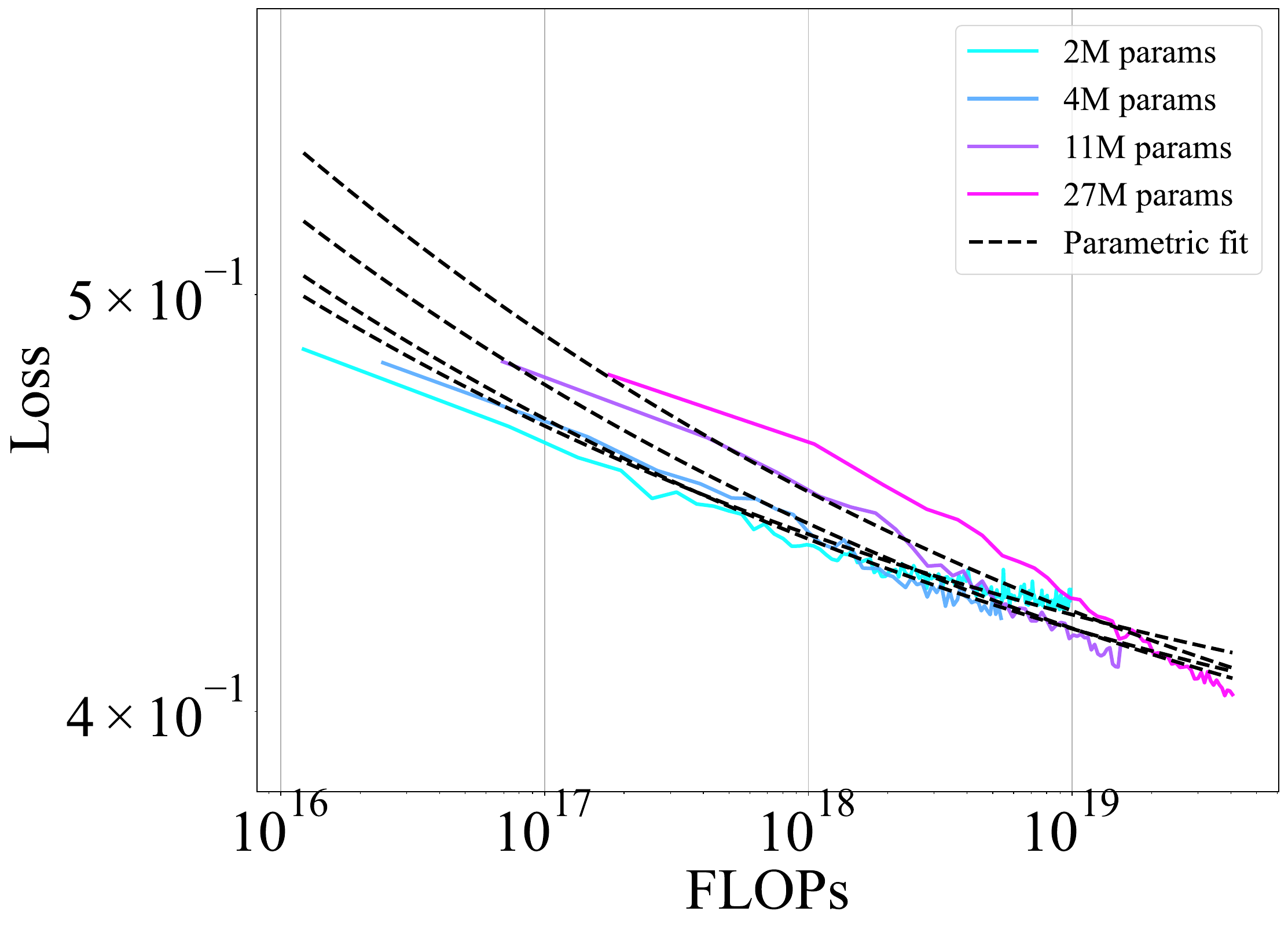}
\includegraphics[width=0.32\columnwidth]{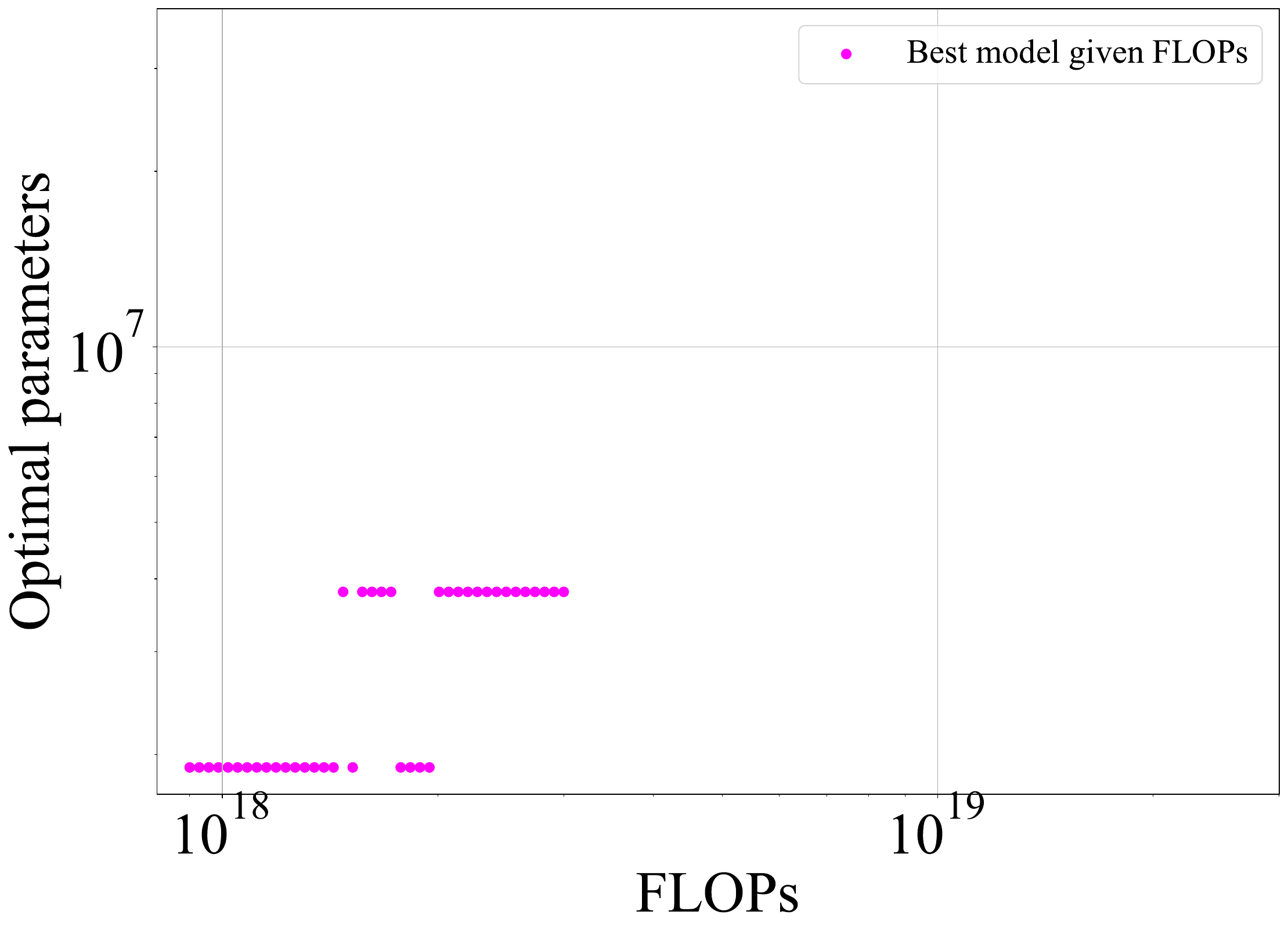}
\includegraphics[width=0.32\columnwidth]{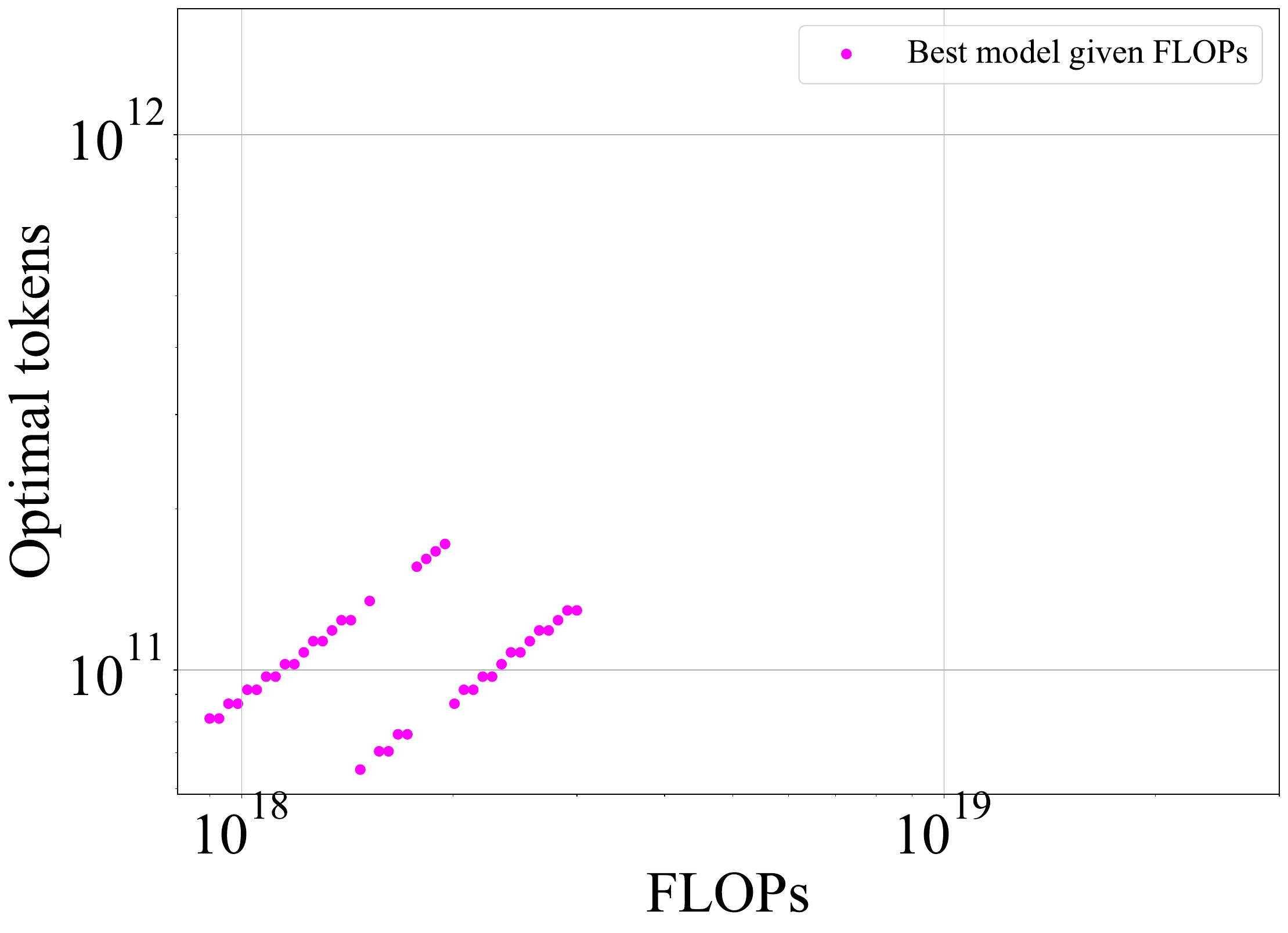}
\caption{BC-Token scaling with $d_z=$ 540 tokens per image observation. Models above 2M parameters do not saturate over the FLOPs range considered and coefficients can not be inferred using the \textit{frontier fit} method. 
}
\label{fig_BC_540_tok}
\end{center}
\end{figure}

\begin{figure}[h!]
\begin{center}
\includegraphics[width=0.32\columnwidth]{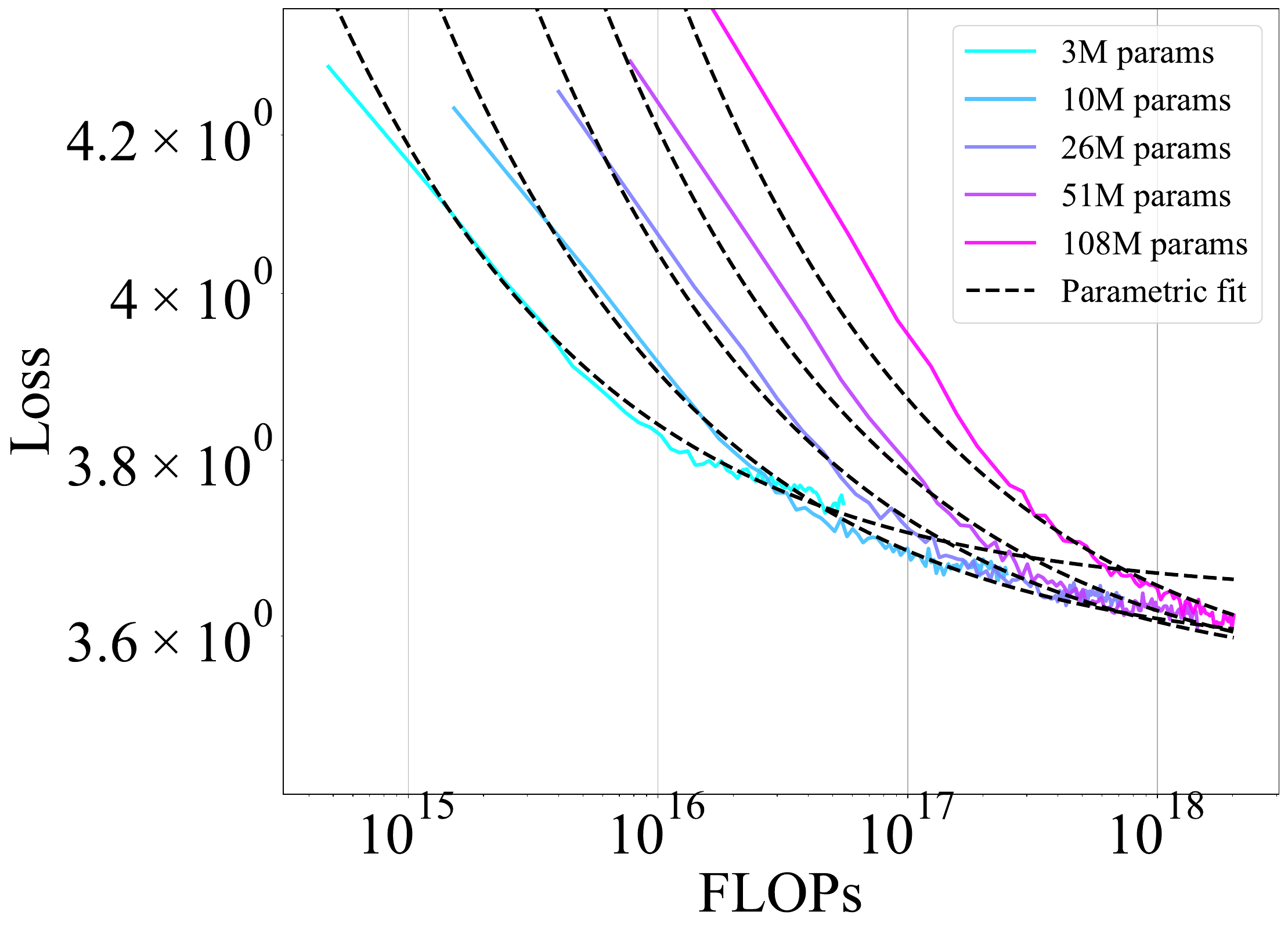}
\includegraphics[width=0.32\columnwidth]{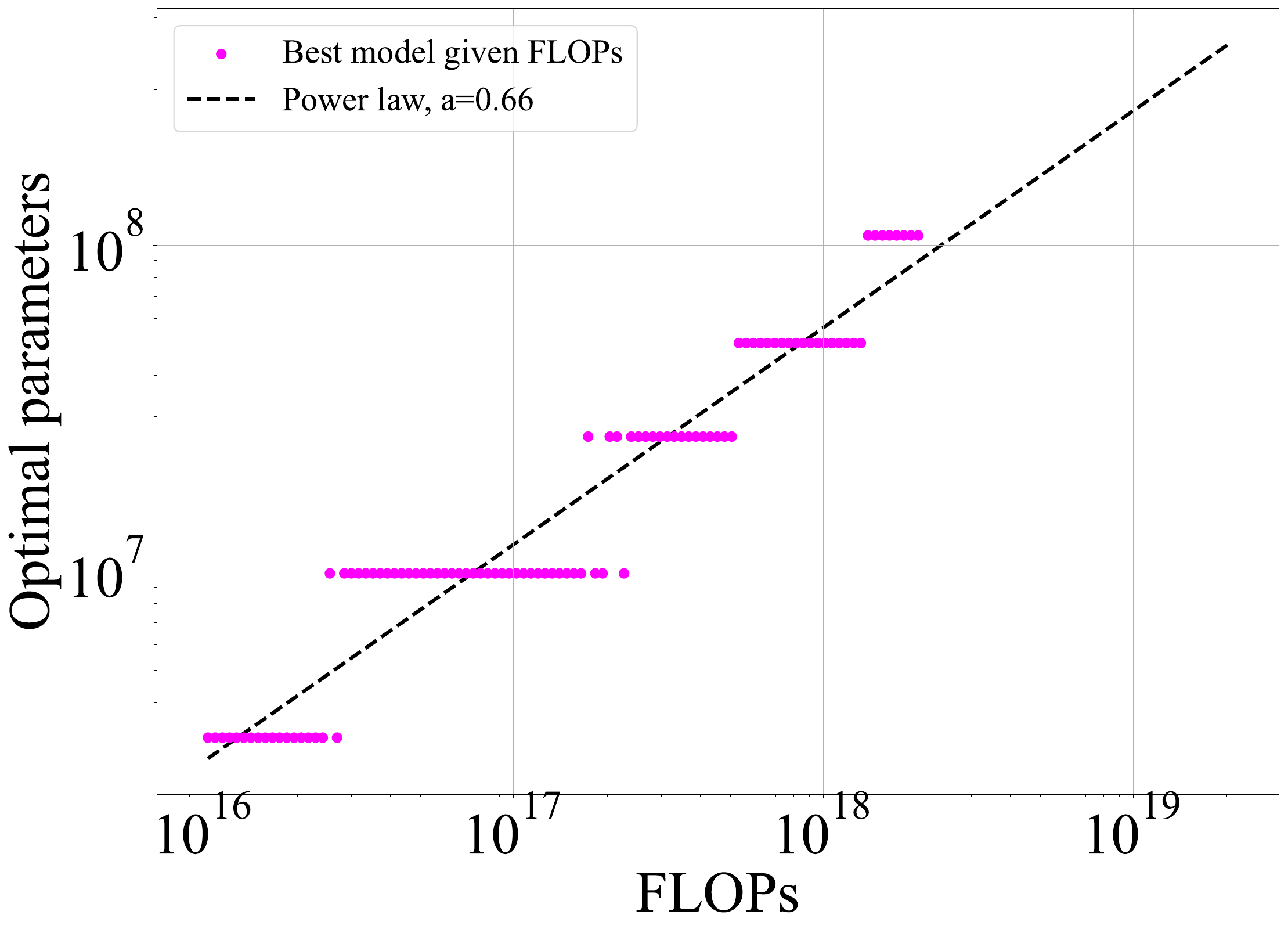}
\includegraphics[width=0.32\columnwidth]{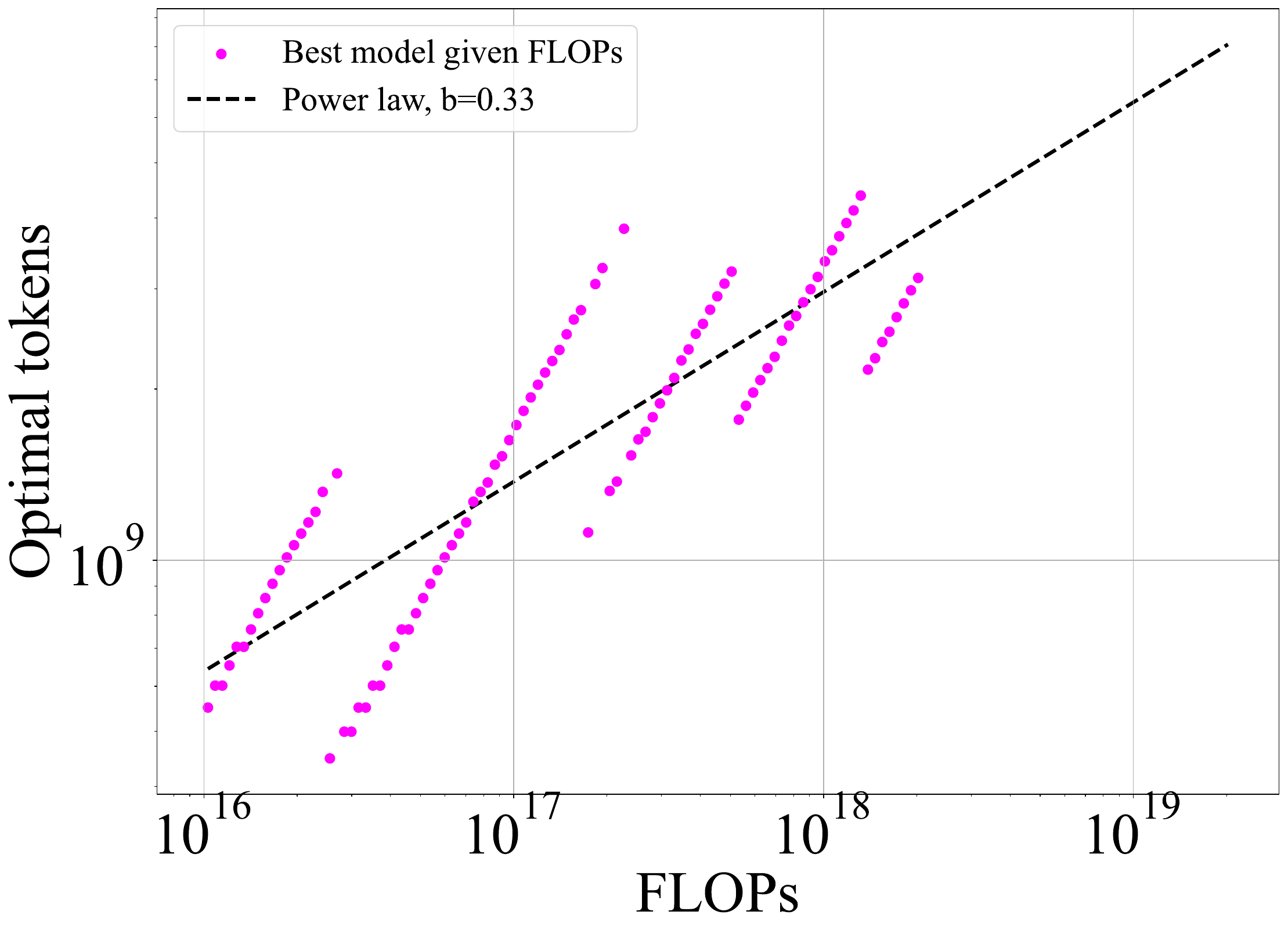}
\caption{BC-CNN scaling.  Left shows the \textit{parametric fit}. Middle \& right show the \textit{frontier fit} estimating optimal model \& dataset size respectively.
Compared to the results for BC-Token the model sizes considered compute-optimal are considerably larger.
The power law coefficient for $N_\text{optimal}$ also increases significantly from $0.32$ to $0.66$ skewing towards scaling model size as opposed to dataset size when scaling up compute.
}
\label{fig_BC_CNN}
\end{center}
\end{figure}

We present our results on the scaling law coefficients for BC-Token in Figure \ref{fig_BC_540_tok}.
Despite sharing an architecture with WM-Token-540 we now observe the opposite dependence on model and dataset sizes.
The coefficients skew heavily towards dataset size; $N_\text{optimal}=0.32$, $D_\text{optimal}=0.68$ (compared to $N_\text{optimal}=0.62$, $D_\text{optimal}=0.37$ -- explained in Section \ref{sec_understand_q1}).
Furthermore, under the same compute budget the compute-optimal model sizes are significantly smaller.
For a compute budget of $10^{18}$ and $10^{19}$ FLOPs we find that model sizes of $2$M and $11$M are compute-optimal for BC-Token-540 compared to $27$M and $110M$ for WM-Token-540.
In our experiments, we observe the losses for the BC-Token models take much longer to plateau leading to less overlap between model sizes.
This results in the \textit{frontier fit} not being suitable for accurately estimating the scaling law coefficients, hence we rely on the \textit{parametric fit} for these results. 

To better understand the change in the scaling law coefficients, we now consider the BC-CNN architecture for the task of BC in Figure \ref{fig_BC_CNN}.
For this architecture, we observe that the coefficients now skew towards model size (similarly to those in \citep{tuyls2023scalingimitation}), with $N_\text{optimal}=0.66$, and $D_\text{optimal}=0.34$.
Section \ref{sec_understand_q2} provides more intuition on the differences between the WM-Token and BC-Token setups that lead to this change.

Further to studying the differences in scaling law coefficients between tasks and architectures, we also study the accuracy of extrapolation. 

\subsection{Extrapolation in world modeling}
\label{sec_mainresults_extrap_wm}

\begin{figure}[h!]
\begin{center}
\includegraphics[width=0.32\columnwidth]{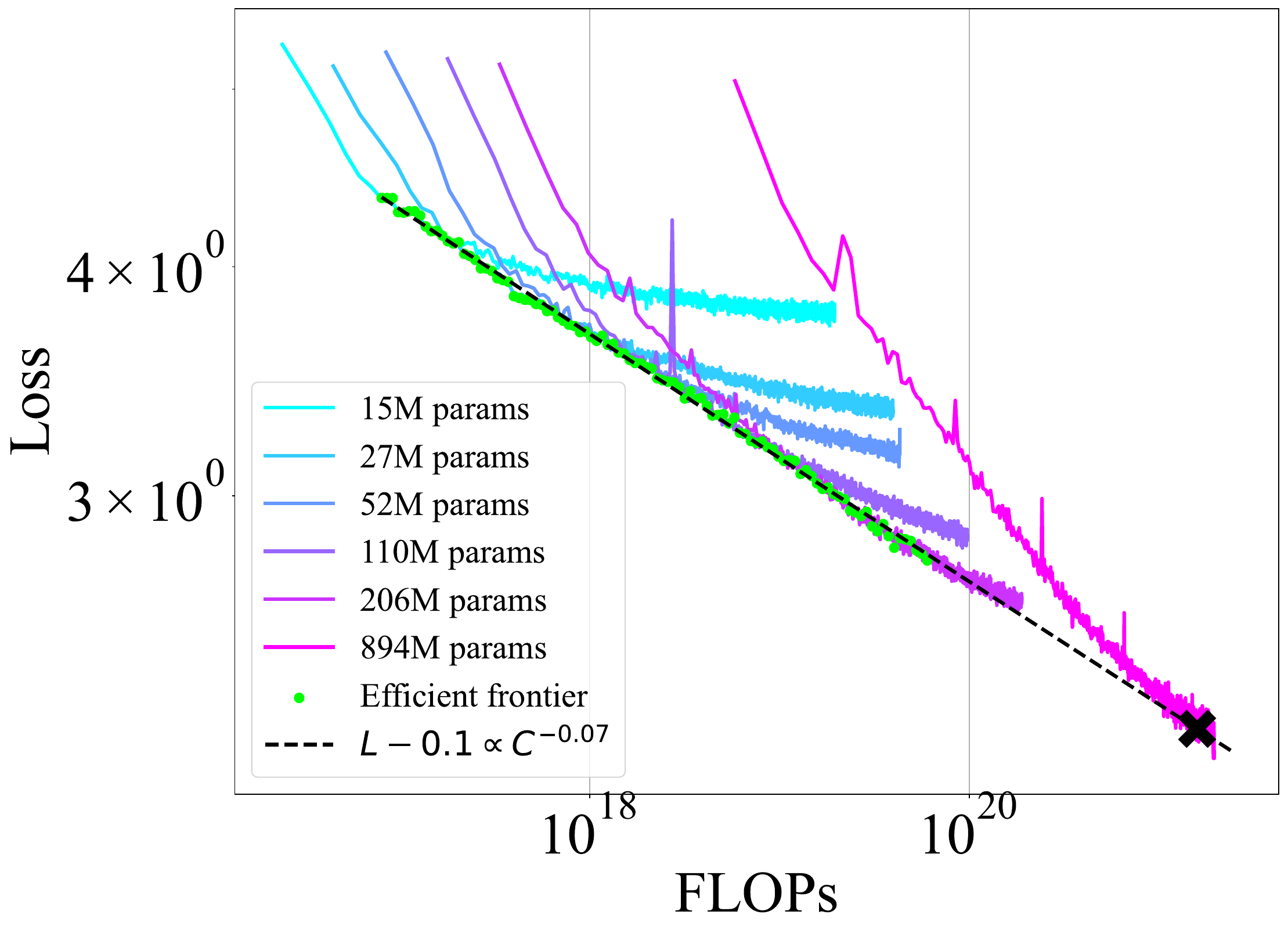}
\includegraphics[width=0.32\columnwidth]{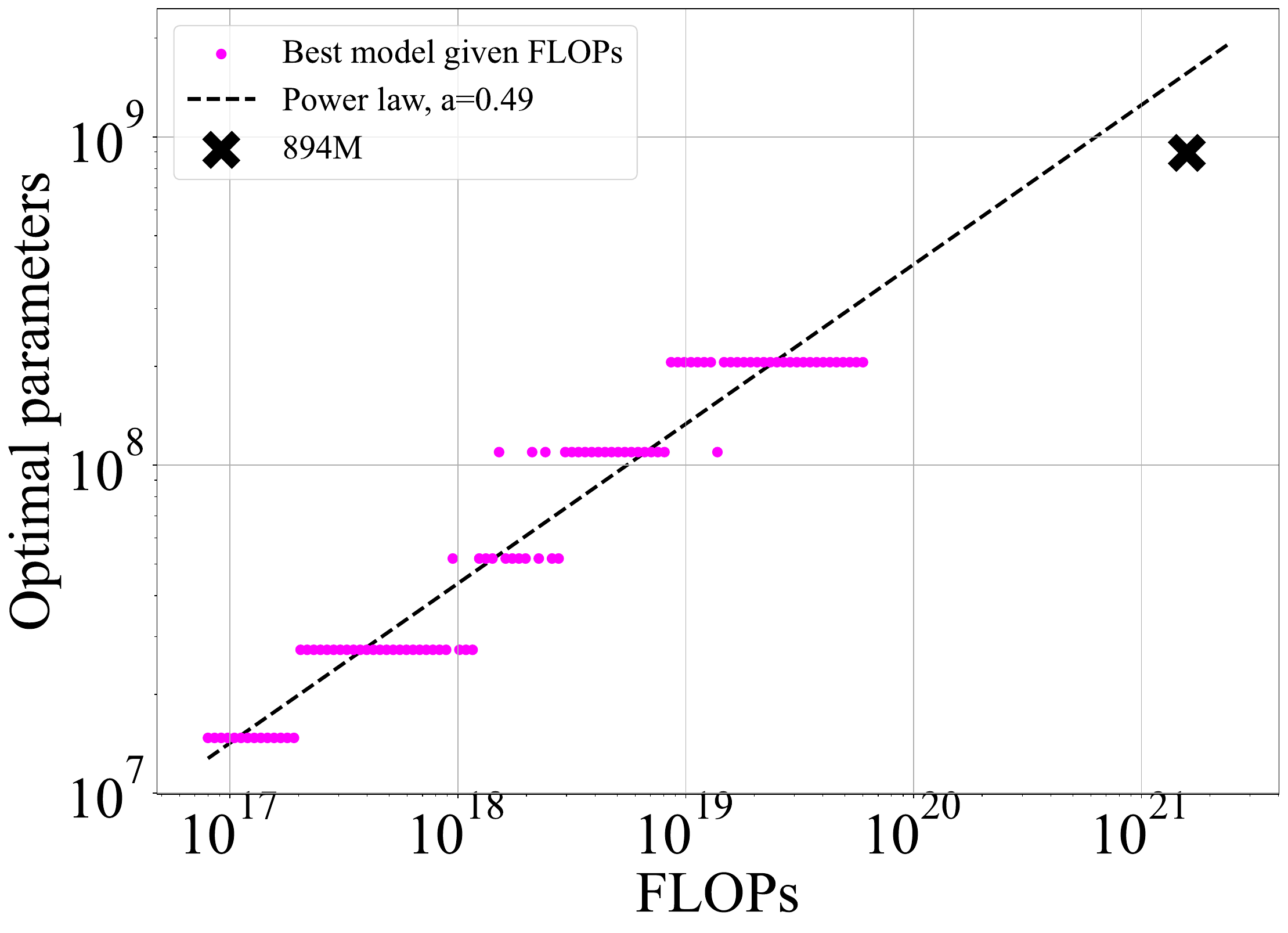}
\includegraphics[width=0.32\columnwidth]{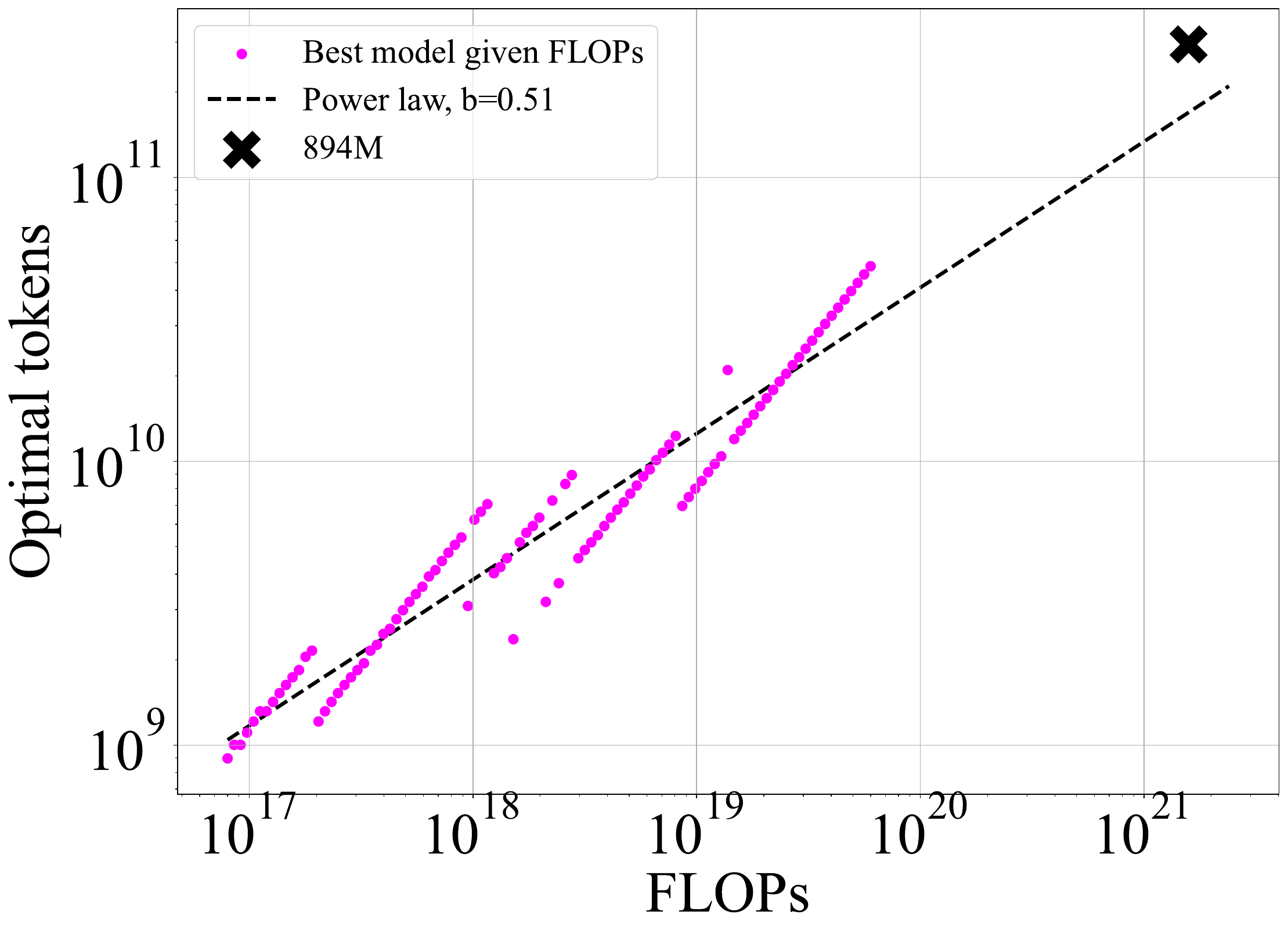}
\caption{Testing the extrapolation capability of our derived scaling law for WM-Token-256 by training an $894$M parameter model with an order of magnitude more compute than was used for the scaling law analyses.
We observe good agreement between our predicted optimal loss/model size/number of training tokens (dotted lines) and our actual training run. 
}
\label{fig_wm_256_extrap}
\end{center}
\end{figure}

To test the extrapolation accuracy of our derived scaling laws, we train a $894M$ parameter WM-Token-256 model with an order of magnitude more compute than used for the scaling law analyses.
Figure \ref{fig_wm_256_extrap} presents both the learning curve as well as the extrapolated lines derived from the Frontier fit method.
We take the point with the loss value closest to our extrapolated loss curve ($\sim1.58 \times 10^{21}$FLOPS), and mark it on the Frontier fit extrapolations.
We observe very good agreement between that point and our compute-optimal predictions for both model and dataset size, demonstrating the accuracy of our derived scaling laws.
The gap between our prediction and the actual training run suggests we could further optimize the hyperparameters (learning rate and batch size in particular) for the $894$M parameter model, which was not extensively tuned due to compute requirements.

\section{Further analysis}
\label{sec_understanding}

Section \ref{sec_mainresults} made several observations about the effect of scale in the pre-training of embodied agents. This section aims to understand these results further, and provide intuition for why they occur. Specifically we target three questions.
\begin{itemize}
    \item Q1: Why does BC-Token produce training curves that do not plateau, while WM-Token does, given an identical architecture and dataset? (Section \ref{sec_understand_q1})
    \item Q2: Why does moving from BC-Token to BC-CNN resolve this issue? (Section \ref{sec_understand_q2})
    \item Q3: Why does increasing the tokens per image observation (256 to 540) lead to an increase in the optimal model size coefficient (0.49 to 0.62)? (Section \ref{sec_understand_q3})
\end{itemize}

\subsection{Q1: BC-Token vs. WM-Token}
\label{sec_understand_q1}

The lack of saturation of BC-Token models compared to WM-Token models can be attributed to two factors. 
The first is a sparser loss. A single observation-action pair is discretized into $d_z + d_a$ total tokens. With the large VQGAN tokenizer, world modeling receives supervision for $d_z/(d_z + d_a) = 540/556 \approx 97\%$ tokens, while BC is supervised for $d_a/(d_z + d_a) = 16/556 \approx 3\%$ of tokens. 

The second factor is the granularity of the targets. The large tokenizer creates a world modeling vocabulary size of 4096. Each vocabulary item roughly corresponds to a specific color and texture for an image patch. Many vocabulary items may only be used to model specific map regions or special abilities. Hence, the world modeling loss is very granular.
On the other hand, a player can take the same action in multiple different situations -- continue straight could be used to escape an enemy, chase an enemy, or navigate to a checkpoint. Hence, the supervision for BC is more vague and abstracted. We can think of this as a \textit{super-classed} label.

To demonstrate the effect of these two factors on optimal model size coefficients, we run a set of tiny-scale experiments in language modeling. 
Transformers are trained on next-character prediction, on a dataset of Shakespeare text\footnote{Shakespeare character dataset from: \url{https://github.com/karpathy/nanoGPT}} using a single character for each token. Model sizes are varied from 4k parameters to 17M parameters. Context length is fixed at 16 characters/tokens. 

\begin{figure}[h!]
\begin{center}
\hspace{0.4in}
Dense loss \hspace{1.2in}
Sparse loss \hspace{0.6in}
Sparse loss, super-classed  \\
\small
$N_\text{optimal} \propto C^{0.63}$, $D_\text{optimal} \propto C^{0.37}$ \hspace{0.01in}
$N_\text{optimal}\propto C^{0.50}$, $D_\text{optimal} \propto C^{0.50}$ \hspace{0.01in}
$N_\text{optimal} \propto C^{0.15}$, $D_\text{optimal} \propto C^{85}$ \\
\includegraphics[width=0.32\columnwidth]{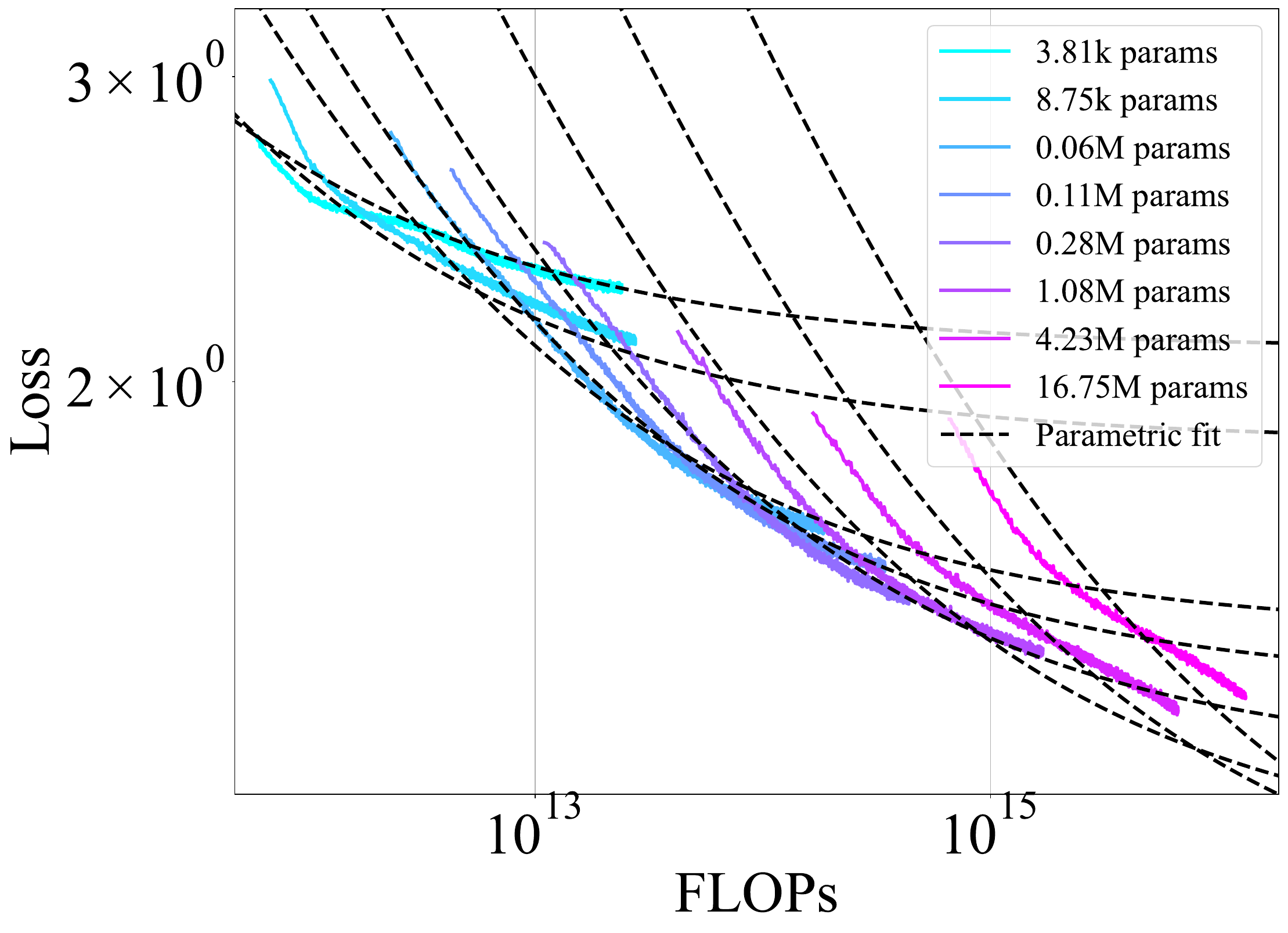}
\includegraphics[width=0.32\columnwidth]{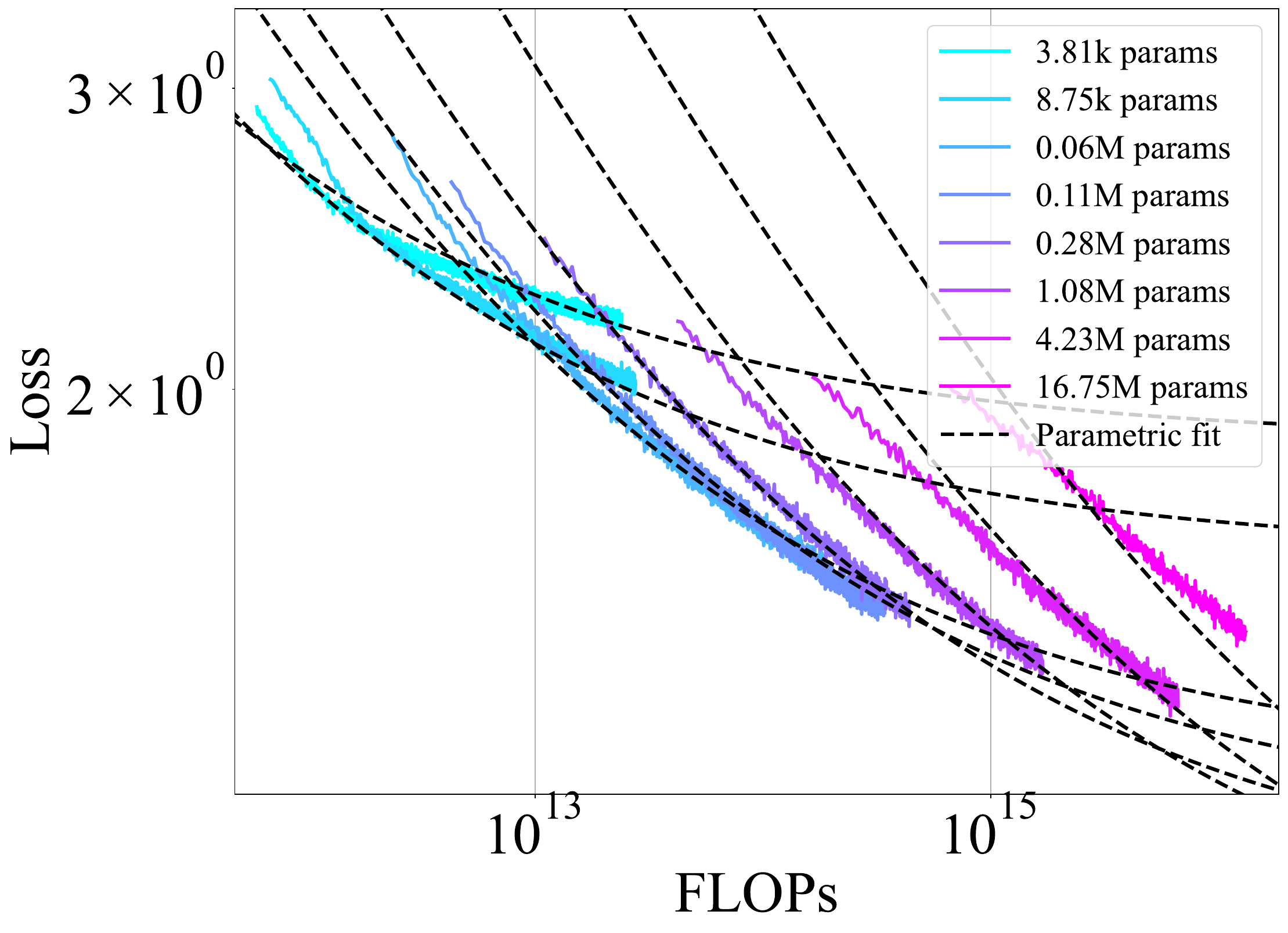}
\includegraphics[width=0.32\columnwidth]{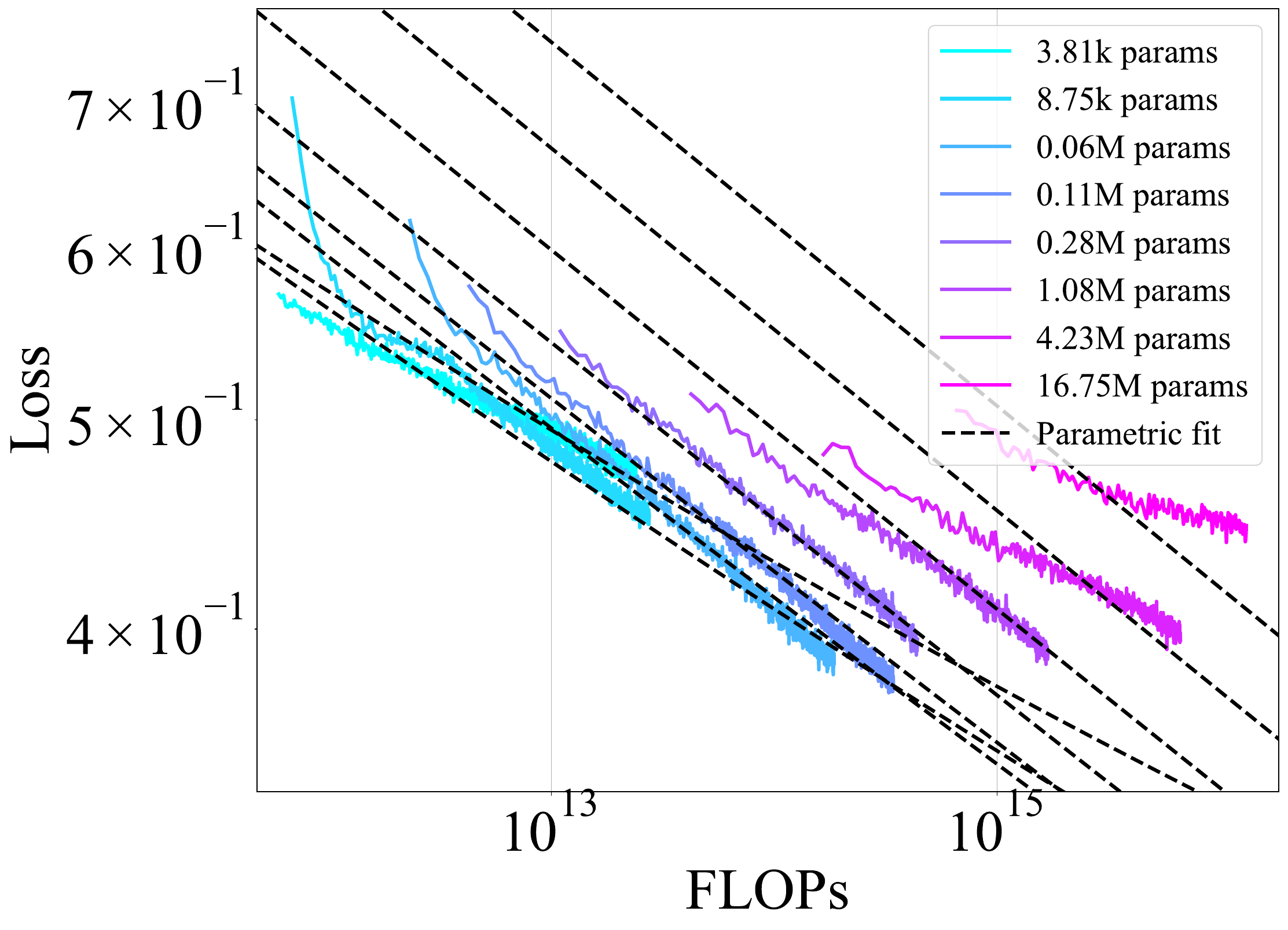}
\caption{Training curves and parametric fit for character modeling experiments. The standard dense LLM loss has been modified to reflect properties of BC -- a sparse loss (one of 16 tokens), and then additionally super-classing the targets into two classes. }
\label{fig_llm_ablation}
\end{center}
\end{figure}

Figure \ref{fig_llm_ablation} (left) shows training curves over all 16 tokens, followed by a \textit{sparse} loss where supervision is only provided from the final token (middle), and then additionally under a \textit{super-classed} setting (right). This super-classes the final target -- rather than using all 128 ASCII characters, they are randomly shuffled into one of two macro classes.

These modifications are intended to mirror the effect of moving from WM-Token to BC-Token. We compute optimal model size coefficients using the parametric fit  as most models are not trained long enough for the frontier fit method. Indeed, we see that the coefficient drops from 0.63 to 0.15 with both the sparse and super-classed loss. This matches the magnitude of decrease seen in Table \ref{tbl_main_coeffs} from 0.66 to 0.32, indicating that the proposed mechanisms explain our findings.

\subsection{Q2: BC-Token vs. BC-CNN}
\label{sec_understand_q2}

Despite the same non-granular loss signal, why does switching architecture from BC-Token to BC-CNN makes the loss of similar model sizes plateau under a much smaller compute budget?

Consider each architecture using a transformer with 1M parameters.
Observe from Figure \ref{fig_architectures_tasks} that BC-Token receives $d_z + d_a = 556$ inputs for every action $\hat{\mathbf{a}}_t$ it predicts, while BC-CNN receives just one input for every action predicted. Hence, BC-Token uses around 556 times more compute in its action prediction ($556 \times 2 \times 1M \approx 1 \times 10^9$ FLOPs) than BC-CNN ($1 \times 2 \times 1M \approx 2 \times 10^6$ FLOPs). This means that even with the same number of parameters, BC-Token can learn a far more expressive function than BC-CNN. Hence, BC-Token requires far more tokens to match this expressivity, and training curves for a given model size plateau much later.

\subsection{Q3: WM-Token-256 vs. WM-Token-540}
\label{sec_understand_q3}

Finally, we seek to understand why the optimal model size coefficient increases when moving from the 256 to the 540 token VQGAN.
As the number of tokens per image observation are increased, the compression rate of the tokenized representation decreases. 
We would expect that each individual token becomes easier to predict in this less compressed representation. This would mean a less expressive function is needed (smaller model size), but also a smaller number of examples would need to be seen (smaller dataset size). 
It is less clear what ratios these ingredients decrease in, and hence what effect a lower compression rate has on the optimal model size coefficient.

Using the small scale RT-1 dataset, we conduct a more thorough investigation of the effect of tokens-per-image observation on scaling coefficients. First we train a range of image tokenizers with $z_o \in [16,36,64,100,256]$, visualized in Figure \ref{fig_rt1_reconstruct}.
For each VQVAE, we then train a range of WM-Token model sizes $N \in [0.08M, 0.2M, 0.28M, 0.54M, 0.99M]$, and measure scaling coefficients using the frontier fit method, repeating three times. 

\begin{figure}[t!]
\begin{center}
\includegraphics[width=0.5\columnwidth]{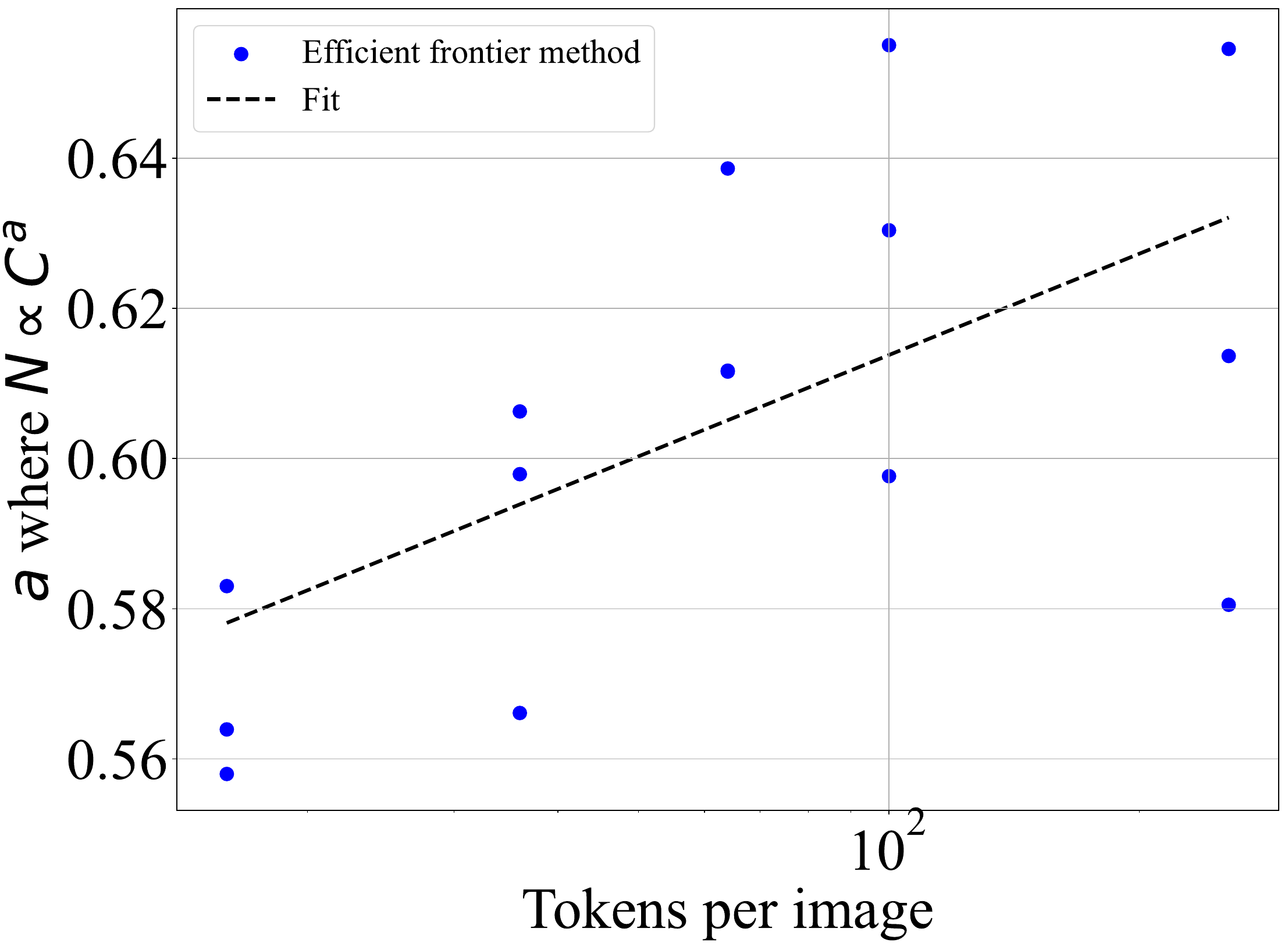}
\caption{RT-1 experiments. Optimal parameter coefficient vs. number of tokens per observation, with three repeated runs per VQVAE.}
\label{fig_rt1_zo_vs_a}
\end{center}
\end{figure}

Figure \ref{fig_rt1_zo_vs_a} plots all coefficient vs. tokens-per-image -- we observe that the optimal parameter scaling coefficient increases with decreasing compression.

To investigate whether compression affects the optimal model size coefficient outside of embodied domains, we ran a small scale experiment in language modeling using two text representations; 1) ASCII character-level tokenization. (low compression) 2) GPT-2 tokenizer (high compression). We used the BookCorpus dataset \citep{Zhu_2015_ICCV}, and trained models past their compute-optimal point, so the \textit{Frontier fit} method could be used for coefficient estimation.

Appendix \ref{sec_app_understanding_exp} shows results. Under the character-level tokenizer (low compression), we find $N_\text{optimal} \propto C^{0.66}$. For the GPT-2 tokenizer (high compression), we find $N_\text{optimal} \propto C^{0.44}$. 
Hence, in language the more compressed representation also leads to a lower optimal model size coefficient.

\section{Discussion \& conclusion}
\label{sec_discussion}

This paper establishes a deeper understanding of scaling laws for world modeling and behavior cloning, two tasks that underpin embodied AI applications in domains such as video games and robotics. Focusing on generative pre-training of such models, we show that
it is possible to recover scaling laws similar to those established in the LLM literature. Establishing such a link is key to making efficient use of available resources, and to training compute-optimal models.

Considering the task of world modeling, we find that models can be smoothly scaled following best practices and insights from the LLM literature. Surprisingly, the scaling coefficients for our WM-Token-256 architecture very closely match those established for LLMs. Comparing to our WM-Token-540 model and additional analysis, we further establish that scaling is affected by the tokenizer's compression rate.

Turning to pre-training BC policies for agents, the choice of architecture is extremely important in determining optimal scaling behavior. 
When using architectures with tokenized image observations, dataset size should be increased much more rapidly than model size. Meanwhile, for BC-CNN architectures, model size should be increased faster than dataset size.

\textbf{Limitations.} 
While we show that scaling laws can be precisely described in the infinite data regime and for appropriate architectures, future work is needed to establish scaling laws for alternative models and under varying dataset quality. In addition, we focus on loss as an intermediate quantity that can be effectively optimized in pre-training. Many additional considerations are required for effective AI models, such as downstream task performance and model inference times. How valuable scaling laws can be in providing insights relevant to those choices remains an open question.

\textbf{Ethics Statement.} 
Data for this project was provided via a partnership with \emph{Ninja Theory}, who collected a large corpus of human gameplay data for their game \emph{Bleeding Edge}. Data collection was covered by an End User License Agreement (EULA) and our use of the data was governed by a data sharing agreement with the game studio, and approved by our institution's IRB. This data was recorded between September 2020 and October 2022. To minimize risk to human subjects, any personally identifiable information (Xbox user ID) was removed from the data. The resulting data was cleaned to remove errors and data from bad actors.



\printbibliography

\newpage
\appendix

The appendix is organized as follows.
\begin{itemize}
    \item Appendix \ref{sec_app_main_exp} contains details on the training of the model configurations, hyperparameters, and a description of the datasets used. 
    \item Appendix \ref{sec_app_understanding_exp} contains results from Section \ref{sec_understand_q3}.
    \item Appendix \ref{sec_app_robotics} contains further details on training world models on robotics.
    \item Appendix \ref{sec_app_pretrain_evidence} contains further results demonstrating the link between pre-training loss and performance. 
\end{itemize} 




\section{Scaling experiments further details}
\label{sec_app_main_exp}

This section provides experimental details for all experiments on the primary Bleeding Edge dataset.

\subsection{Hyperparameters}
We trained two VQGANs from scratch with reconstruction losses. 
\begin{itemize}
    \item \textbf{BE-Small.} Based on \cite{esser2021vqgan},  uses $d_z=256$, $V_o=4096$, $h=w=128$, with 28M parameters, and a CNN design. It was trained on a single SkyGarden Bleeding Edge map. 
    \item \textbf{BE-Large.} Based on \cite{yu2022vqganvit},  uses $d_z=540$, $V_o=4096$, $h=180$, $w=300$, with 150M parameters, and a vision transformer design. It was trained on all seven Bleeding Edge maps. 
\end{itemize}
We selected the numbers of tokens per image based on qualitative assessment of reconstructions. We found that 256 tokens per image was the minimum that still allowed a reconstruction to capture the majority of salient gameplay details. However certain details still were lacking, such as an enemy player's health bars -- hence we also considered a 540 token version that provided a higher quality reconstruction.

BC-CNN details. We use $h=w=128$. The $0.6M$ paramter CNN is similar to that used by \citep{baker2022vpt}, however it uses ConvNext blocks \citep{liu_convnet_2022}. 
The CNN produces an embedding of size $1024$ which is then put through a linear layer to obtain a vector matching the transformer's embedding dimension.

Transformer configurations are given in Table \ref{tbl_transfomer_configs}. We describe the parameters for the WM-Token architecture.
Note that MLP layers are four times the width of embed dim.
Model configurations roughly followed the model configurations used in Table A9 of \cite{hoffmann2022training}, where residual stream dimension, number of layers, and number of heads were roughly increased proportionally.

\begin{table}[h!]
\caption{Transformer configurations. Here $N$ is listed for the tokenized architectures. Parameter count varies slightly for BC-CNN due to inclusion of the embedding CNN and differing numbers of embedding parameters sizes.}
\begin{center}
\label{tbl_transfomer_configs}
\begin{tabular}{lrrrrr}
     \multicolumn{1}{c}{\bf $N$}  &\multicolumn{1}{c}{\bf Layers} 
     &\multicolumn{1}{c}{\bf Num heads} 
     &\multicolumn{1}{c}{\bf Embed dim} 
\\ \hline \\
    2M & 3 & 3 & 180 \\
    4M & 4 & 4 & 240 \\
    11M & 6 & 6 & 360 \\
    15M & 4 & 4 & 512 \\
    27M & 8 & 8 & 512 \\
    52M & 10 & 10 & 640 \\
    110M & 15 & 12 & 768 \\
    206M & 16 & 16 & 1024 \\
    894M & 23 & 14 & 1792 
\end{tabular}
\end{center}
\end{table}

\subsection{Training details}

All transformers are trained with a variant of nanoGPT \citep{Karpathy2022} using PyTorch Lightning \citep{Falcon_PyTorch_Lightning_2019}.

This section lists key hyperparameters. Note that it was important to find optimization settings that produced the lowest possible loss for a given model size. In general larger models require smaller learning rates. Our approach first optimized the smallest model through a grid sweep, we would then sequentially run a sweep over the next largest model, starting at the smaller model's optimized learning rate. Table 3-6 provide final settings.

\begin{table}[h!]
\caption{Hyperparameters for WM-Token with $d_z=$256 tokens per image observation. 
}
\begin{center}
\label{tbl_training_hyperparams_WM_256Token_BE}
\begin{tabular}{lrrrrr}
     \multicolumn{1}{c}{\bf $N$}  
     & \multicolumn{1}{c}{\bf Seq len} 
     & \multicolumn{1}{c}{\bf Context length} 
     & \multicolumn{1}{c}{\bf Tokens per update} 
     & \multicolumn{1}{c}{\bf Learning rate} 
     \\ \hline \\
    15M & 10 & 2,720 & 522,240 & 0.0007 \\
    27M & 10 & 2,720 & 522,240 & 0.0007 \\
    52M & 10 & 2,720 & 522,240 & 0.0007 \\
    110M & 10 & 2,720 & 522,240 & 0.0007 \\
    206M & 10 & 2,720 & 522,240 & 0.00057 \\
    894M & 10 & 2,720 & 2M  & 0.00028 \\
\end{tabular}
\end{center}
\end{table}

\begin{table}[h!]
\caption{Hyperparameters for WM-Token with $d_z=$540 tokens per image observation.}
\begin{center}
\label{tbl_training_hyperparams_WM_540Token_BE}
\begin{tabular}{lrrrrr}
     \multicolumn{1}{c}{\bf $N$}  
     & \multicolumn{1}{c}{\bf Seq len} 
     & \multicolumn{1}{c}{\bf Context length} 
     & \multicolumn{1}{c}{\bf Tokens per update} 
     & \multicolumn{1}{c}{\bf Learning rate} 
     \\ \hline \\
    4M & 10 & 5,560 & 533,760 & 0.005 \\
    11M & 10 & 5,560 & 533,760 & 0.001 \\
    27M & 10 & 5,560 & 533,760 & 0.001 \\
    52M & 10 & 5,560 & 533,760 & 0.001 \\
    110M & 10 & 5,560 & 533,760 & 0.0005 \\
    206M & 10 & 5,560 & 533,760 & 0.0005 \\
\end{tabular}
\end{center}
\end{table}

\begin{table}[h!]
\caption{Hyperparameters for BC-Token with $d_z=$540 tokens per image observation.}
\begin{center}
\label{tbl_training_hyperparams_BC_540Token_BE}
\begin{tabular}{lrrrrr}
     \multicolumn{1}{c}{\bf $N$}  
     & \multicolumn{1}{c}{\bf Seq len} 
     & \multicolumn{1}{c}{\bf Context length} 
     & \multicolumn{1}{c}{\bf Tokens per update} 
     & \multicolumn{1}{c}{\bf Learning rate} 
     \\ \hline \\
    2M & 10 & 5,560 & 533,760 & 0.0005 \\
    4M & 10 & 5,560 & 533,760 & 0.0005 \\
    11M & 10 & 5,560 & 533,760 & 0.0001 \\
    27M & 10 & 5,560 & 533,760 & 0.0001 \\
\end{tabular}
\end{center}
\end{table}

\begin{table}[h!]
\caption{Hyperparameters for BC-CNN.}
\begin{center}
\label{tbl_training_hyperparams_BC_CNN}
\begin{tabular}{lrrrrr}
     \multicolumn{1}{c}{\bf $N$}  
     & \multicolumn{1}{c}{\bf Seq len} 
     & \multicolumn{1}{c}{\bf Context length} 
     & \multicolumn{1}{c}{\bf Items per update} 
     & \multicolumn{1}{c}{\bf Learning rate} 
     \\ \hline \\
    2M & 10 & 10 & 2560 & 0.0003 \\
    3M & 10 & 10 & 2560 & 0.0003 \\
    10M & 10 & 10 & 2560 & 0.0003 \\
    26M & 10 & 10 & 2560 & 0.0003 \\
    51M & 10 & 10 & 2560 & 0.0003 \\
\end{tabular}
\end{center}
\end{table}

\subsection{Dataset details}
\label{subsec:data-details}

Image observations were stored in MP4 format at 60fps, alongside binary files containing the associated controller actions. A time code extracted from the game was stored for each frame, to ensure actions and frames remained in sync at training time. 

The \textit{7 Maps} dataset comprised {60,986} matches, yielding {530,713} individual player trajectories (each around 9 minutes), totaling 27.89 TiB on disk. This amounted to around 8.6 years of gameplay. After downsampling to 10Hz (the frequency models are trained on), this equated to 1.63B frames. This was then divided into training / validation / test sets by dividing the matches with an 80:10:10 split.

Our filtered \textit{Sky Garden} dataset used the same 80:10:10 split and 10Hz downsampling, but focused on just one map, yielding {71,940} individual player trajectories, or 355.5M frames (around 1.12 years of game play).

For discretizing the controller actions, while the buttons are natively discrete, we discretize the x and y values of the left and right joysticks into eleven buckets.


\subsubsection{Infinite data regime allowed FLOPs}
\label{sec_appendix_infinite}
We wish to study scaling in the infinite data regime, where training loss is not significantly effected by models repeatedly training on the same datapoints which can lead to overfitting effects.
This section calculates the number of training tokens allowed for each model family trained in this work. 
Viewing Figure \ref{fig_intro_overview} alongside these numbers confirms that models remain in the infinite data regime for all our experiments.

\textbf{WM-Token-540, BC-Token-540.} 
We trained on the \textit{7 maps} dataset, with 1.63B observation-action pairs. Models used the tokenized architecture with the large VQGAN, so each observation-action pair creates $540+16=556$ transformer inputs, for a total of 1.63B$\times 556 = 906$B training tokens. 
\cite{muennighoff2024repeatscaling} observe that tokens may be reused up to four times with negligible departure from the infinite data regime. This produces 3.6T tokens. 
For a 200M parameter model the compute allowed by the infinite data regime is $C=6ND=6 \times 200\text{M} \times 3.6\text{T} = 4.3\times 10^{21}$ FLOPs.

\textbf{WM-Token-256.} This is trained on the \textit{Sky Garden} dataset, with 355M observation-action pairs. Each pair is split into $256+16=272$ tokens, for 97B training tokens, or 97B$\times 4 = 386$B effective tokens. 
For a 200M parameter model the compute allowed by the `infinite data regime' is $C=6ND=6 \times 200\text{M} \times 386\text{B} = 4.6\times 10^{20}$ FLOPs.

\textbf{BC-CNN.}  Trained on \textit{7 maps} dataset, but now with one token per observation-action pair, this creates a possible $1.63\text{B} \times 4 = 6.52\text{B}$ effective tokens. A 50M parameter model uses $C=6ND=6 \times 50\text{M} \times 6.52\text{B} = 2.0\times 10^{18}$ FLOPs.

\section{Further analysis details}
\label{sec_app_understanding_exp}

Experimental results supporting Section \ref{sec_understand_q3}.

\begin{figure}[h!]
\begin{center}
\includegraphics[width=0.32\columnwidth]{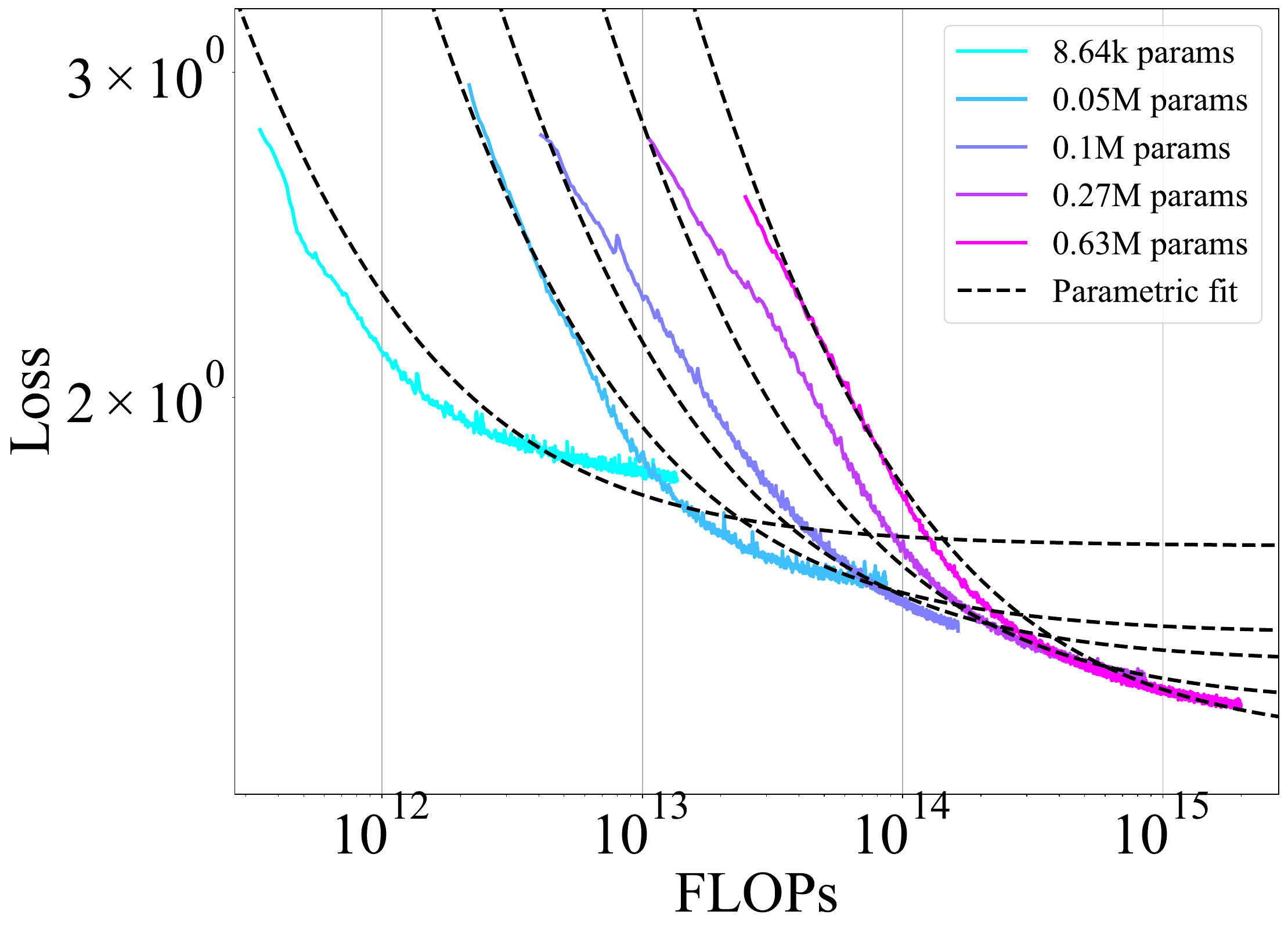}
\includegraphics[width=0.32\columnwidth]{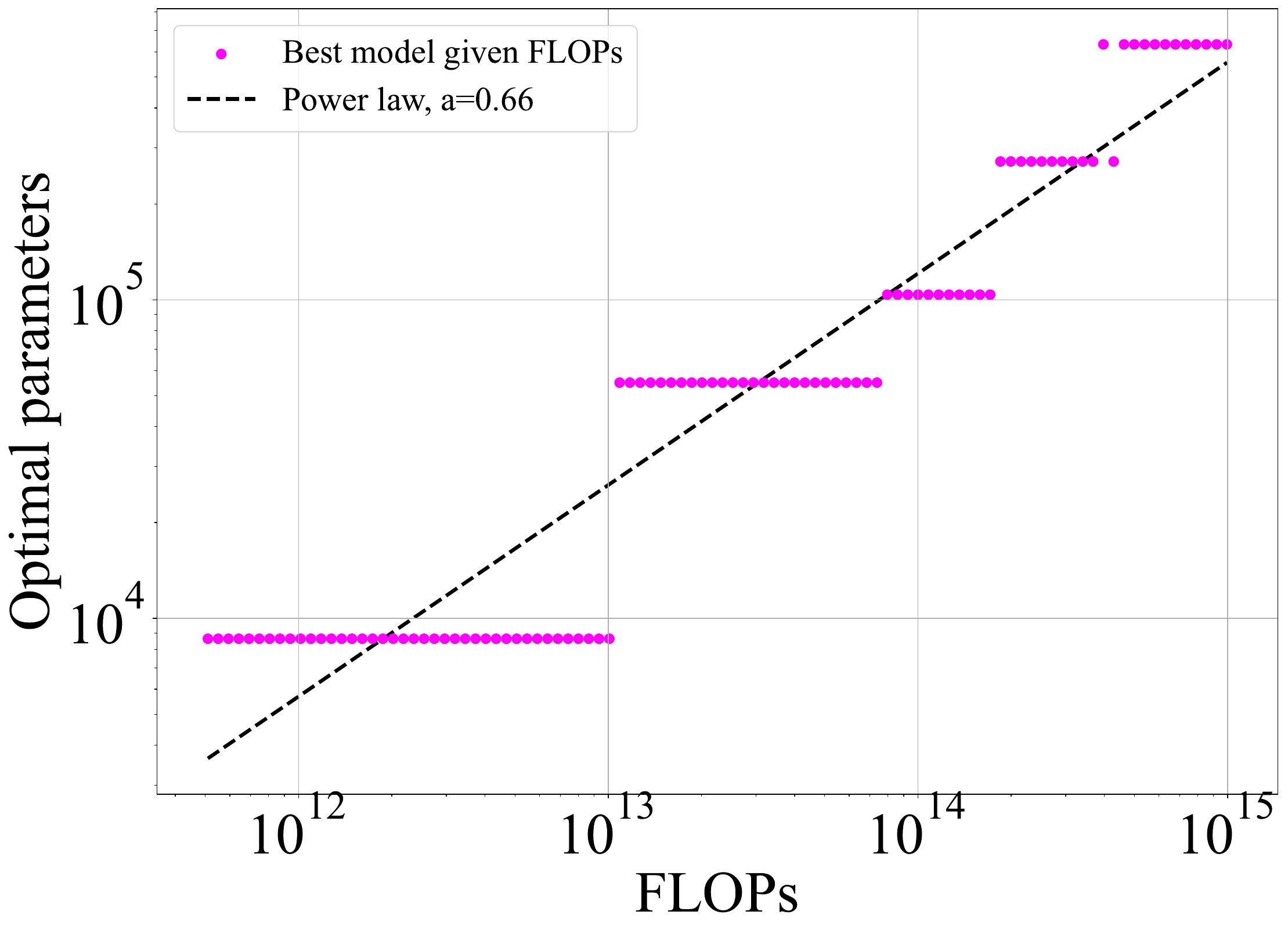}
\includegraphics[width=0.32\columnwidth]{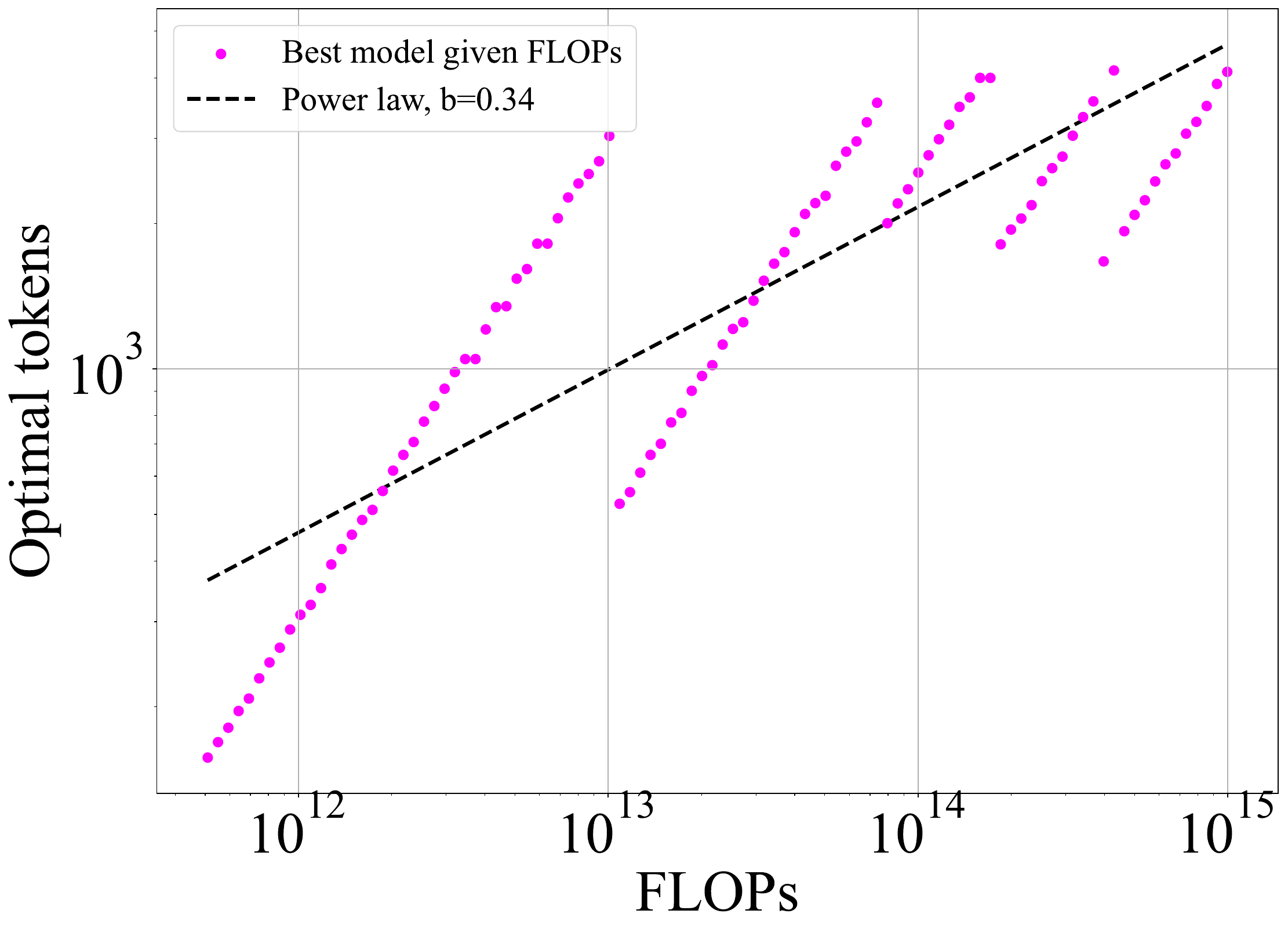}
\caption{Relating to Section \ref{sec_understand_q3}, character-level (low compression).
Utilising the \textit{frontier} fit (middle and right) we derive the power law coefficient for $N_\text{optimal}$ as $0.66$ and $D_\text{optimal}$ as $0.34$.
}
\label{fig_understand_compression_char}
\end{center}
\end{figure}

\begin{figure}[h!]
\begin{center}
\includegraphics[width=0.32\columnwidth]{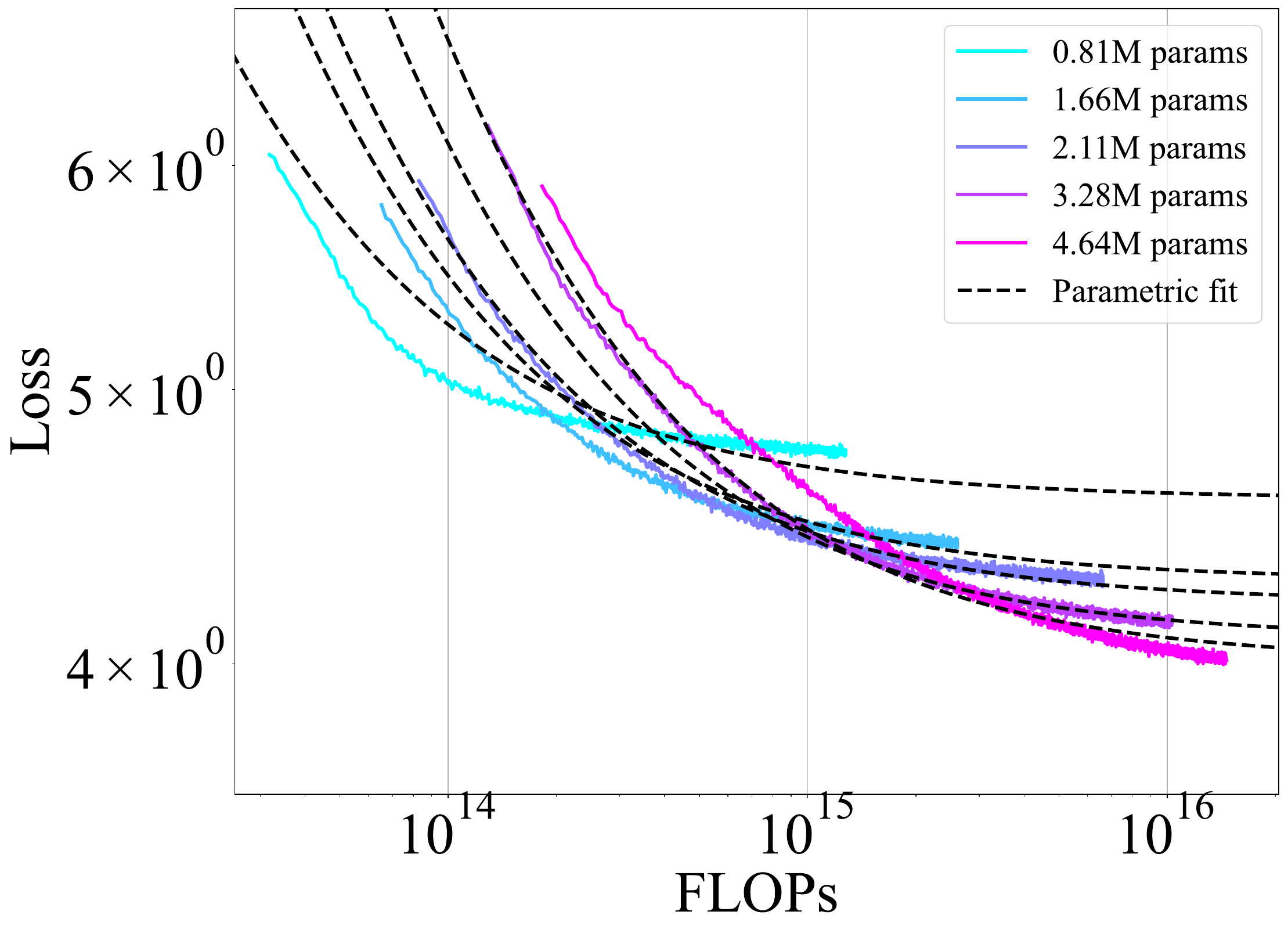}
\includegraphics[width=0.32\columnwidth]{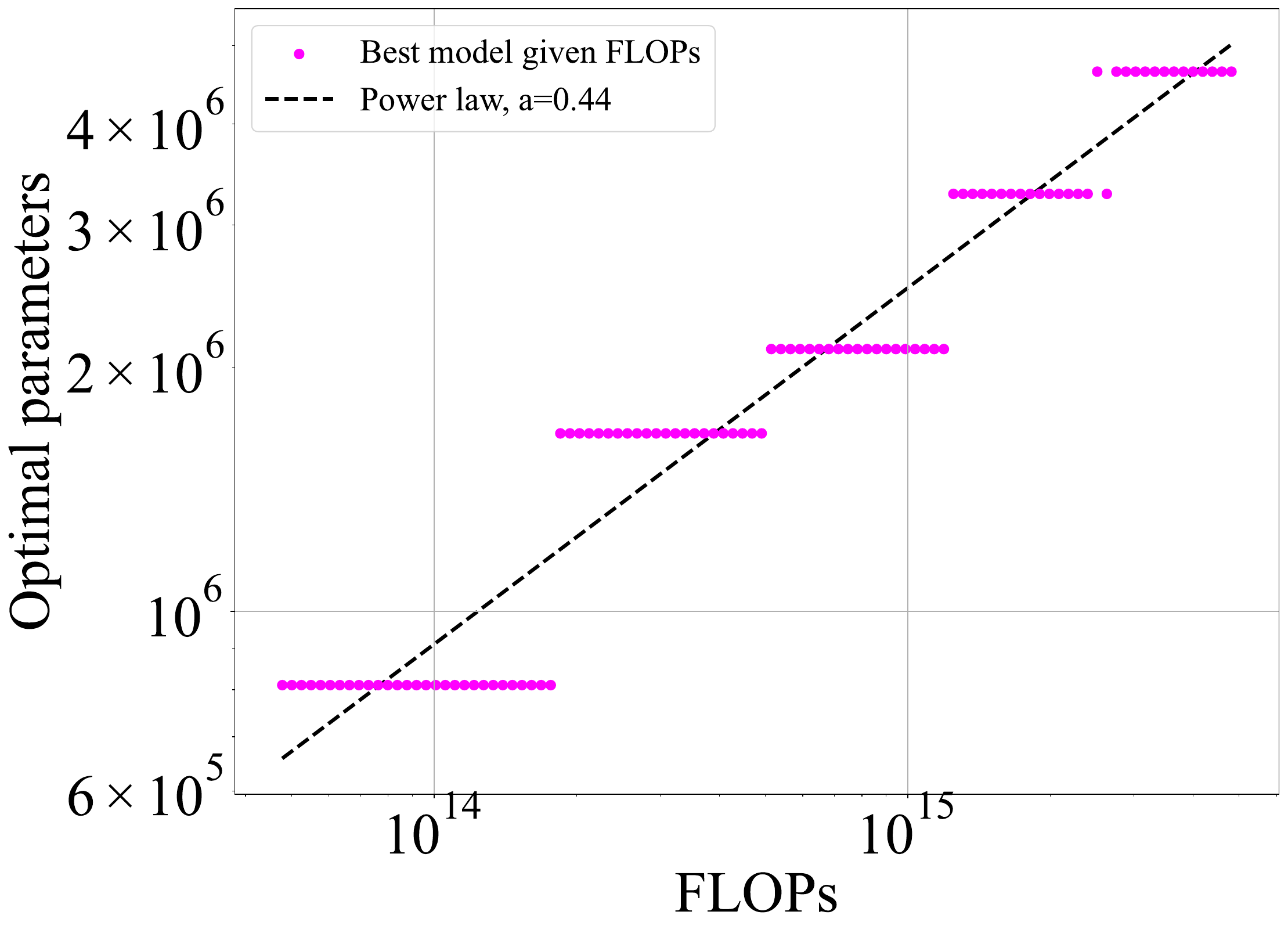}
\includegraphics[width=0.32\columnwidth]{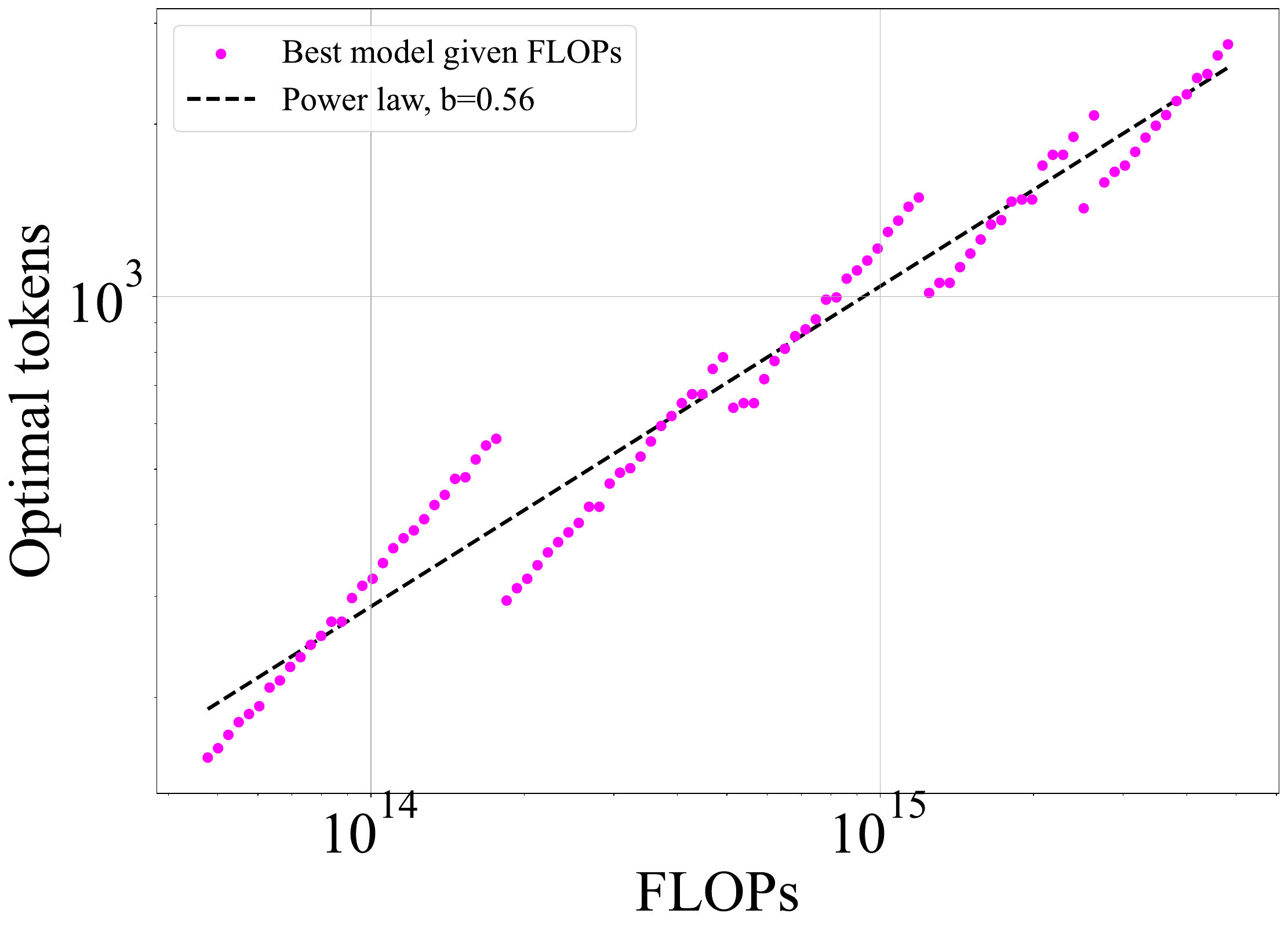}
\caption{Relating to Section \ref{sec_understand_q3}, GPT-2 tokenizer (high compression).
Utilising the \textit{frontier} fit (middle and right) we derive the power law coefficient for $N_\text{optimal}$ as $0.44$ and $D_\text{optimal}$ as $0.56$, an increase from $0.66$ in Figure \ref{fig_understand_compression_char} found when utilising a lower compression character-level tokenizer.
}
\label{fig_understand_compression_token}
\end{center}
\end{figure}

\newpage
\section{World modeling for robotics experimental details}
\label{sec_app_robotics}

This section provides experimental details for WM experiments on the secondary RT-1 dataset.

\subsection{Dataset}
\label{sec_appendix_rt1_dataset}

We resized the RT-1 dataset to 128x128 pixels per image. 
For action labels, we take the 3D \verb|world_vector| coordinates, combined with the 1D \verb|gripper_closedness_action| vector, to make an action vector with four dimensions. All are in the range -1 to 1, and these are discretized into 500 evenly spaced buckets.

\subsection{VQVAEs}

\begin{figure}[h!]
\begin{center}
\includegraphics[width=0.99\columnwidth]{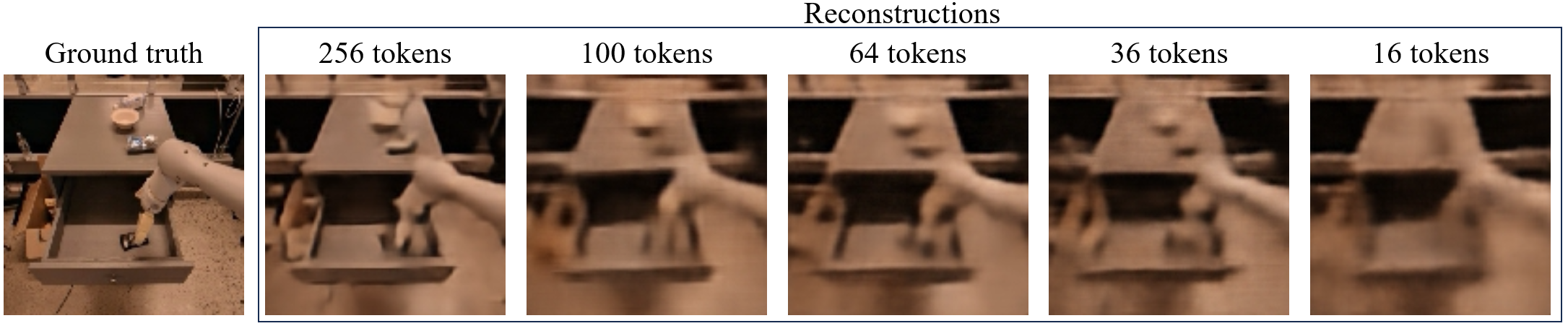}
\caption{VQVAE reconstructions on the RT-1 dataset for differing numbers of tokens per observation, $z_o \in [16, 36, 64, 100, 256]$.
}
\label{fig_rt1_reconstruct}
\end{center}
\end{figure}

We trained a set of five VQVAEs using the implementation from \url{https://github.com/nadavbh12/VQ-VAE}.
We set $z_o \in [16, 36, 64, 100, 256]$ and $V_o = 4096$, training each VQVAE for 40,000 updates on batches of 128. Reconstructions are visualized in Figure \ref{fig_rt1_reconstruct}.

\subsection{Transformer training details}

\begin{figure}[h!]
\begin{center}
$z_o = 16$, $N_\text{optimal} \propto C^{0.56}$, $N_\text{optimal} \propto D^{0.44}$ \\
\includegraphics[width=0.32\columnwidth]{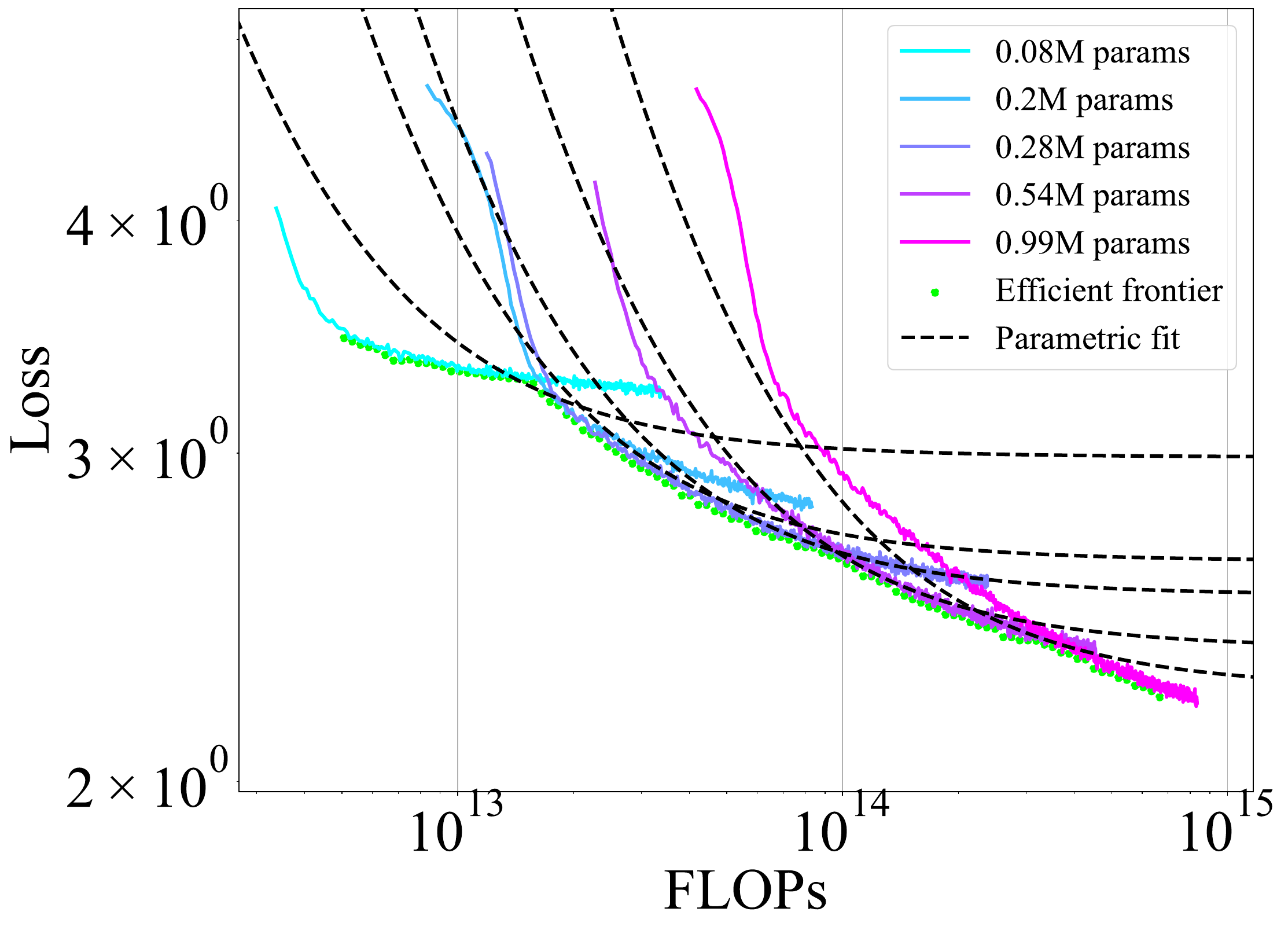}
\includegraphics[width=0.32\columnwidth]{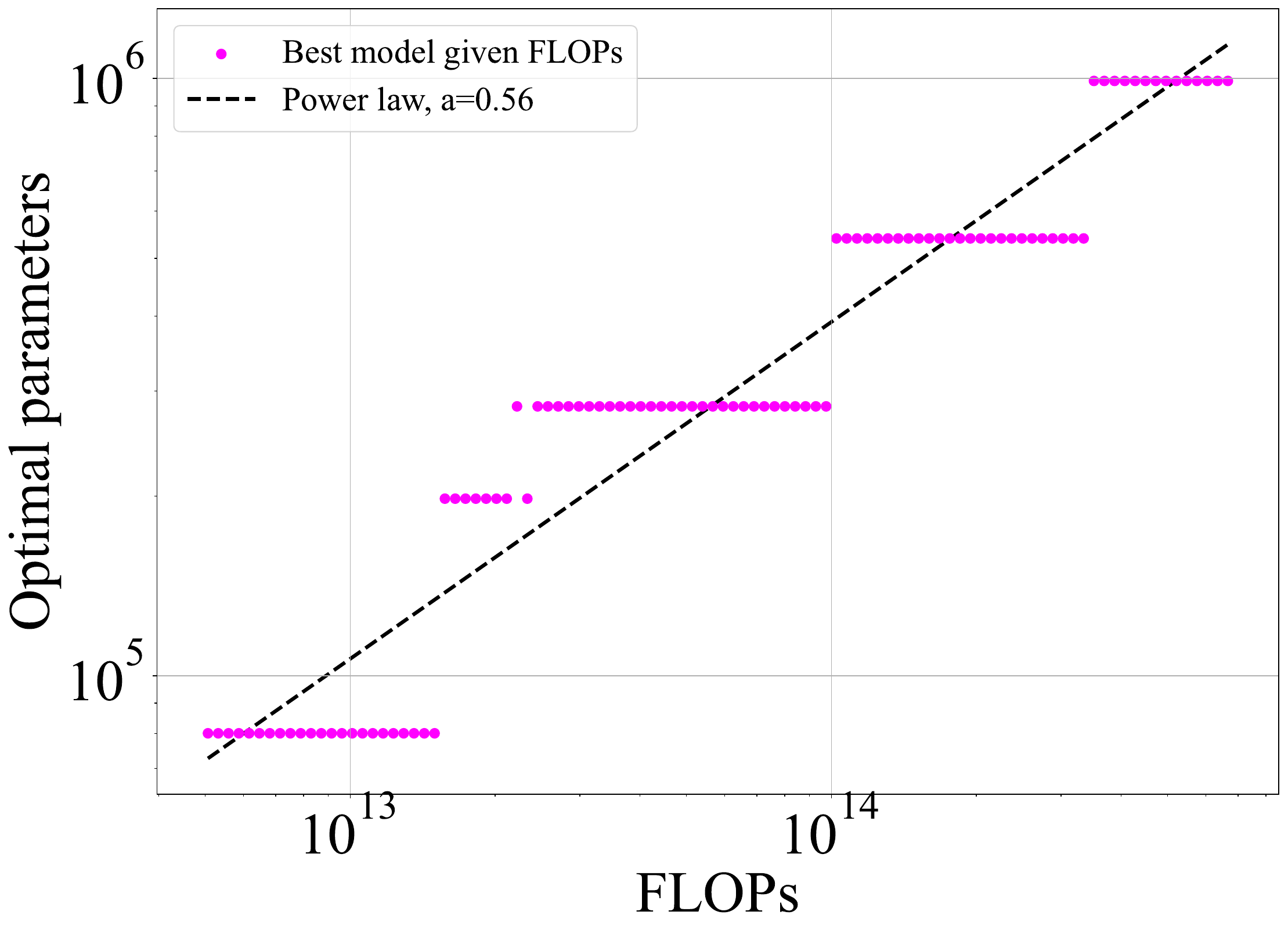}
\includegraphics[width=0.32\columnwidth]{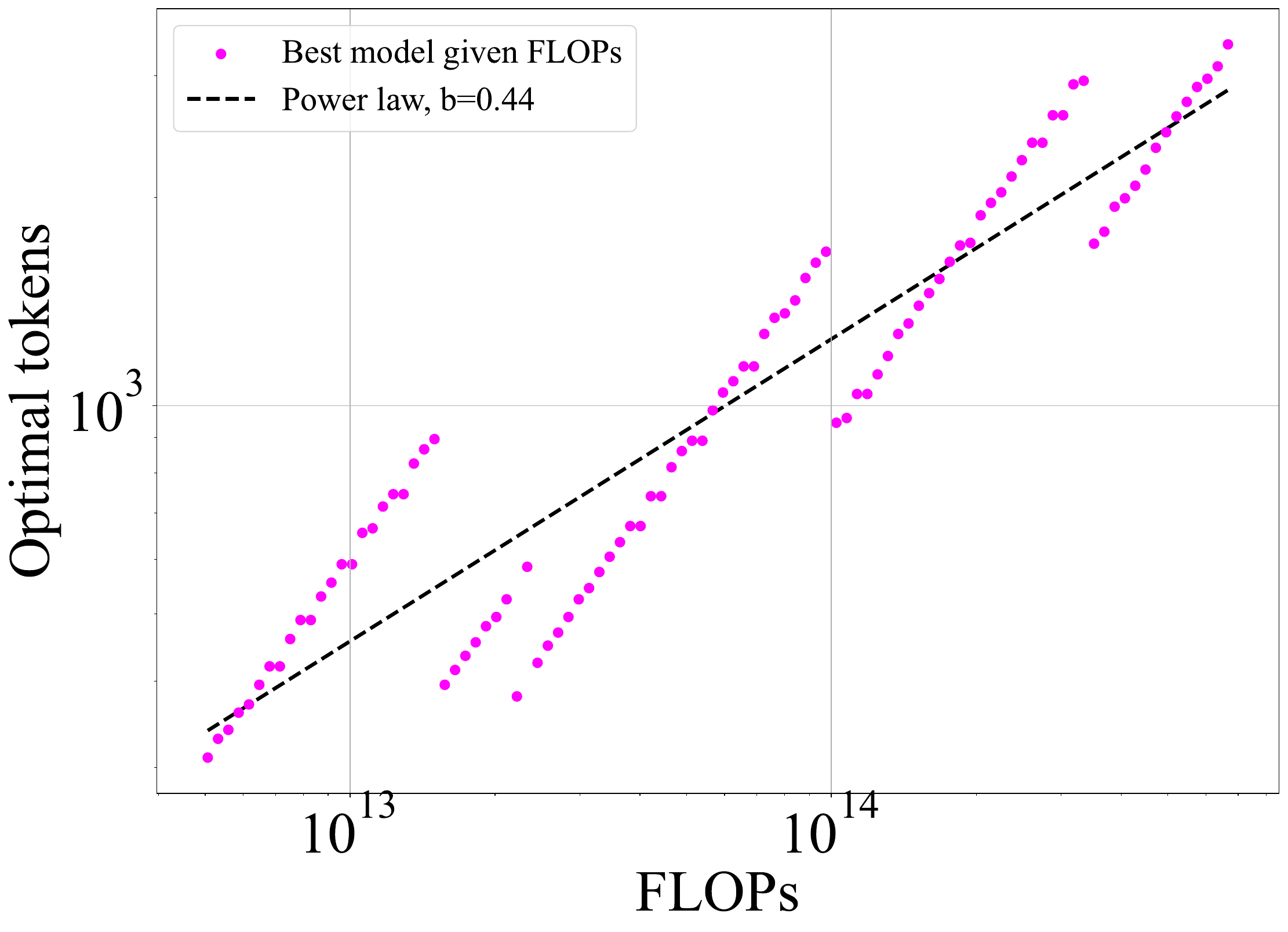} \\

$z_o = 36$, $N_\text{optimal} \propto C^{0.60}$, $N_\text{optimal} \propto D^{0.40}$ \\
\includegraphics[width=0.32\columnwidth]{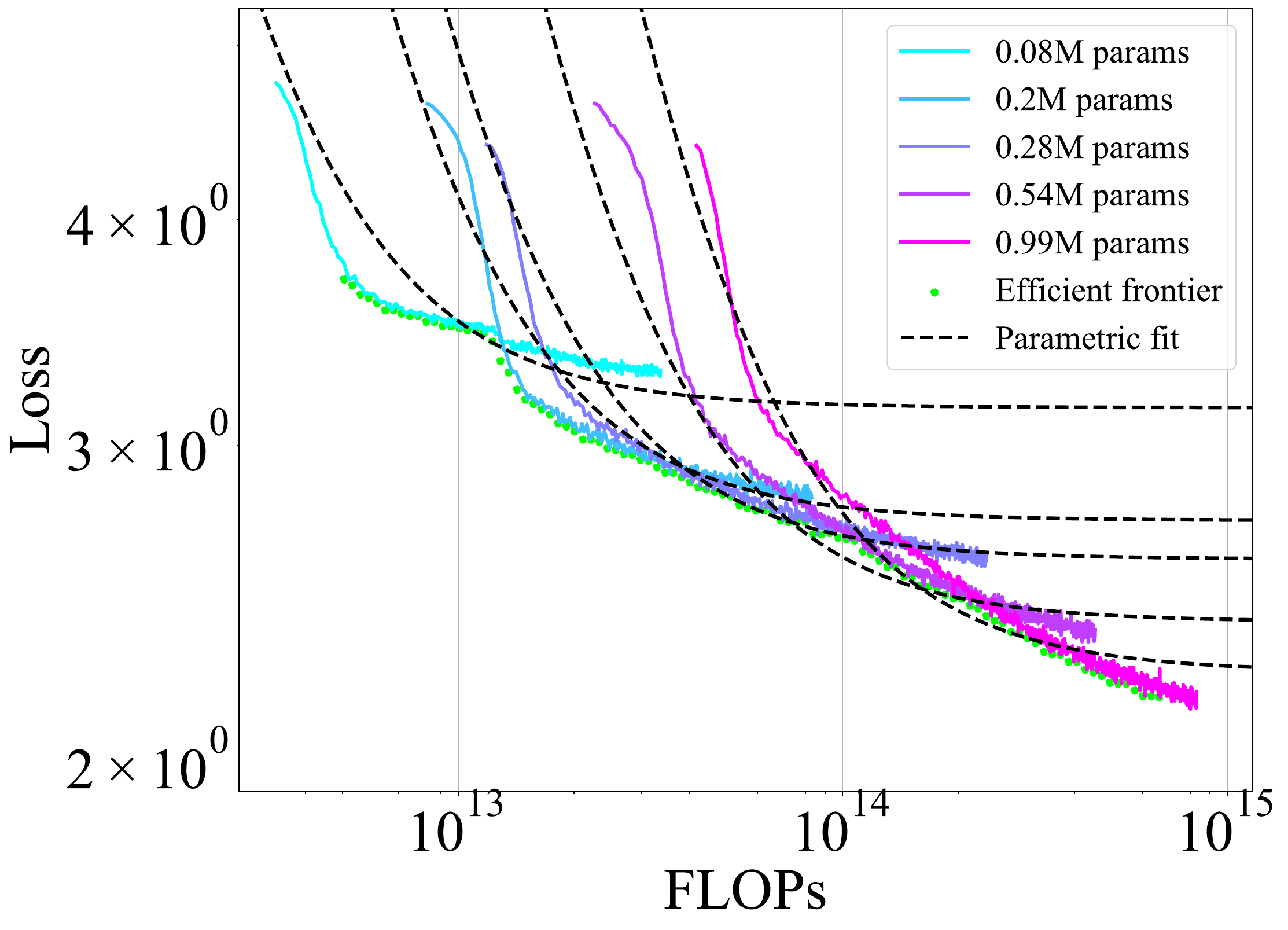}
\includegraphics[width=0.32\columnwidth]{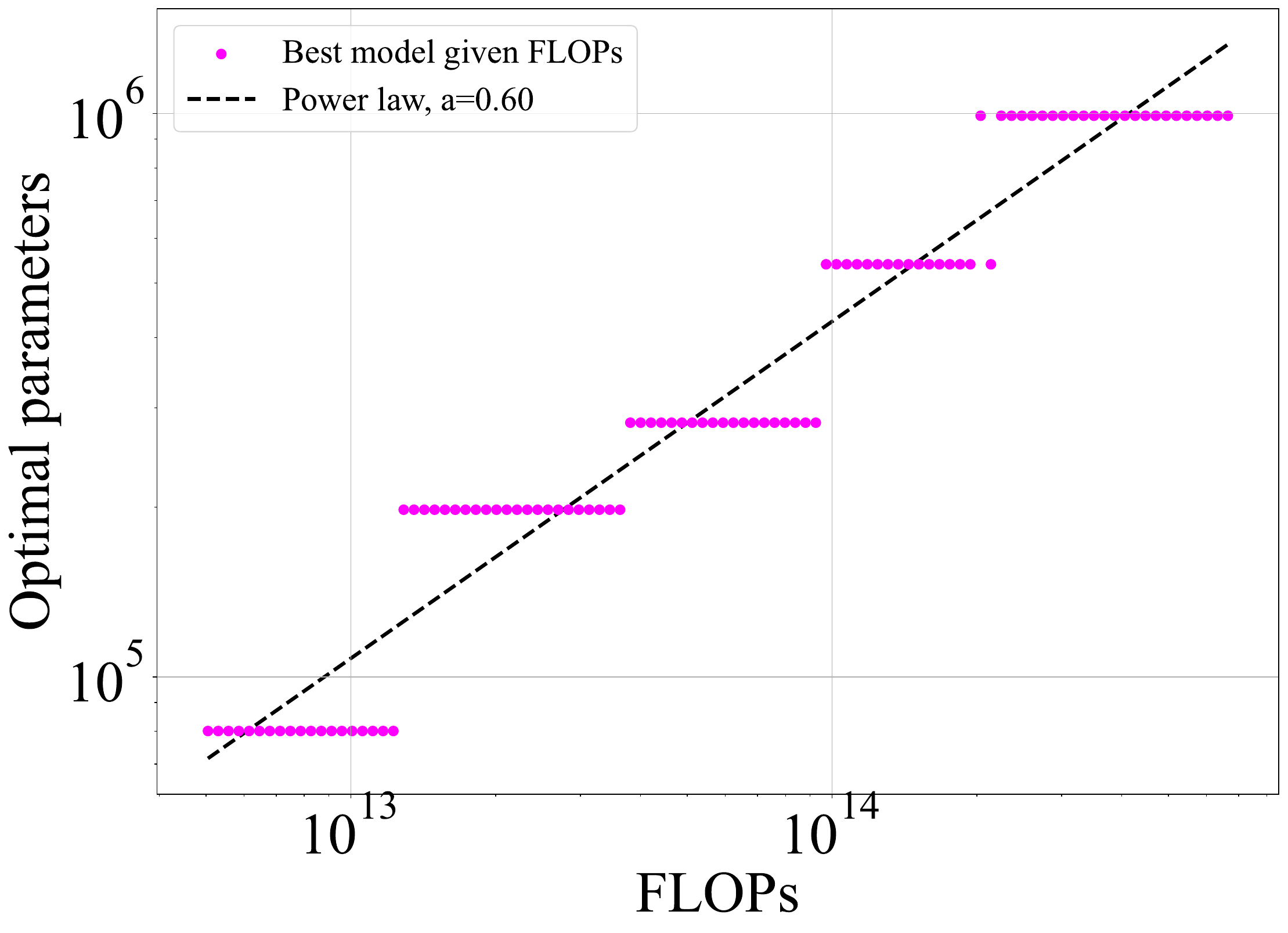}
\includegraphics[width=0.32\columnwidth]{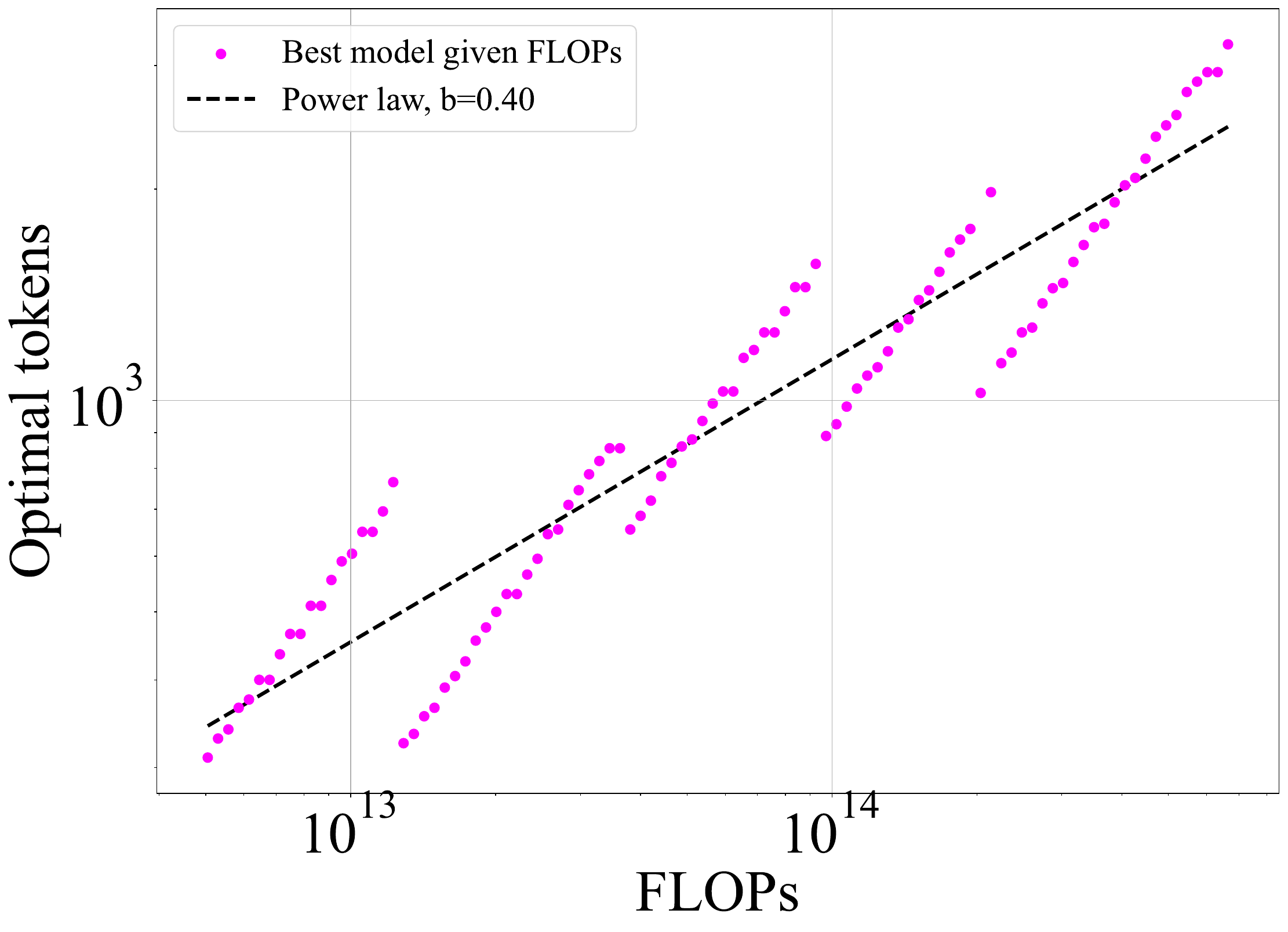} \\

$z_o = 64$, $N_\text{optimal} \propto C^{0.61}$, $N_\text{optimal} \propto D^{0.39}$ \\
\includegraphics[width=0.32\columnwidth]{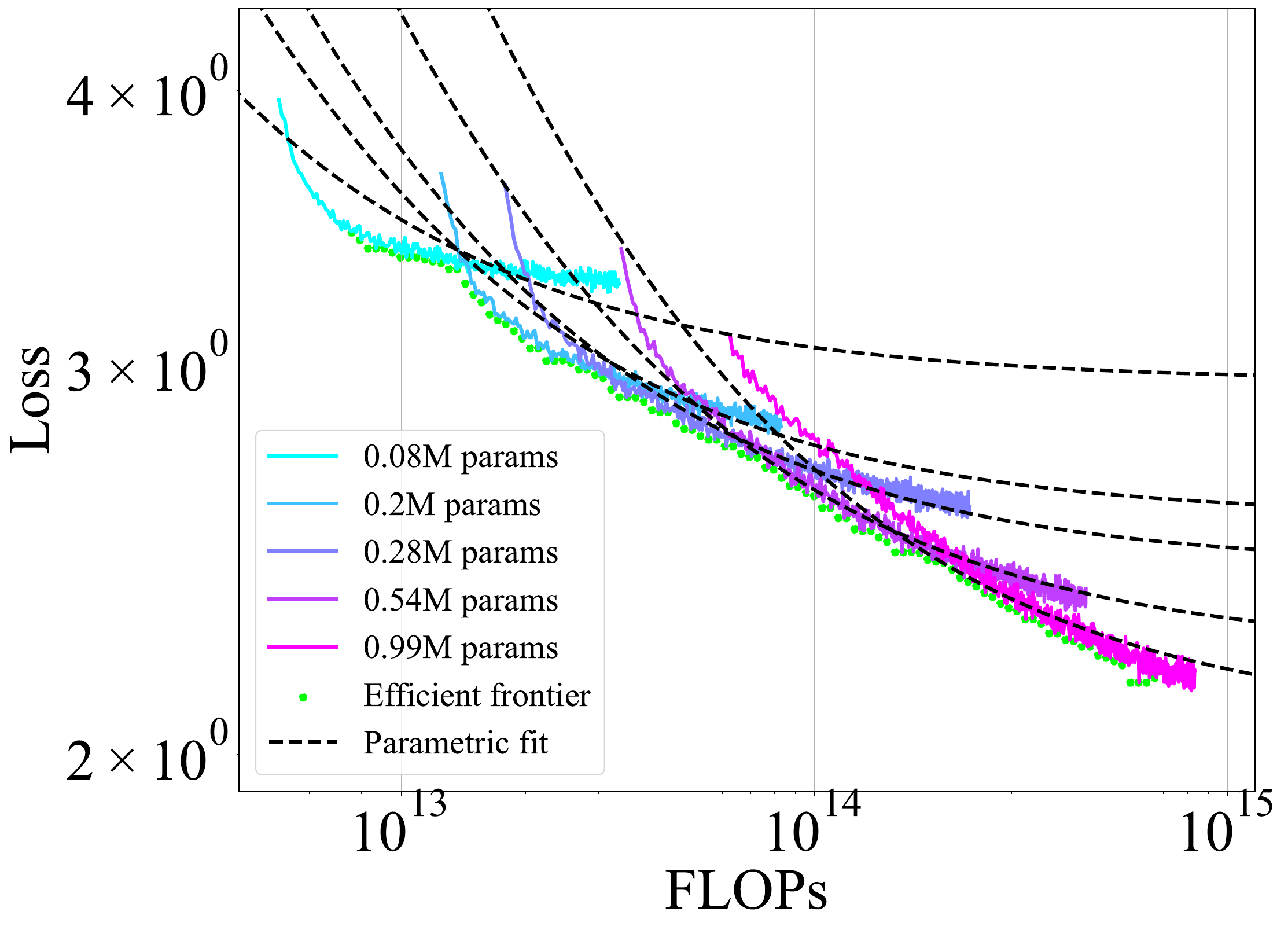}
\includegraphics[width=0.32\columnwidth]{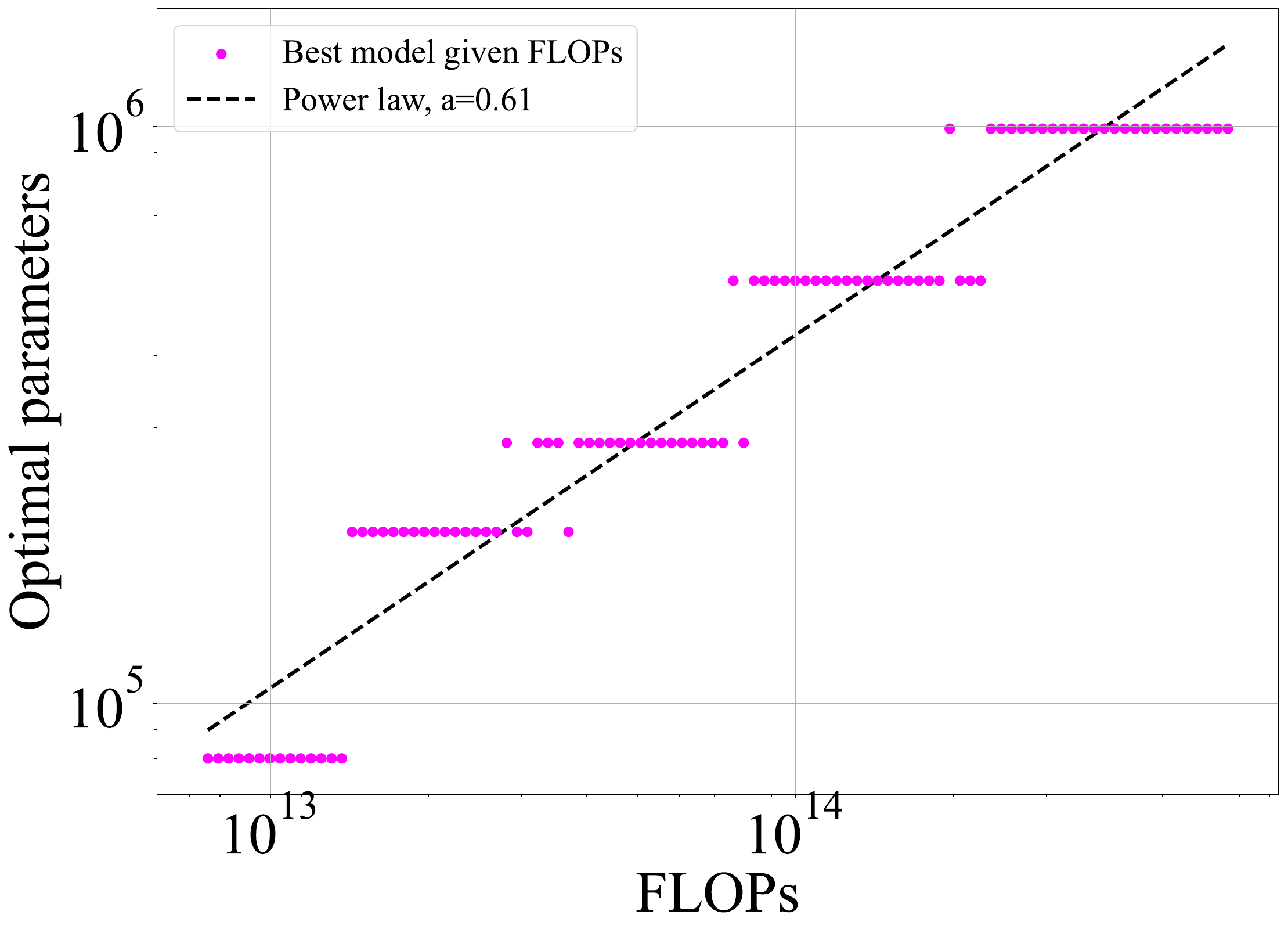}
\includegraphics[width=0.32\columnwidth]{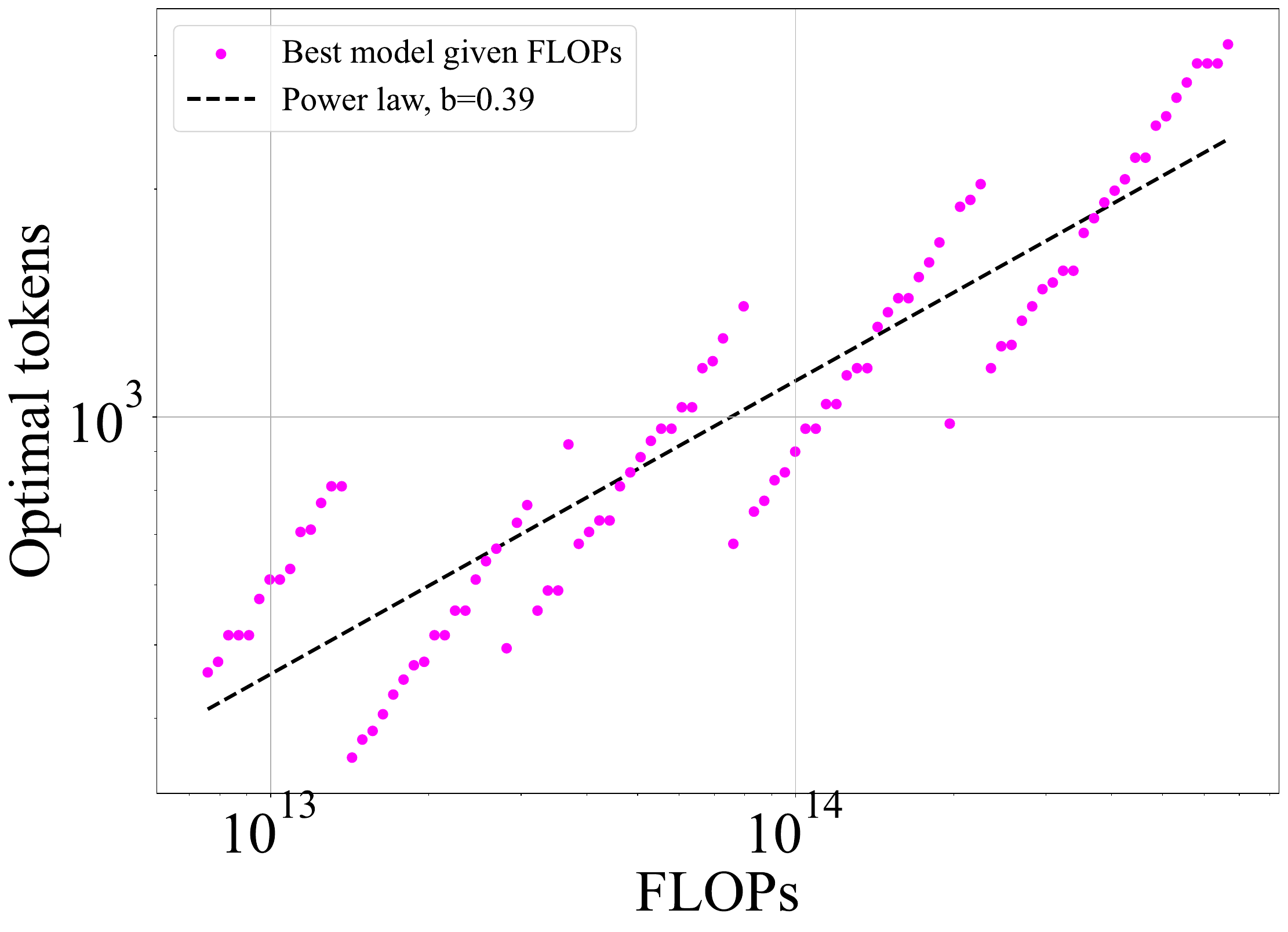} \\

$z_o = 100$, $N_\text{optimal} \propto C^{0.60}$, $N_\text{optimal} \propto D^{0.40}$ \\
\includegraphics[width=0.32\columnwidth]{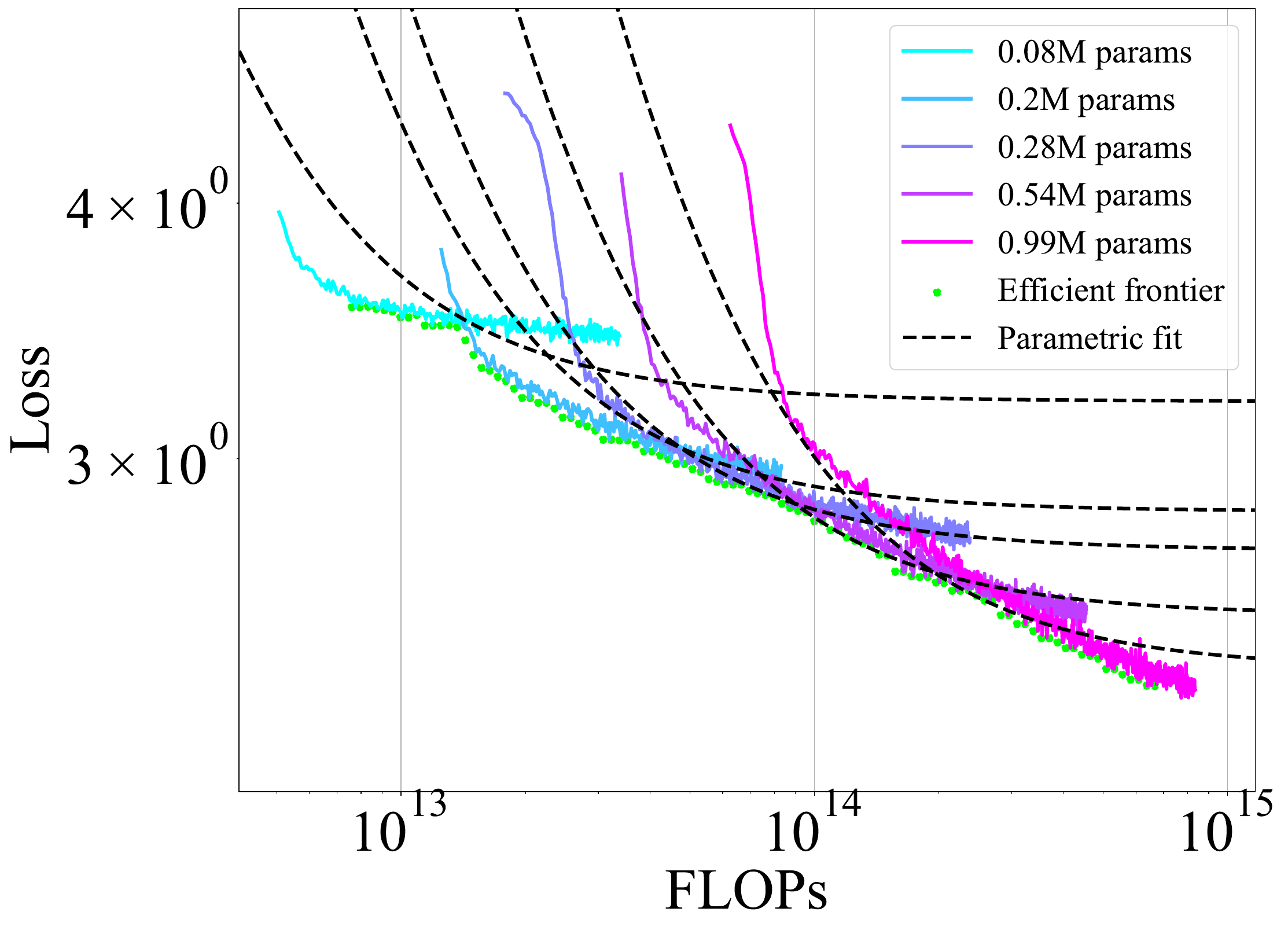}
\includegraphics[width=0.32\columnwidth]{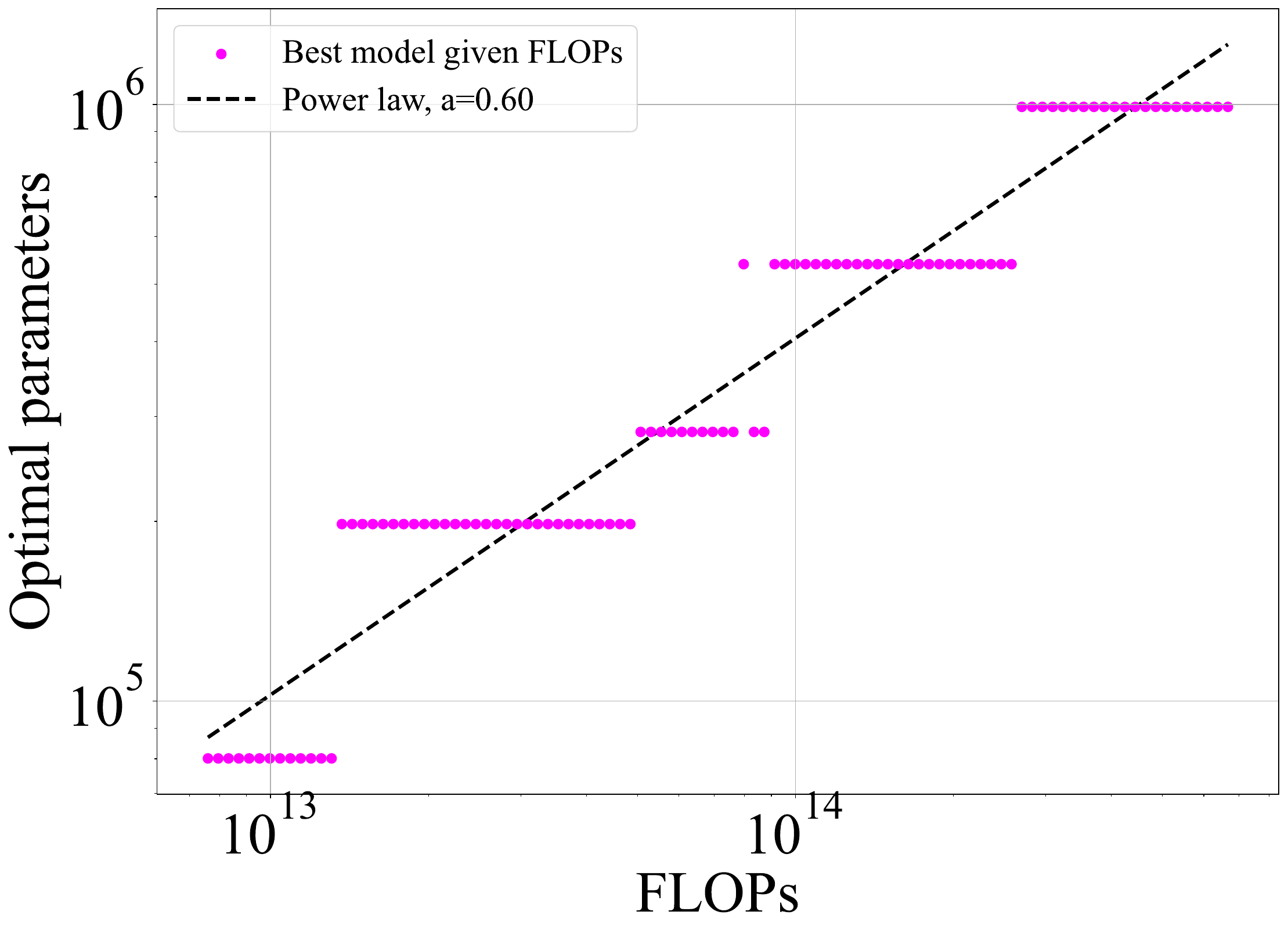}
\includegraphics[width=0.32\columnwidth]{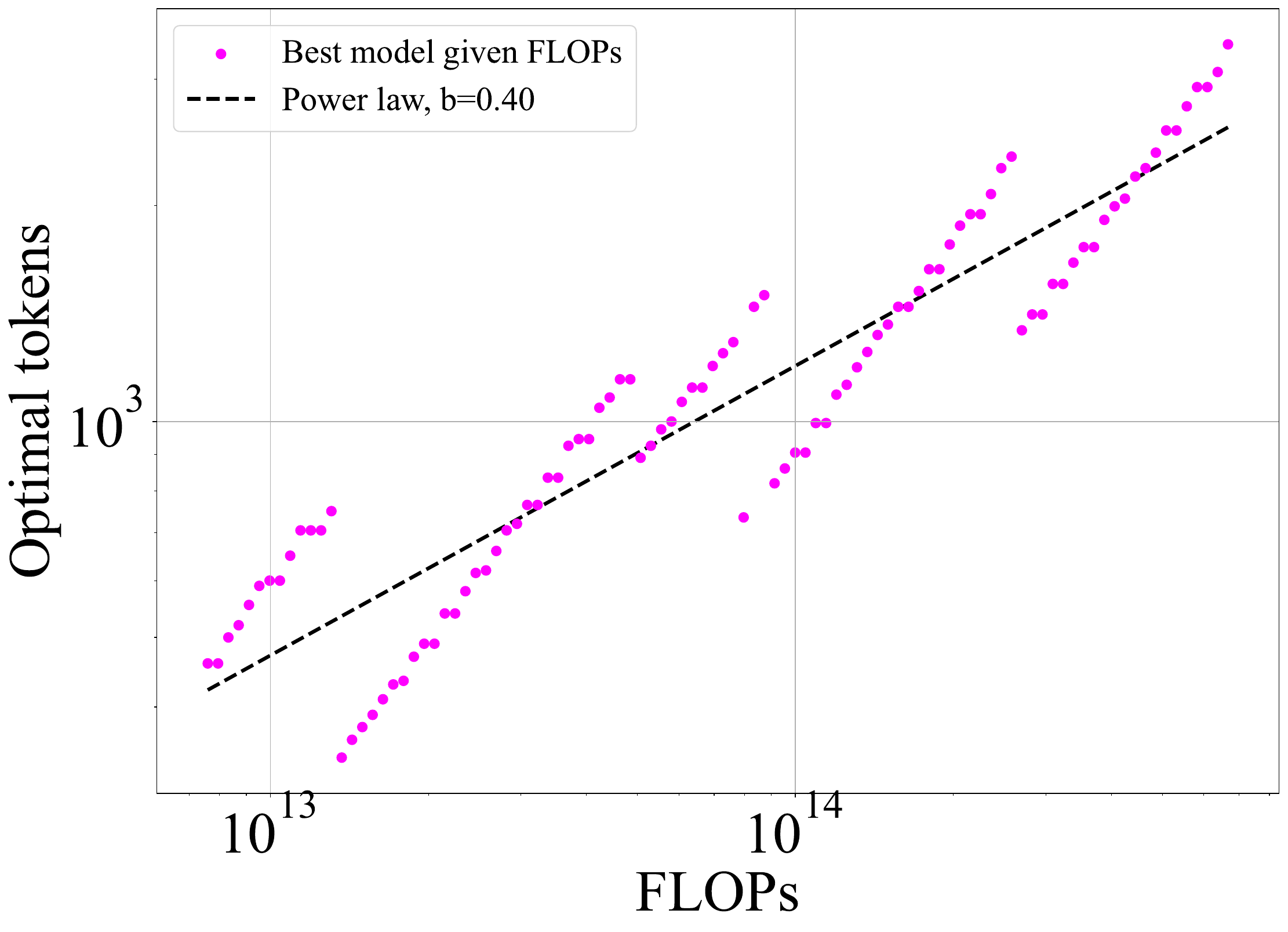} \\

$z_o = 256$, $N_\text{optimal} \propto C^{0.65}$, $N_\text{optimal} \propto D^{0.34}$ \\
\includegraphics[width=0.32\columnwidth]{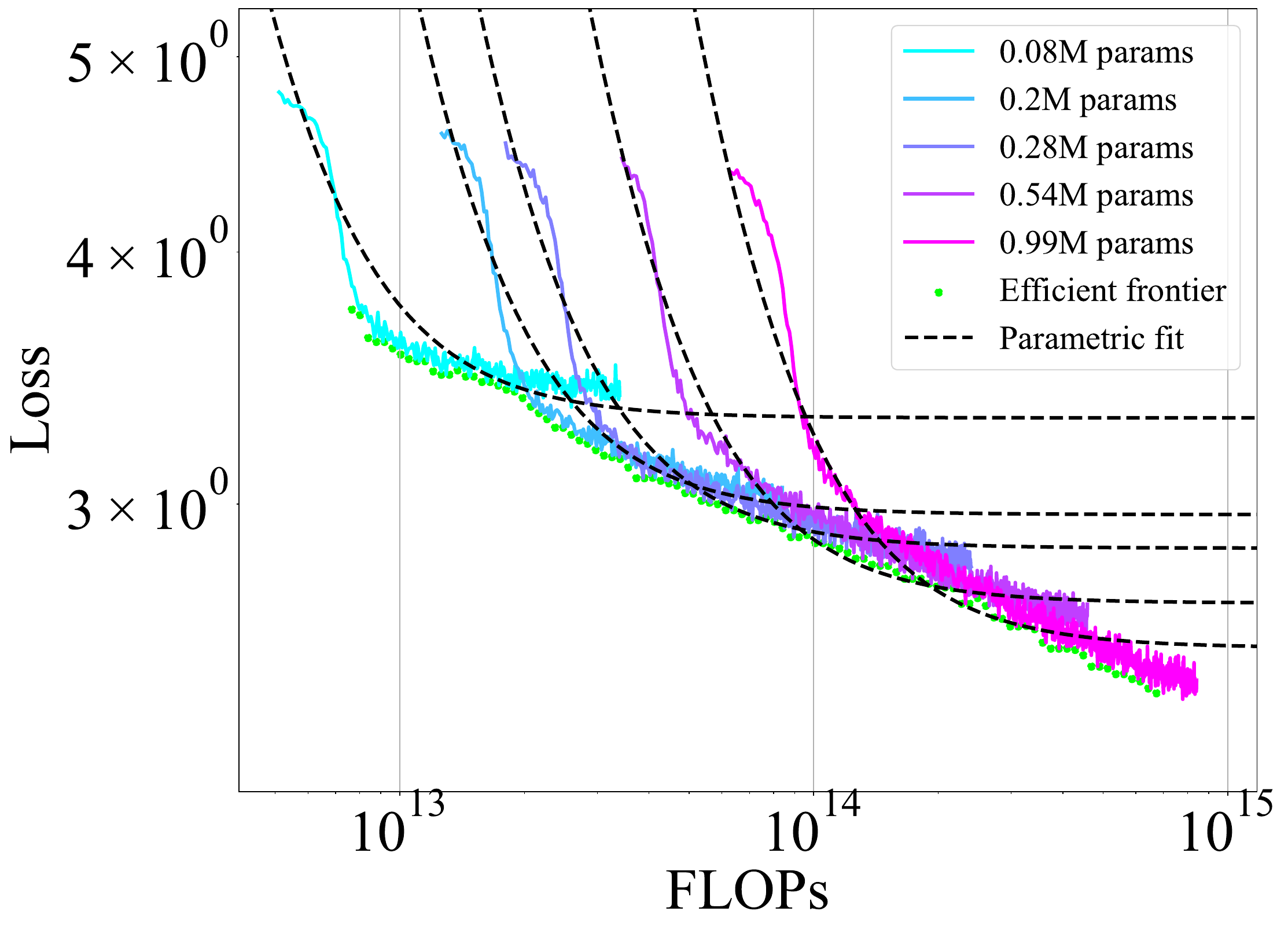}
\includegraphics[width=0.32\columnwidth]{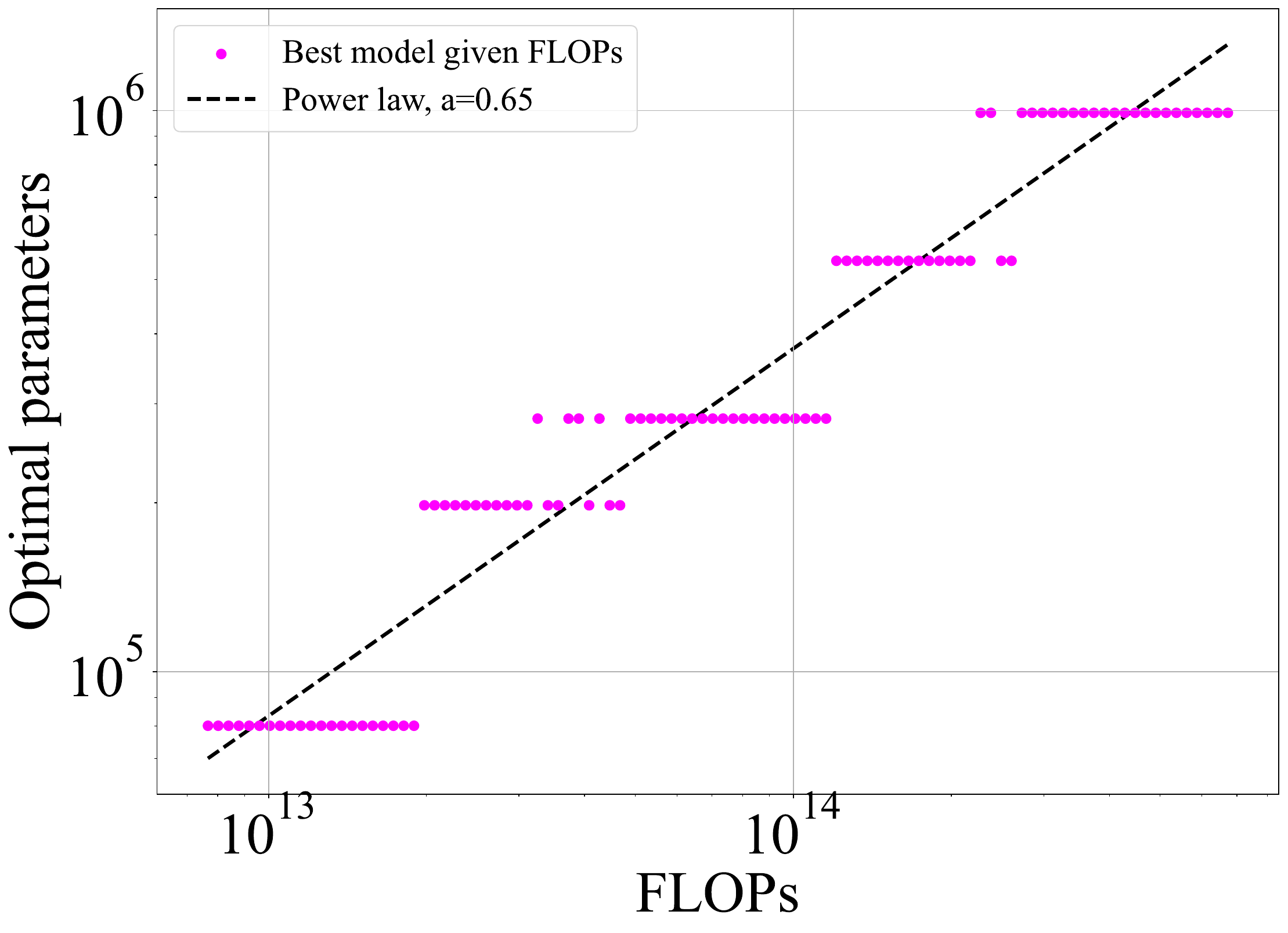}
\includegraphics[width=0.32\columnwidth]{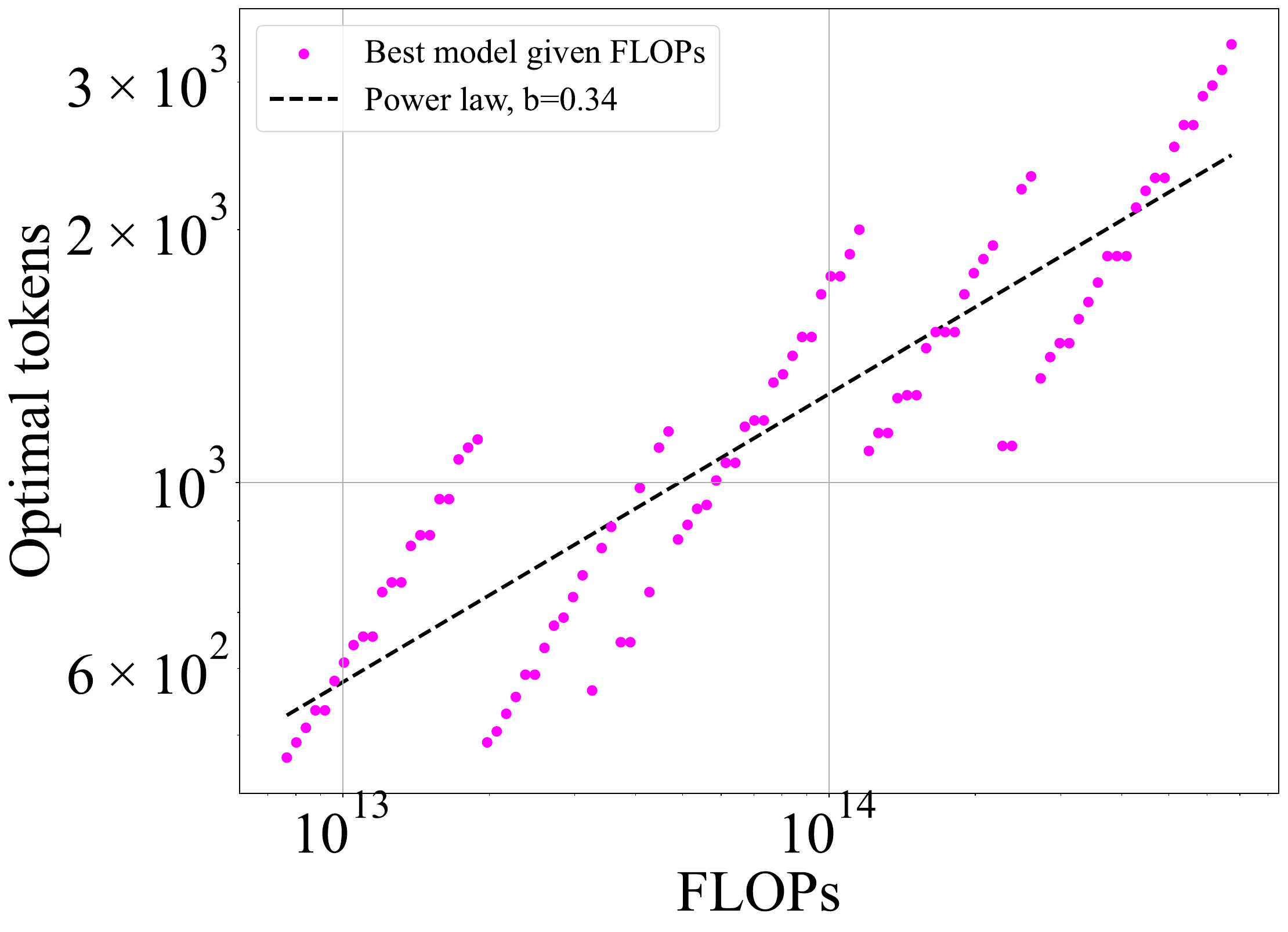} \\

\caption{RT-1 experiments. Note that the optimal parameter coefficient increases with the number of tokens per observation.}
\label{fig_rt1_scaling}
\end{center}
\end{figure}

Table \ref{tbl_training_hyperparams_WM_RT1} provides training details for the model sizes tested. Figure \ref{fig_rt1_scaling} shows one example set of training curves per VQVAE.  

\begin{table}[h!]
\caption{Hyperparameters for WM-Token in RT-1 experiments.}
\begin{center}
\label{tbl_training_hyperparams_WM_RT1}
\begin{tabular}{lrrrrr}
     \multicolumn{1}{c}{\bf $N$}  
     & \multicolumn{1}{c}{\bf Seq len} 
     & \multicolumn{1}{c}{\bf Context length} 
     & \multicolumn{1}{c}{\bf Tokens per update} 
     & \multicolumn{1}{c}{\bf Learning rate} 
     \\ \hline \\
    0.08M & 2 & 2$(z_o + 4)$ & 34,000 & 0.01 \\
    0.2M & 2 & 2$(z_o + 4)$ & 34,000 & 0.005 \\
    0.28M & 2 & 2$(z_o + 4)$ & 34,000 & 0.004 \\
    0.54M & 2 & 2$(z_o + 4)$ & 34,000 & 0.0027 \\
    0.99M & 2 & 2$(z_o + 4)$ & 34,000 & 0.002 \\
\end{tabular}
\end{center}
\end{table}

\FloatBarrier
\newpage
\section{Pre-training loss vs. world modeling metrics}
\label{sec_app_pretrain_evidence}

This section presents evidence for pre-training loss correlating with WM performance.
We use metrics commonly used to assess the quality of the world models \citep{yang2023unisim}, originally developed in the video generation literature. Conditioned on an initial real frame and a sequence of real actions, we compare the observations generated by a world model, with the real sequence of observations, measuring FVD and LPIPS. Specifically, we generate 1024 videos each of 10 seconds. We perform this for various checkpoints on each size in our WM-Token-256 set of models. This allows a plot of the checkpoint pre-training loss vs video generation metric to be assessed. 

Figure \ref{fig_wmtok256_loss_vs_metrics_main} shows results.
We find correlations of 0.77, 0.83 for LPIPS and FVD respectively. Two early checkpoints from the 894M model are the only significant anomalies to trend of metrics improving with loss.
This evidences the strong relationship between pre-training loss and world model quality.

\end{document}